\documentclass{article}

\usepackage{microtype}
\usepackage{graphicx}
\usepackage{subfigure}
\usepackage{booktabs} 

\usepackage{hyperref}

\usepackage[accepted]{icml2025}

\usepackage{amssymb}
\usepackage{mathtools}
\usepackage{amsthm}
\usepackage{xcolor} 
\usepackage{wrapfig}
\usepackage{wrapfig}
\usepackage{multirow}
\usepackage{makecell}
\usepackage{subfigure}
\usepackage{placeins}
\usepackage{longtable}
\usepackage{setspace} 
\usepackage{stix} 
\usepackage[font=small,labelfont=it]{caption} 
\makeatletter
\renewcommand{\@makecaption}[2]{%
  \vskip 10pt
  \baselineskip 11pt
  \sbox\@tempboxa{\small\textit{#1.} #2}%
  \ifdim \wd\@tempboxa >\hsize
    \sbox\@tempboxa{\small\textit{#1.} }%
    \parbox[t]{\hsize}{\usebox\@tempboxa {\footnotesize #2}}%
  \else
    \centerline{\usebox\@tempboxa}%
  \fi}
\makeatother

\usepackage[capitalize,noabbrev]{cleveref}

\theoremstyle{plain}

\theoremstyle{definition}

\theoremstyle{remark}

\usepackage[textsize=tiny]{todonotes}

\newcommand{\data}{\textbf{s1K}}
\newcommand{\model}{\textbf{s1-32B}}

\icmltitlerunning{\textbf{s1}: Simple test-time scaling}

\makeatletter
\AtBeginDocument{
\renewcommand{\sectionautorefname}{\S\@gobble}

\renewcommand{\subsectionautorefname}{\S\@gobble} 
\renewcommand{\subsubsectionautorefname}{\S\@gobble}
\renewcommand{\appendixautorefname}{\S\@gobble}

}
\makeatother

\definecolor{olmoeDarkYellow}{HTML}{fdac15}
\definecolor{defaultblue}{HTML}{0077B6}
\definecolor{defaultlightblue}{HTML}{00B4D8}
\definecolor{blue}{HTML}{03045E}
\definecolor{blueb}{HTML}{0077B6}
\definecolor{bluec}{HTML}{00B4D8}
\definecolor{blued}{HTML}{90E0EF}
\definecolor{bluee}{HTML}{CAF0F8}

\newcommand{\defaultblue}[1]{{\leavevmode\color{blueb}#1}}
\newcommand{\defaultlightblue}[1]{{\leavevmode\color{bluec}#1}}
\newcommand{\defaultyellow}[1]{{\leavevmode\color{olmoeDarkYellow}#1}}

\makeatletter
\let\@oldmaketitle\@maketitle
\renewcommand{\@maketitle}{\@oldmaketitle
  \includegraphics[width=\linewidth,height=4\baselineskip]
    {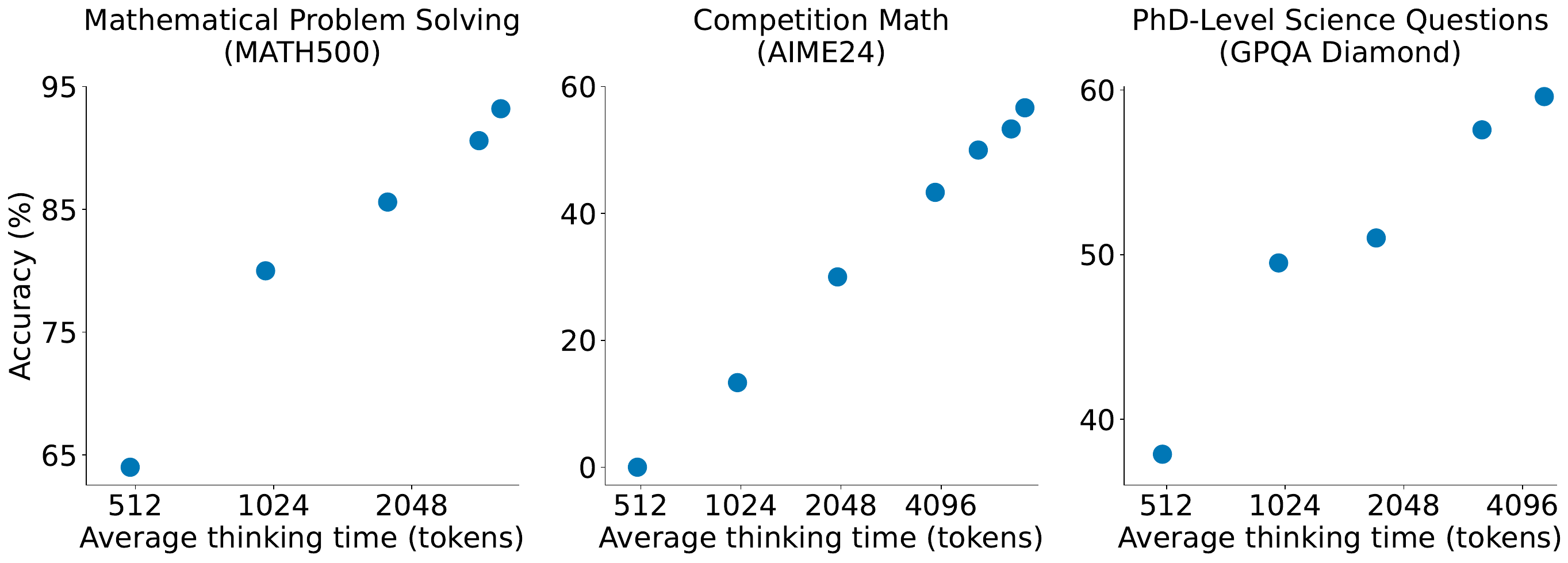}\bigskip}
\makeatother

\begin{document}

\twocolumn[{
\icmltitle{\textbf{s1}: Simple test-time scaling}

\icmlsetsymbol{equal}{*}

\begin{icmlauthorlist}
\icmlauthor{Niklas Muennighoff}{equal,s,a,c}
\icmlauthor{Zitong Yang}{equal,s}
\icmlauthor{Weijia Shi}{equal,w,a}
\icmlauthor{Xiang Lisa Li}{equal,s}
\icmlauthor{Li Fei-Fei}{s}
\icmlauthor{Hannaneh Hajishirzi}{w,a}
\icmlauthor{Luke Zettlemoyer}{w}
\icmlauthor{Percy Liang}{s}
\icmlauthor{Emmanuel Cand\`es}{s}
\icmlauthor{Tatsunori Hashimoto}{s}
\end{icmlauthorlist}

\icmlaffiliation{s}{Stanford University}
\icmlaffiliation{w}{University of Washington, Seattle}
\icmlaffiliation{a}{Allen Institute for AI}
\icmlaffiliation{c}{Contextual AI}


\icmlkeywords{Machine Learning, ICML, Large language models, Test-time scaling, Test-time compute}

\vskip 0.3in
}]

\printAffiliationsAndNotice{\icmlEqualContribution}

\begin{abstract}
Test-time scaling is a promising new approach to language modeling that uses extra test-time compute to improve performance. Recently, OpenAI's o1 model showed this capability but did not publicly share its methodology, leading to many replication efforts. We seek the simplest approach to achieve test-time scaling and strong reasoning performance. First, we curate a small dataset \data{} of 1,000 questions paired with reasoning traces relying on three criteria we validate through ablations: difficulty, diversity, and quality. Second, we develop budget forcing to control test-time compute by forcefully terminating the model's thinking process or lengthening it by appending ``Wait'' multiple times to the model's generation when it tries to end. This can lead the model to double-check its answer, often fixing incorrect reasoning steps. After supervised finetuning the Qwen2.5-32B-Instruct language model on \data{} and equipping it with budget forcing, our model \model{} exceeds o1-preview on competition math questions by up to 27\% (MATH and AIME24). Further, scaling \model{} with budget forcing allows extrapolating beyond its performance without test-time intervention: from 50\% to 57\% on AIME24. Our model, data, and code are open-source at \url{https://github.com/simplescaling/s1}.

\end{abstract}

\section{Introduction}
\label{sec:intro}

\begin{figure}
\centering
\includegraphics[width=\columnwidth]{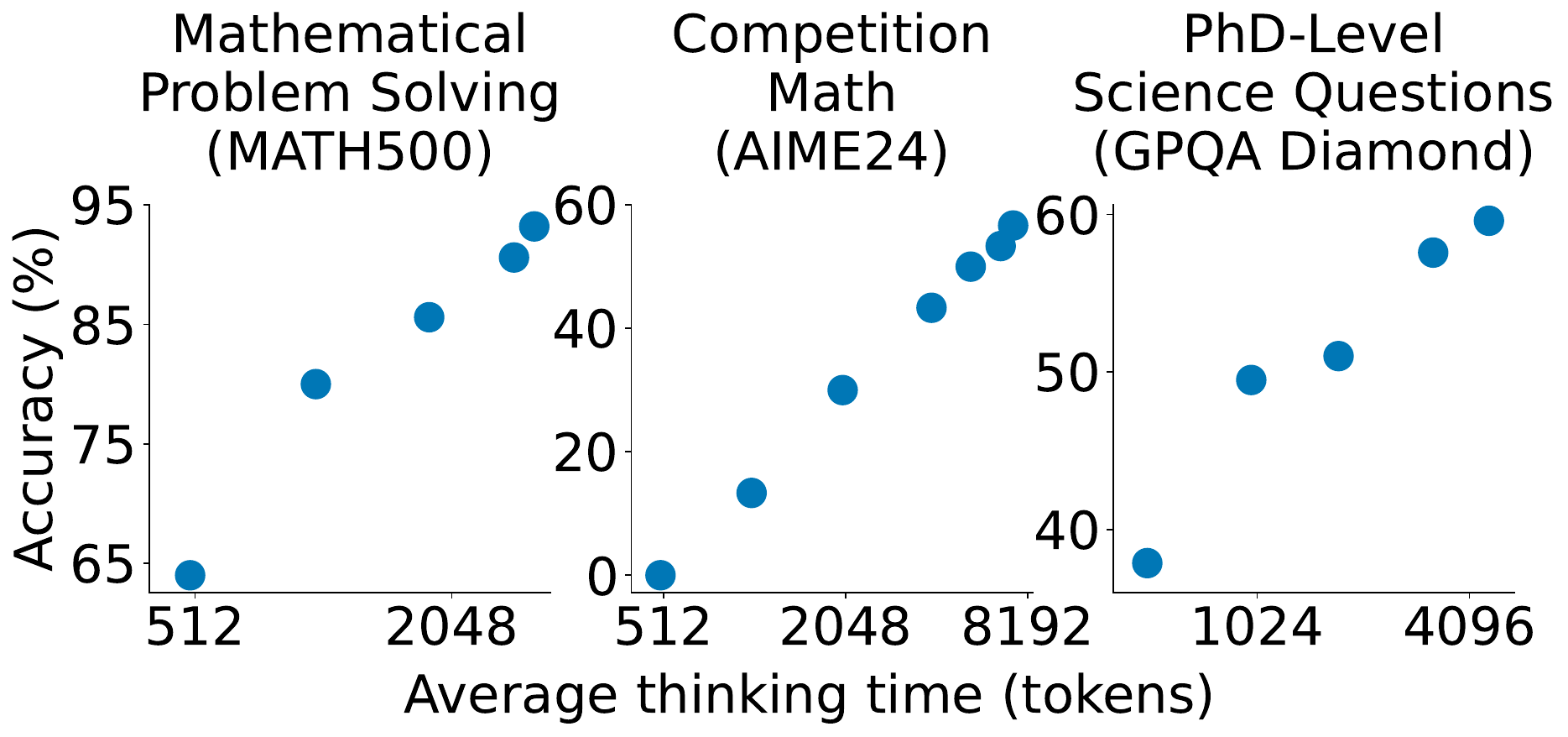}
\caption{\textbf{Test-time scaling with \model{}.} We benchmark \model{} on reasoning-intensive tasks and vary test-time compute.}
\label{fig:scaling}
\end{figure}

Performance improvements of language models (LMs) over the past years have largely relied on scaling up train-time compute using large-scale self-supervised  pretraining~\citep{kaplan2020scalinglawsneurallanguage,hoffmann2022trainingcomputeoptimallargelanguage}. The creation of these powerful models has set the stage for a new scaling paradigm built on top of them: \textit{test-time scaling}. The aim of this approach is to increase the compute at test time to get better results. There has been much work exploring this idea~\citep{snell2024scalingllmtesttimecompute,welleck2024decodingmetagenerationinferencetimealgorithms}, and the viability of this paradigm was recently validated by OpenAI o1~\citep{o1}. o1 has demonstrated strong reasoning performance with consistent gains from scaling test-time compute. OpenAI describes their approach as using large-scale reinforcement learning (RL) implying the use of sizable amounts of data~\citep{o1}. This has led to various attempts to replicate their models relying on techniques like Monte Carlo Tree Search~\citep{gao2024interpretablecontrastivemontecarlo,zhang2024o1codero1replicationcoding}, multi-agent approaches~\citep{qin2024o1replicationjourneystrategic}, and others~\citep{wang2024drto1optimizeddeepreasoning,huang2024o1replicationjourney,huang2025o1replicationjourney}. Among these approaches, DeepSeek R1~\citep{r1} has successfully replicated o1-level performance, also employing reinforcement learning via millions of samples and multiple training stages. However, despite the large number of o1 replication attempts, none have openly replicated a clear test-time scaling behavior. Thus, we ask: what is the simplest approach to achieve both test-time scaling and strong reasoning performance?

\begin{figure*}[t]
\centering
\subfigure{\includegraphics[width=0.45\textwidth]{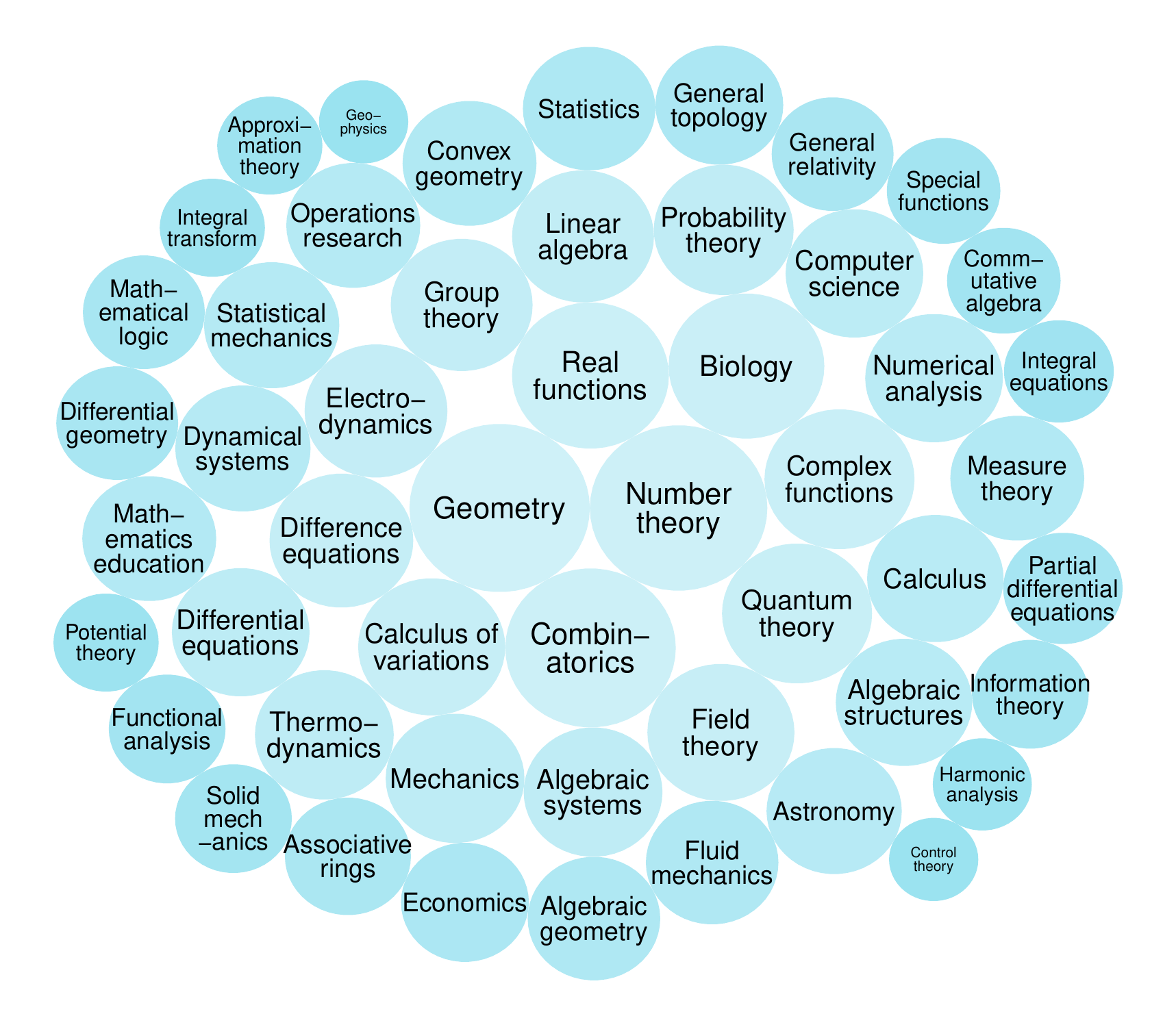}}
\subfigure{\includegraphics[width=0.5\textwidth]{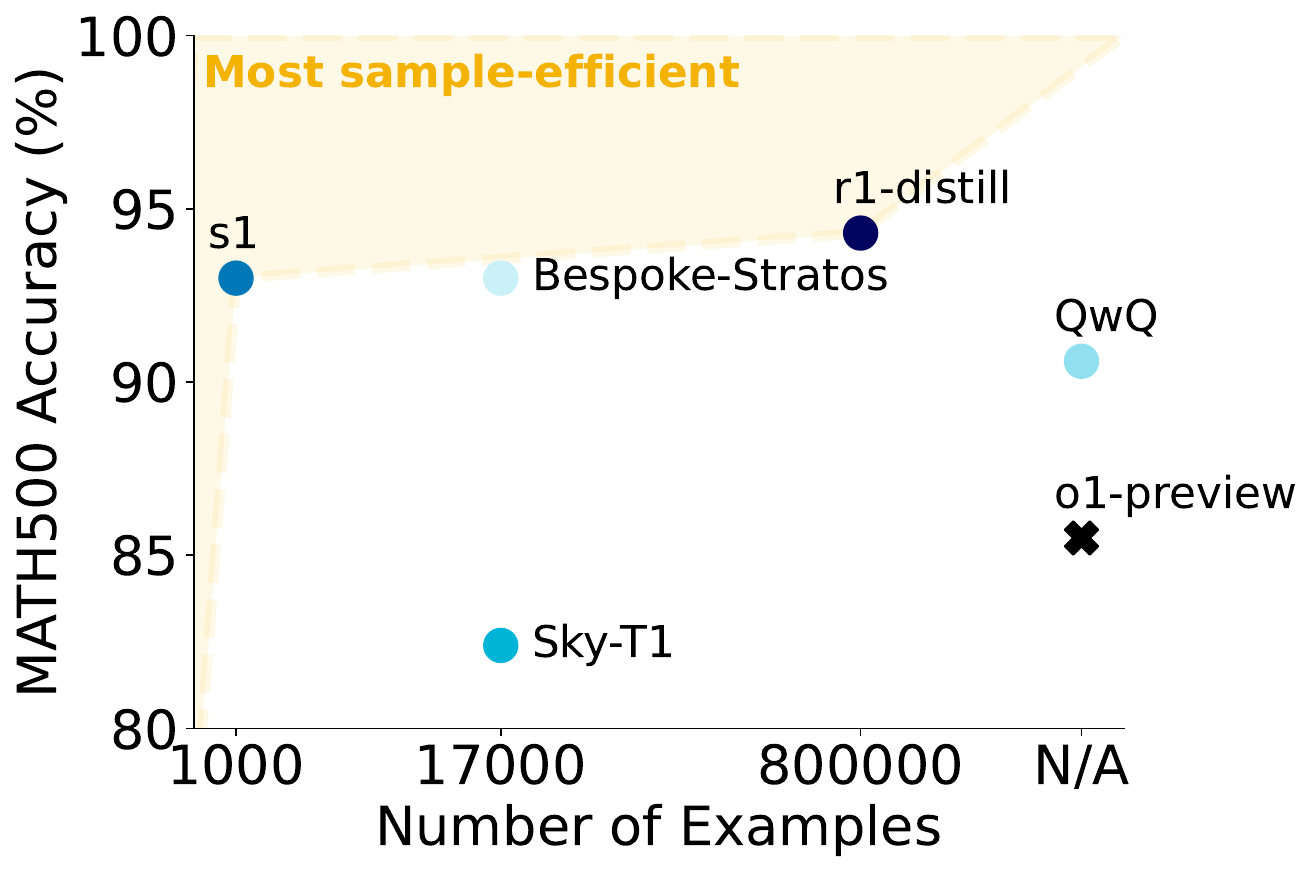}}
\caption{\textbf{\data{} and \model{}.} \textit{(left)}~\data{} is a dataset of 1,000 high-quality, diverse, and difficult questions with reasoning traces. \textit{(right)}~\model{}, a 32B parameter model finetuned on \data{} is on the sample-efficiency frontier. See \autoref{tab:perf} for details on other models.}
\label{fig:s1k-bar}
\end{figure*}

We show that training on only 1,000 samples with next-token prediction and controlling thinking duration via a simple test-time technique we refer to as \textit{budget forcing} leads to a strong reasoning model that scales in performance with more test-time compute. Specifically, we construct \data{}, which consists of 1,000 carefully curated questions paired with reasoning traces and answers distilled from Gemini Thinking Experimental~\citep{geminithinking}. We perform supervised fine-tuning (SFT) of an off-the-shelf pretrained model on our small dataset requiring just 26 minutes of training on 16 H100 GPUs. After training, we control the amount of test-time compute our model spends using \textit{budget forcing}: \textbf{(I)} If the model generates more thinking tokens than a desired limit, we forcefully end the thinking process by appending an end-of-thinking token delimiter. Ending the thinking this way makes the model transition to generating its answer. \textbf{(II)} If we want the model to spend more test-time compute on a problem, we suppress the generation of the end-of-thinking token delimiter and instead append ``Wait'' to the model's current reasoning trace to encourage more exploration. Equipped with this simple recipe -- SFT on 1,000 samples and test-time budget forcing -- our model \model{} exhibits test-time scaling (\autoref{fig:scaling}). Further, \model{} is the most sample-efficient reasoning model and outperforms closed-source models like OpenAI's o1-preview (\autoref{fig:s1k-bar}).

We conduct extensive ablation experiments targeting (a) our selection of 1,000 (1K) reasoning samples and (b) our test-time scaling. For \textbf{(a)}, we find that jointly incorporating difficulty, diversity, and quality measures into our selection algorithm is important. Random selection, selecting samples with the longest reasoning traces, or only selecting maximally diverse samples all lead to significantly worse performance (around $-$30\% on AIME24 on average). Training on our full data pool of 59K examples, a superset of \data{}, does not offer substantial gains over our 1K selection. This highlights the importance of careful data selection and echoes prior findings for instruction tuning~\citep{zhou2023lima}. For \textbf{(b)}, we define desiderata for test-time scaling methods to compare different approaches. Budget forcing leads to the best scaling as it has perfect controllability with a clear positive slope leading to strong performance.

In summary, our contributions are: We develop simple methods for creating a sample-efficient reasoning dataset (\autoref{sec:data}) and test-time scaling (\autoref{sec:ttc});
Based on these we build \model{} which is competitive with o1-preview (\autoref{sec:results}); We ablate subtleties of data (\autoref{sec:dataabl}) and test-time scaling (\autoref{sec:scaleabl}). We end with a discussion to motivate future work on simple reasoning (\autoref{sec:disc}). Our code, model, and data are open-source at \url{https://github.com/simplescaling/s1}.

\section{Reasoning data curation to create \data{}}
\label{sec:data}

In this section, we describe our process for creating a large dataset first in \autoref{sec:59K} and then filtering it down to \data{} in \autoref{sec:selection-criteria}.

\subsection{Initial collection of 59K samples}
\label{sec:59K}

We collect an initial 59,029 questions from 16 sources following three guiding principles. \textbf{Quality}: Datasets should be high-quality; we always inspect samples and ignore datasets with, e.g., poor formatting; \textbf{Difficulty}: Datasets should be challenging and require significant reasoning effort; \textbf{Diversity}: Datasets should stem from various fields to cover different reasoning tasks. We collect datasets of two categories:

\paragraph{Curation of existing datasets} Our largest source is NuminaMATH \citep{numina_math_datasets} with 30,660 mathematical problems from online websites. We also include historical AIME problems (1983-2021). To enhance diversity, we add OlympicArena \citep{huang2024olympicarenabenchmarkingmultidisciplinecognitive} with 4,250 questions spanning Astronomy, Biology, Chemistry, Computer Science, Geography, Mathematics, and Physics from various Olympiads. OmniMath \citep{gao2024omnimathuniversalolympiadlevel} adds 4,238 competition-level mathematics problems. We also include 2,385 problems from AGIEval \citep{zhong2023agievalhumancentricbenchmarkevaluating}, which features questions from standardized tests like SAT and LSAT, covering English, Law, and Logic. We refer to \autoref{tab:ds} in \autoref{sec:details} for our other sources.

\paragraph{New datasets in quantitative reasoning} To complement these existing datasets, we create two original datasets. s1-prob consists of 182 questions from the probability section of Stanford University's Statistics Department's PhD Qualifying Exams (\url{https://statistics.stanford.edu}), accompanied by handwritten solutions that cover difficult proofs. The probability qualifying exam is held yearly and requires professional-level mathematical problem-solving. s1-teasers comprises 23 challenging brain-teasers commonly used in interview questions for quantitative trading positions. Each sample consists of a problem and solution taken from PuzzledQuant (\url{https://www.puzzledquant.com/}). We only take examples with the highest difficulty level ("Hard").

For each question, we generate a reasoning trace and solution using the Google Gemini Flash Thinking API~\citep{geminithinking} extracting its reasoning trace and response. This yields 59K triplets of a question, generated reasoning trace, and generated solution. Examples from our dataset are in \autoref{sec:samples}. We decontaminate all samples against our evaluation questions (MATH500, GPQA Diamond, AIME24; \autoref{sec:decontaminate}) using 8-grams and deduplicate the data.

\subsection{Final selection of 1K samples}
\label{sec:selection-criteria}

We could directly train on our pool of 59K questions, however, our goal is to find the \textit{simplest} approach with minimal resources. Thus, we go through three stages of filtering to arrive at a minimal set of 1,000 samples relying on our three guiding data principles: Quality, Difficulty, and Diversity. 

\paragraph{Quality} We first remove any questions where we ran into any API errors reducing our dataset to \defaultlightblue{54,116} samples. Next, we filter out low-quality examples by checking if they contain any string patterns with formatting issues, such as ASCII art diagrams, non-existent image references, or inconsistent question numbering reducing our dataset to \defaultlightblue{51,581} examples. From this pool, we identify \defaultlightblue{384} samples for our final 1,000 samples from datasets that we perceive as high-quality and not in need of further filtering (see \autoref{sec:algo} for details).

\paragraph{Difficulty} For difficulty, we use two indicators: model performance and reasoning trace length. We evaluate two models on each question: Qwen2.5-7B-Instruct and Qwen2.5-32B-Instruct~\citep{qwen2024qwen25technicalreport}, with correctness assessed by Claude 3.5 Sonnet comparing each attempt against the reference solution (see \autoref{sec:grading} for the grading protocol). We measure the token length of each reasoning trace to indicate problem difficulty using the Qwen2.5 tokenizer. This relies on the assumption that more difficult problems require more thinking tokens. Based on the grading, we remove questions that either Qwen2.5-7B-Instruct or Qwen2.5-32B-Instruct can solve correctly and thus may be too easy. By using two models we reduce the likelihood of an easy sample slipping through our filtering due to a rare mistake on an easy question of one of the models. This brings our total samples down to \defaultlightblue{24,496}, setting the stage for the next round of subsampling based on diversity. While filtering with these two models may be optimized for our setup as we will also use Qwen2.5-32B-Instruct as our model to finetune, the idea of model-based filtering generalizes to other setups.

\paragraph{Diversity} To quantify diversity, we classify questions into domains using Claude 3.5 Sonnet based on the Mathematics Subject Classification (MSC) system (e.g., geometry, combinatorics, etc.) from the American Mathematical Society.\footnote{\url{https://mathscinet.ams.org/mathscinet/msc/msc2020.html}} The taxonomy focuses on topics in mathematics but also includes other sciences such as biology, physics, and economics. To select our final examples from the pool of \defaultlightblue{24,496} questions, we first choose one domain uniformly at random. Then, we sample one problem from this domain according to a distribution that favors longer reasoning traces (see \autoref{sec:algo} for details) as motivated in \textit{Difficulty}. We repeat this process until we have \defaultlightblue{1,000} total samples spanning 50 domains.

In \autoref{sec:dataabl}, we will show that using our three criteria in combination is important, as only relying on quality, diversity, or difficulty in isolation leads to worse datasets. Some distilled generations are incorrect, which we allow in our data as we focus on capturing the reasoning process rather than entirely correct solutions. Our grader (\autoref{sec:grading}) deems 53.6\% correct in \data{} and 63.0\% in our follow-up \textbf{s1K-1.1} (see \autoref{sec:s11}).

\FloatBarrier

\section{Test-time scaling}
\label{sec:ttc}

\begin{figure}
\includegraphics[width=\columnwidth]{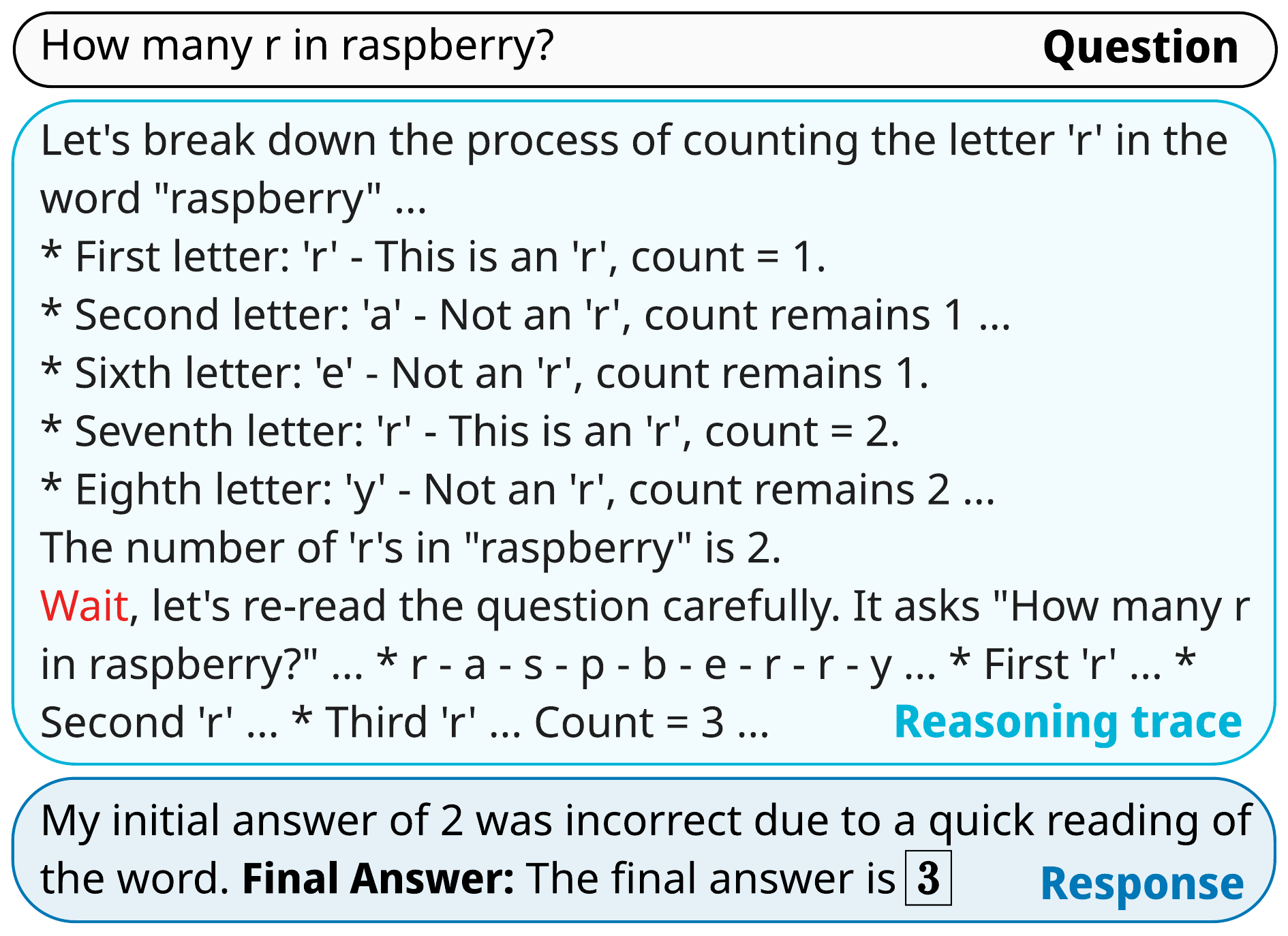}
\caption{\textbf{Budget forcing with \model{}.} The model tries to stop after ``...is 2.'', but we suppress the end-of-thinking token delimiter instead appending ``Wait'' leading \model{} to self-correct its answer.}
\label{fig:raspberry}
\end{figure}

\subsection{Method}

We classify test-time scaling methods into \textbf{1) Sequential}, where later computations depend on earlier ones (e.g., a long reasoning trace), and \textbf{2) Parallel}, where computations run independently (e.g., majority voting)~\citep{snell2024scalingllmtesttimecompute,brown2024largelanguagemonkeysscaling}. We focus on sequential scaling as intuitively we believe it should scale better, since later computations can build on intermediate results, allowing for deeper reasoning and iterative refinement. We propose new sequential scaling methods and ways to benchmark them.

\paragraph{Budget forcing} We propose a simple decoding-time intervention by forcing a maximum and/or minimum number of thinking tokens. Specifically, we enforce a maximum token count by simply appending the end-of-thinking token delimiter and optionally ``\texttt{Final Answer:}'' to early exit the thinking stage and make the model provide its current best answer. To enforce a minimum, we suppress the generation of the end-of-thinking token delimiter and optionally append the string ``Wait'' to the model's current reasoning trace to encourage the model to reflect on its current generation. \autoref{fig:raspberry} contains an example of how this simple approach can lead the model to arrive at a better answer.

\paragraph{Baselines} We benchmark budget forcing with: \textbf{(I) Conditional length-control methods}, which rely on telling the model in the prompt how long it should generate for. We group them by granularity into (a) Token-conditional control: We specify an upper bound of thinking tokens in the prompt; (b) Step-conditional control: We specify an upper bound of thinking steps, where each step is around 100 tokens; (c) Class-conditional control: We write two generic prompts that tell the model to either think for a short or long amount of time (see \autoref{sec:scalingdetails} for details). \textbf{(II) Rejection sampling}, which samples until a generation fits a predetermined compute budget. This oracle captures the posterior over responses conditioned on its length.

\subsection{Metrics}
\label{sec:metrics}

We establish a set of desiderata as evaluation metrics to measure test-time scaling across methods. Importantly, we do not only care about the accuracy a method can achieve but also its controllability and test-time scaling slope. For each method we consider, we run a set of evaluations $a \in \mathcal{A}$ varying test-time compute on a fixed benchmark, e.g. AIME24. This produces a piece-wise linear function $f$ with compute as the x-axis measured in thinking tokens and accuracy as the y-axis (see \autoref{fig:scaling}, where the rightmost dot for AIME24 corresponds to $f(7320)=57\%$). We measure three metrics:
\begin{align}
\text{Control} = \frac{1}{|\mathcal{A}|} \sum_{a \in \mathcal{A}} \mathbb{I}(a_{\text{min}} \leq a \leq a_{\text{max}})
\end{align}
where $a_{\text{min}}, a_{\text{max}}$ refer to a pre-specified minimum and maximum amount of test-time compute; in our case thinking tokens. We usually only constrain $a_{\text{max}}$. As tokens generated correspond to the amount of test-time compute spent, this metric measures the extent to which a method allows controllability over the use of that test-time compute. We report it as a percentage with 100\% being perfect control.
\begin{align}
\text{Scaling} = \frac{1}{\binom{|\mathcal{A}|}{2}} \sum_{\substack{a, b \in \mathcal{A} \\ b > a}} \frac{f(b) - f(a)}{b - a}
\end{align}
$\text{Scaling}$ is the average slope of the piece-wise linear function. It must be positive for useful methods and larger is better.
\begin{align}
\text{Performance} &= \max_{a \in \mathcal{A}} f(a)
\end{align}
Performance is simply the maximum performance the method achieves on the benchmark. A method with monotonically increasing scaling achieves 100\% performance on any benchmark in the limit. However, the methods we investigate eventually flatten out or further scaling fails due to control or context window limitations.

\begin{figure*}[t]
\centering
\subfigure[Sequential scaling via budget forcing\label{fig:forcing}]{\includegraphics[width=0.49\textwidth]{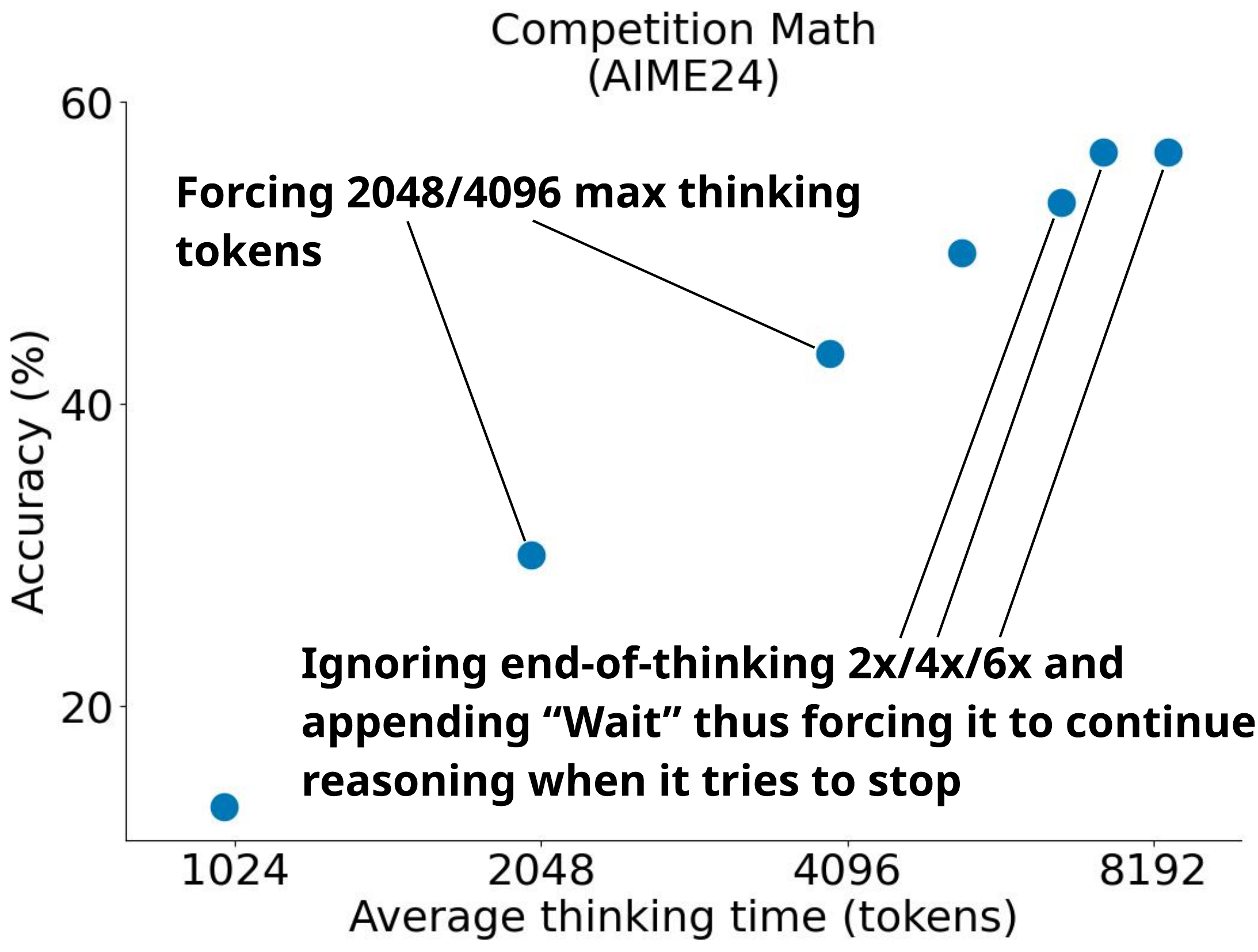}}
\subfigure[Parallel scaling via majority voting\label{fig:majority}]{\includegraphics[width=0.49\textwidth]{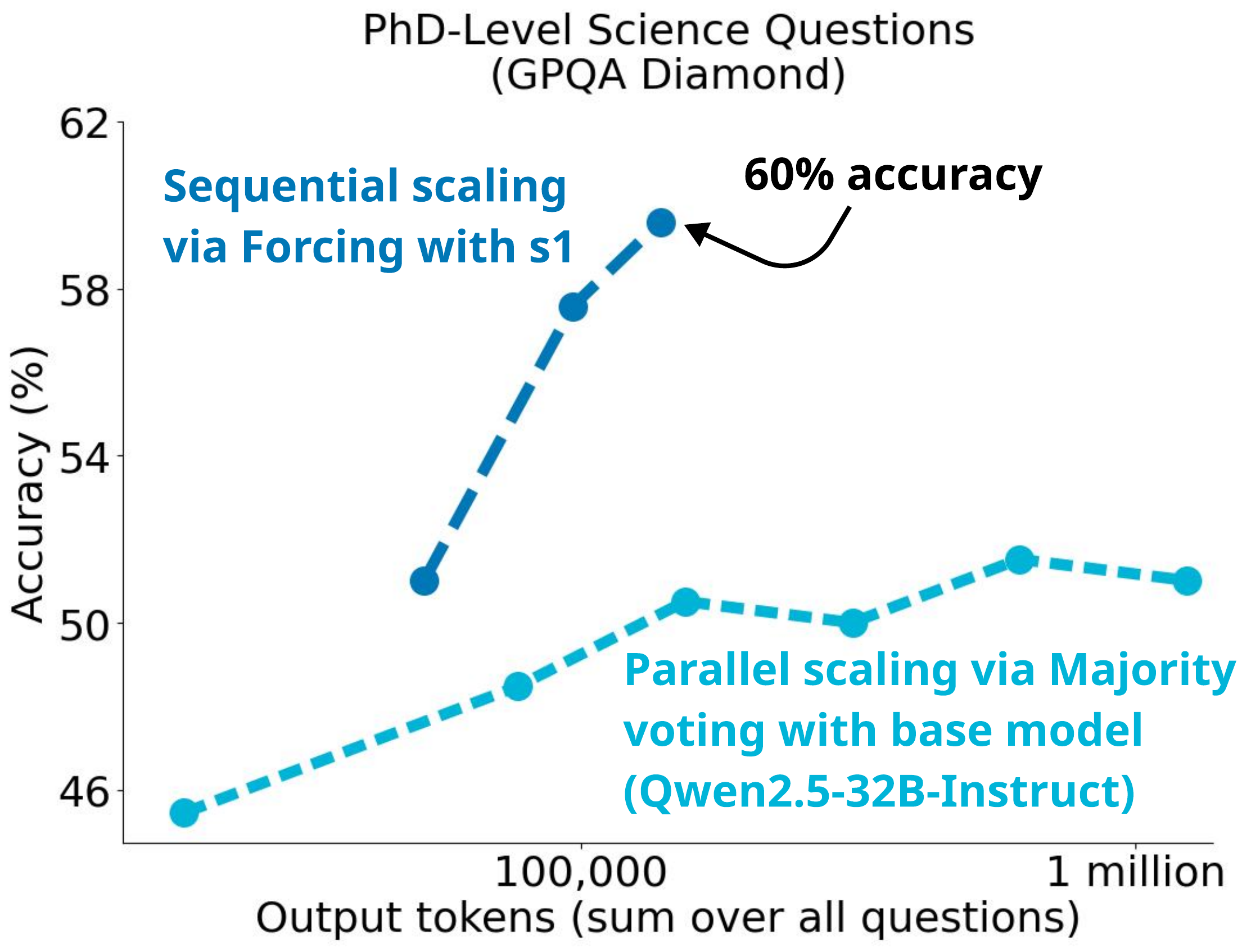}}
\caption{\textbf{Sequential and parallel test-time scaling.} \textit{(a):} Budget forcing shows clear scaling trends and extrapolates to some extent. For the three rightmost dots, we prevent the model from stopping its thinking 2/4/6 times, each time appending ``Wait'' to its current reasoning trace. \textit{(b):} For Qwen2.5-32B-Instruct we perform 64 evaluations for each sample with a temperature of 1 and visualize the performance when majority voting across 2, 4, 8, 16, 32, and 64 of these.}
\label{fig:scaling2}
\end{figure*}

\section{Results}
\label{sec:results}

\subsection{Setup}

\paragraph{Training} We perform supervised finetuning on Qwen2.5-32B-Instruct using \data{} to obtain our model \model{} using basic hyperparameters outlined in \autoref{sec:details-training}. Finetuning took 26 minutes on 16 NVIDIA H100 GPUs with PyTorch FSDP.

\paragraph{Evaluation} We select three representative reasoning benchmarks widely used in the field: \textbf{AIME24}~\citep{aime} has 30 problems that were used in the 2024 American Invitational Mathematics Examination (AIME) held from January 31 – February 1, 2024. AIME tests mathematical problem-solving with arithmetic, algebra, counting, geometry, number theory, probability, and other secondary school math topics. High-scoring high school students in the test are invited to participate in the United States of America Mathematics Olympiad (USAMO). All AIME answers are integers ranging from $000$ to $999$, inclusive. Some AIME problems rely on figures that we provide to our model using the vector graphics language Asymptote as it cannot take image inputs. \textbf{MATH500}~\citep{hendrycks2021measuringmathematicalproblemsolving} is a benchmark of competition math problems of varying difficulty. We evaluate on the same 500 samples selected by OpenAI in prior work~\citep{lightman2023letsverifystepstep}. \textbf{GPQA Diamond}~\citep{rein2023gpqagraduatelevelgoogleproofqa} consists of 198 PhD-level science questions from Biology, Chemistry and Physics. Experts with PhDs in the corresponding domains only achieved 69.7\% on GPQA Diamond~\citep{o1}. When we write ``GPQA'' in the context of evaluation in this work, we always refer to the Diamond subset. We build on the ``lm-evaluation-harness'' framework~\citep{eval-harness,biderman2024lessons}. Unless otherwise specified, we evaluate with a temperature of 0 (greedy) and measure accuracy (equivalent to pass@1).

\paragraph{Other models} We benchmark \model{} against: \textbf{OpenAI o1 series}~\citep{o1}, closed-source models that popularized test-time scaling; \textbf{DeepSeek r1 series}~\citep{r1}, open-weight reasoning models with up to o1-level performance; Qwen's \textbf{QwQ-32B-preview}~\citep{qwq-32b-preview}, a 32B open-weight reasoning model without disclosed methodology; \textbf{Sky-T1-32B-Preview}~\citep{sky_t1} and \textbf{Bespoke-32B}~\citep{bespoke_stratos}, open models with open reasoning data distilled from QwQ-32B-preview and r1; \textbf{Google Gemini 2.0 Flash Thinking Experimental}~\citep{geminithinking}, the API that we distill from. As it has no official evaluation scores, we use the Gemini API to benchmark it ourselves. However, the ``recitation error'' of the Gemini API makes evaluation challenging.\footnote{\url{https://github.com/google/generative-ai-docs/issues/257}} We circumvent this, by manually inserting all 30 AIME24 questions in its web interface where the error does not appear. However, we leave out MATH500 (500 questions) and GPQA Diamond (198 questions), thus they are N.A. in \autoref{tab:perf}. Our model, \model{}, is fully open including weights, reasoning data, and code.

\begin{table}[htbp]
\centering
\caption{\textbf{\model{} is a strong open reasoning model.} We evaluate \model{}, Qwen, and Gemini (some entries are unknown (N.A.), see \autoref{sec:results}). Other results are from the respective reports~\citep{qwen2024qwen25technicalreport,qwq-32b-preview,o1,r1,bespoke_stratos,sky_t1}. \# ex. = number examples used for reasoning finetuning; BF = budget forcing. See \autoref{sec:s11} for our better \textbf{s1.1} model.}
\begin{tabular}{lrrrr}
\toprule
Model & \makecell{\# ex.} & \makecell{AIME\\2024} & \makecell{MATH\\500} & \makecell{GPQA\\Diamond} \\
\midrule
\multicolumn{5}{c}{\textbf{API only}} \\
\midrule
o1-preview & N.A. & 44.6 & 85.5 & 73.3 \\
o1-mini & N.A. & 70.0 & 90.0 & 60.0 \\
o1 & N.A. & \textbf{74.4} & \textbf{94.8} & \textbf{77.3} \\
Gemini 2.0 & \multirow{2}{*}{N.A.} & \multirow{2}{*}{60.0} & \multirow{2}{*}{N.A.} & \multirow{2}{*}{N.A.} \\
Flash Think. & & & & \\
\midrule
\multicolumn{5}{c}{\textbf{Open Weights}} \\
\midrule
Qwen2.5- & \multirow{2}{*}{N.A.} & \multirow{2}{*}{26.7} & \multirow{2}{*}{84.0} & \multirow{2}{*}{49.0} \\
32B-Instruct & & & & \\
QwQ-32B & N.A. & 50.0 & 90.6 & 54.5 \\
r1 & $\gg$800K & \textbf{79.8} & \textbf{97.3} & \textbf{71.5} \\
r1-distill & 800K & 72.6 & 94.3 & 62.1 \\
\midrule
\multicolumn{5}{c}{\textbf{Open Weights and Open Data}} \\
\midrule
Sky-T1 & 17K & 43.3 & 82.4 & 56.8 \\
Bespoke-32B & 17K & \textbf{63.3} & 93.0 & 58.1 \\
\midrule
s1 w/o BF & \textbf{1K} & 50.0 & 92.6 & 56.6 \\
\textbf{s1-32B} & \textbf{1K} & 56.7 & \textbf{93.0} & \textbf{59.6} \\
\bottomrule
\label{tab:perf}
\end{tabular}
\end{table}

\subsection{Performance}

\paragraph{Test-time scaling} \autoref{fig:scaling} shows the performance of \model{} with budget forcing scales with more test-time compute. In \autoref{fig:scaling2} (left), we expand the plot from \autoref{fig:scaling} (middle) showing that while we can improve AIME24 performance using our budget forcing technique (\autoref{sec:ttc}) and more test-time compute it does eventually flatten out at six times. Suppressing the end-of-thinking token delimiter too often can lead the model into repetitive loops instead of continued reasoning. In \autoref{fig:scaling2} (right), we show that after training Qwen2.5-32B-Instruct on our 1,000 samples to produce \model{} and equipping it with the simple budget forcing technique, it operates in a different scaling paradigm. Scaling test-time compute on the base model via majority voting cannot catch up with the performance of \model{} which validates our intuition from \autoref{sec:ttc} that sequential scaling is more effective than parallel. We provide example generations of \model{} in \autoref{fig:ex}.

\paragraph{Sample-efficiency} In \autoref{fig:s1k-bar} (right) and \autoref{tab:perf} we compare \model{} with other models. We find that \model{} is the most sample-efficient open data reasoning model. It performs significantly better than our base model (Qwen2.5-32B-Instruct) despite just training it on an additional 1,000 samples. The concurrently released r1-32B shows stronger performance than \model{} while also only using SFT~\citep{r1}. However, it is trained on 800 $\times$ more reasoning samples. It is an open question whether one can achieve their performance with just 1,000 samples. Finally, our model nearly matches Gemini 2.0 Thinking on AIME24. As the data for \model{} is distilled from Gemini 2.0, this shows our distillation procedure was likely effective. 

\begin{figure*}[p]
\footnotesize
\centering
\begin{tabular}{@{}p{0.39\textwidth}p{0.27\textwidth}p{0.28\textwidth}@{}}
\textbf{[AIME24]}

~

\textbf{Alice and Bob play the following game. A stack of $n$ tokens lies before them. The players take turns with Alice going first. On each turn, the player removes either $1$ token or $4$ tokens from the stack. Whoever removes the last token wins. Find the number of positive integers $n$ less than or equal to $2024$ for which there exists a strategy for Bob that guarantees that Bob will win the game regardless of Alice's play.
}

~ 

\textcolor{defaultlightblue}{
The problem asks for the number of positive integers $n \le 2024$ such that Bob has a winning strategy. This is a typical impartial game that can be analyzed using the Sprague-Grundy theorem or by directly finding the winning and losing positions. Let $W$ be the set of winning positions (number of tokens $n$ such that the player whose turn it is can force a win) and $L$ be the set of losing positions. \textcolor[HTML]{808080}{[...]}
}

~

\textcolor{defaultlightblue}{
Let $P(n)$ be the statement "the player whose turn it is when there are $n$ tokens has a winning strategy. $P(0)$ is false (the player whose turn it is and faces $0$ tokens has already lost).
Consider the small values of $n$:}

\textcolor{defaultlightblue}{
$n=1$: The player whose turn it is can take $1$ token and win. So $P(1)$ is true. $1 \in W$.}

\textcolor{defaultlightblue}{
$n=2$: The player whose turn it is can take $1$ token, leaving $1$ token for the opponent. Since $1 \in W$, the opponent wins. So $P(2)$ is false. $2 \in L$.}

\textcolor{defaultlightblue}{
$n=3$: The player whose turn it is can take $1$ token, leaving $2$ tokens for the opponent. Since $2 \in L$, the player whose turn it is wins. So $P(3)$ is true. $3 \in W$.\textcolor[HTML]{808080}{[...]}}


~

\textcolor{defaultlightblue}{
We are looking for the number of positive integers $n \le 2024$ for which Bob has a winning strategy. This means that if the game starts with $n$ tokens, and Alice goes first, Bob can win. This happens if and only if $n$ is a losing position (because if $n \in L$, then Alice's first move must lead to a winning position for Bob). \textcolor[HTML]{808080}{[...]}
}

\textcolor{defaultlightblue}{
Let's try to find a pattern in the losing positions $L = \{0, 2, 5, 7, 10, 12, 15, 17, 20, 22, ...\}$.
Differences between consecutive terms: $2, 3, 2, 3, 2, 3, 2, 3, 2, ...$
The pattern of differences is alternating $2, 3$. \textcolor[HTML]{808080}{[...]}
}

\textcolor{defaultlightblue}{
The losing positions are of the form $5m$ or $5m+2$, where $m \ge 0$.
We need to find the number of positive integers $n \le 2024$ that are in the set of losing positions $L$. \textcolor[HTML]{808080}{[...]}
}

\textcolor{defaultlightblue}{
Case 1: $n = 5m$, where $m \ge 1$.
$1 \le 5m \le 2024$
$1/5 \le m \le 2024/5 = 404.8$
Since $m$ is an integer, $1 \le m \le 404$.
The number of values of $m$ is $404$. 
}

\textcolor{defaultlightblue}{
Case 2: $n = 5m+2$, where $m \ge 0$.
$5m+2 \le 2024$
$5m \le 2022$
$m \le 2022/5 = 404.4$
Since $m$ is an integer, $0 \le m \le 404$.
The number of values of $m$ is $404 - 0 + 1 = 405$.
}

~

\textcolor{defaultblue}{
The total number of positive integers $n \le 2024$ that are losing positions is $404 + 405 = \boxed{809}$.
}

&
\textbf{[MATH500]}

~

\textbf{
Find the projection of $a$ onto $b = \begin{pmatrix} 2 \\ 6 \\ 3 \end{pmatrix}$ if $a \cdot b = 8.$
}

~ 

\textcolor{defaultlightblue}{
The projection of vector $\mathbf{a}$ onto vector $\mathbf{b}$ is given by the formula:
$$\text{proj}_{\mathbf{b}} \mathbf{a} = \frac{\mathbf{a} \cdot \mathbf{b}}{\|\mathbf{b}\|^2} \mathbf{b}$$
We are given $\mathbf{b} = \begin{pmatrix} 2 \\ 6 \\ 3 \end{pmatrix}$ and $\mathbf{a} \cdot \mathbf{b} = 8.$
}

~

\textcolor{defaultlightblue}{
First, we need to find the magnitude squared of $\mathbf{b}$, which is $\|\mathbf{b}\|^2$.
$$\|\mathbf{b}\|^2 = \mathbf{b} \cdot \mathbf{b} = 2^2 + 6^2 + 3^2 = 4 + 36 + 9 = 49.$$
}

~

\textcolor{defaultlightblue}{
Now we can plug the values into the projection formula:
$$\text{proj}_{\mathbf{b}} \mathbf{a} = \frac{8}{49} \mathbf{b}$$
Substitute the vector $\mathbf{b}$:
$$\text{proj}_{\mathbf{b}} \mathbf{a} = \frac{8}{49} \begin{pmatrix} 2 \\ 6 \\ 3 \end{pmatrix}$$}

~

\textcolor{defaultlightblue}{
Multiply the scalar by each component of the vector:
$$\text{proj}_{\mathbf{b}} \mathbf{a} = \begin{pmatrix} \frac{8}{49} \times 2 \\ \frac{8}{49} \times 6 \\ \frac{8}{49} \times 3 \end{pmatrix} = \begin{pmatrix} \frac{16}{49} \\ \frac{48}{49} \\ \frac{24}{49} \end{pmatrix}$$
}

~

\textcolor{defaultblue}{
Final Answer: The final answer is 
$$\boxed{\begin{pmatrix} \frac{16}{49} \\ \frac{48}{49} \\ \frac{24}{49} \end{pmatrix}}$$
}

~

&
\textbf{[GPQA]}

~

\textbf{
A quantum mechanical particle of mass $m$ moves in two dimensions in the following potential, as a function of the polar coordinates $(r, \theta)$:
$$V(r, \theta) = \frac{1}{2} k r^2 + \frac{3}{2} k r^2 \cos^2(\theta)$$
Find the energy spectrum.
\textbf{Hint:} Write the potential in Cartesian coordinates.
}

~

\textcolor{defaultlightblue}{
The problem asks for the energy spectrum of a particle in a given 2D potential. The potential is given in polar coordinates $(r, \theta)$:
$$V(r, \theta) = \frac{1}{2} kr^2 + \frac{3}{2} kr^2 \cos^2(\theta).$$
We can rewrite the potential as:
$$V(r, \theta) = \frac{1}{2} kr^2 (1 + 3\cos^2(\theta)).$$
}
\textcolor{defaultlightblue}{
The Hamiltonian for a particle in 2D is given by:
$$H = -\frac{\hbar^2}{2m} \left( \frac{\partial^2}{\partial x^2} + \frac{\partial^2}{\partial y^2} \right) + V(x, y).$$
In polar coordinates $(r, \theta)$, the Laplacian operator is:
$$\nabla^2 = \frac{1}{r} \frac{\partial}{\partial r} \left(r \frac{\partial}{\partial r} \right) + \frac{1}{r^2} \frac{\partial^2}{\partial \theta^2}.$$
So the Hamiltonian in polar coordinates is:
$$H = -\frac{\hbar^2}{2m} \left[ \frac{1}{r} \frac{\partial}{\partial r} \left(r \frac{\partial}{\partial r} \right) + \frac{1}{r^2} \frac{\partial^2}{\partial \theta^2} \right] + V(r, \theta).$$
}
\textcolor{defaultlightblue}{
The time-independent Schrödinger equation is $H\psi = E\psi$.
We can try to separate variables by assuming a solution of the form $\psi(r, \theta) = R(r)\Theta(\theta)$. \textcolor[HTML]{808080}{[...]}
}

~

\textcolor{defaultblue}{
This confirms the energy spectrum is:
$$E = \hbar\sqrt{\frac{k}{m}} \left(2n_x + n_y + \frac{3}{2}\right).$$
Final Answer: $E = \hbar\sqrt{\frac{k}{m}} \left(2n_x + n_y + \frac{3}{2}\right)$.
}
\\
\bottomrule
\end{tabular}
\vspace{-5pt}
\caption{\textbf{Example model outputs.} We pick a question from AIME24 (\textit{left}), MATH500 (\textit{middle}), and GPQA (\textit{right}), where our model generates the correct answer. The black text is the prompt, the \textcolor{defaultlightblue}{light blue} text is the reasoning trace, and the \textcolor{defaultblue}{blue} text is the answer of \model{}. The gray ellipsis \textcolor[HTML]{808080}{[...]} indicates that the text was trimmed to fit this page, but the generated text is actually longer.}
\label{fig:ex}
\end{figure*}

\section{Ablations}
\label{sec:abl}

\subsection{Data Quantity, Diversity, and Difficulty}
\label{sec:dataabl}

\begin{table}[htbp]
\caption{\textbf{\data{} data ablations.} We budget force (BF) a maximum of around 30,000 thinking tokens for all scores in this table. This performs slightly better than the scores without BF (\autoref{tab:perf}) as it allows the model to finish with a best guess when stuck in an infinite loop. We report 95\%  paired bootstrap confidence intervals for differences relative to the \data{} model using 10,000 bootstrap samples. E.g., the interval [-13\%, 20\%] means that, with 95\% confidence, the true difference between 59K-full and \data{} is between -13\% and +20\%. If the entire interval is negative, e.g. [-27\%, -3\%], we can confidently say that the performance is worse than \data{}.}
\begin{tabular}{l|ccc}
\toprule 
Model & \makecell{AIME\\2024} & \makecell{MATH\\500} & \makecell{GPQA\\Diamond} \\ 
\midrule
\multirow{2}{*}{1K-random} & 36.7 & 90.6 & 52.0 \\
& \scriptsize{[-26.7\%, -3.3\%]} & \scriptsize{[-4.8\%, 0.0\%]} & \scriptsize{[-12.6\%, 2.5\%]} \\
\multirow{2}{*}{1K-diverse} & 26.7 & 91.2 & 54.6 \\
& \scriptsize{[-40.0\%, -10.0\%]} & \scriptsize{[-4.0\%, 0.2\%]} &  \scriptsize{[-10.1\%, 5.1\%]}\\
\multirow{2}{*}{1K-longest} & 33.3 & 90.4 & 59.6 \\
& \scriptsize{[-36.7\%, 0.0\%]} & \scriptsize{[-5.0\%, -0.2\%]} & \scriptsize{[-5.1\%, 10.1\%]} \\
\multirow{2}{*}{59K-full} & 53.3 & 92.8 & 58.1 \\
& \scriptsize{[-13.3\%, 20.0\%]} & \scriptsize{[-2.6\%, 2.2\%]} & \scriptsize{[-6.6\%, 8.6\%]} \\
\midrule
\data{} & 50.0 & 93.0 & 57.6 \\
\bottomrule
\end{tabular}
\label{tab:datablation}
\end{table}

In \autoref{sec:data} we outlined our three guiding principles in curating \data{}: Quality, Difficulty, and Diversity. Here we test the importance of combining them and the overall efficacy of our selection. \textbf{Only Quality (1K-random)}: After obtaining our high-quality reasoning chains from Gemini, we select 1,000 samples at random; not relying on our difficulty and diversity filtering at all. \autoref{tab:datablation} shows this approach performs much worse than \data{} across all benchmarks. \textbf{Only Diversity (1K-diverse)}: For this dataset, we sample uniformly across domains to maximize diversity disregarding any notion of difficulty. This approach also leads to poor performance similar to 1K-random. \textbf{Only Difficulty (1K-longest)}: Here we rely on one of our difficulty indicators introduced in \autoref{sec:data} by selecting the 1,000 samples with the longest reasoning traces. This approach significantly boosts GPQA performance but overall still falls short of using \data{}. \textbf{Maximize Quantity}: Finally, we compare with just training on all of our 59K samples, a superset of all the 1K-sample versions. This leads to a strong model but uses much more resources. To finetune on 59K samples, we use 394 H100 GPU hours while \model{} only required 7 H100 GPU hours.  Moreover, relying only on \data{} is extremely competitive as shown in \autoref{sec:data}. Overall, combining all three criteria -- \textit{Quality}, \textit{Difficulty}, \textit{Diversity} -- via our methodology in \autoref{sec:data} is key for sample-efficient reasoning training.

\subsection{Test-time scaling methods}
\label{sec:scaleabl}

\begin{table}[htbp]
\centering
\caption{\textbf{Ablations on methods to scale test-time compute on AIME24.} $|\mathcal{A}|$ refers to the number of evaluation runs used to estimate the properties; thus a higher value indicates more robustness. \textbf{Bold} indicates our chosen method and the best values. BF = budget forcing, TCC/SCC/CCC = token/step/class-conditional control, RS = rejection sampling.}
\begin{tabular}{l|cccc}
\toprule
Method & Control & Scaling & Performance & $|\mathcal{A}|$ \\
\midrule
\textbf{BF} & \textbf{100\%} & 15 & \textbf{56.7} & 5 \\
\midrule
TCC & 40\% & -24 & 40.0 & 5 \\
TCC + BF & \textbf{100\%} & 13 & 40.0 & 5 \\
SCC & 60\% & 3 & 36.7 & 5 \\
SCC + BF & \textbf{100\%} & 6 & 36.7 & 5 \\
CCC & 50\% & \textbf{25} & 36.7 & 2 \\
\midrule
RS & \textbf{100\%} & -35 & 40.0 & 5 \\
\bottomrule
\end{tabular}
\label{tab:scalingabl}
\end{table}

\paragraph{Budget forcing} In \autoref{tab:scalingabl} we compare the test-time scaling methods we have introduced in \autoref{sec:ttc}. Overall, we find that \textit{budget forcing} provides perfect control, good scaling, and leads to our best AIME24 score. Thus, this is the method we use for \model{} in \autoref{fig:scaling} and in \autoref{sec:results}. In \autoref{tab:forcingablation}, we compare different strings for extrapolating performance. We find that ``Wait'' generally gives the best performance.

\textbf{Class-conditional control} We provide benchmark scores for this method in \autoref{sec:scalingdetails} and summarize three findings here: \textbf{(1)}~Token-conditional control fails without budget forcing, as our model cannot reliably count tokens - even when trained to do so. \textbf{(2)}~Under step-conditional control, the model generates a similar total number of tokens when given different step targets, as the model goes from few steps with many tokens per step, to many steps with few tokens in each step. Thus, the model learns to hack its way around the compute constraint making the controllability of this method mediocre. \textbf{(3)}~Class-conditional control can work - telling a model to simply think longer can increase its test-time compute and performance, which leads good scaling in \autoref{tab:scalingabl}.

\begin{table}[htbp]
\centering
\caption{\textbf{Budget forcing extrapolation ablations.} We compare ignoring the end-of-thinking delimiter twice and appending none or various strings.}
\begin{tabular}{l|ccc}
\toprule
Model & \makecell{AIME\\2024} & \makecell{MATH\\500} & \makecell{GPQA\\Diamond} \\
\midrule
No extrapolation & 50.0 & \textbf{93.0} & 57.6 \\
\midrule
2x without string & 50.0 & 90.2 & 55.1 \\
2x ``Alternatively'' & 50.0 & 92.2 & \textbf{59.6} \\
2x ``Hmm'' & 50.0 & \textbf{93.0} & \textbf{59.6} \\
2x ``Wait'' & \textbf{53.3} & \textbf{93.0} & \textbf{59.6} \\
\bottomrule
\end{tabular}
\label{tab:forcingablation}
\end{table}

\FloatBarrier

\begin{figure}
\includegraphics[width=\columnwidth]{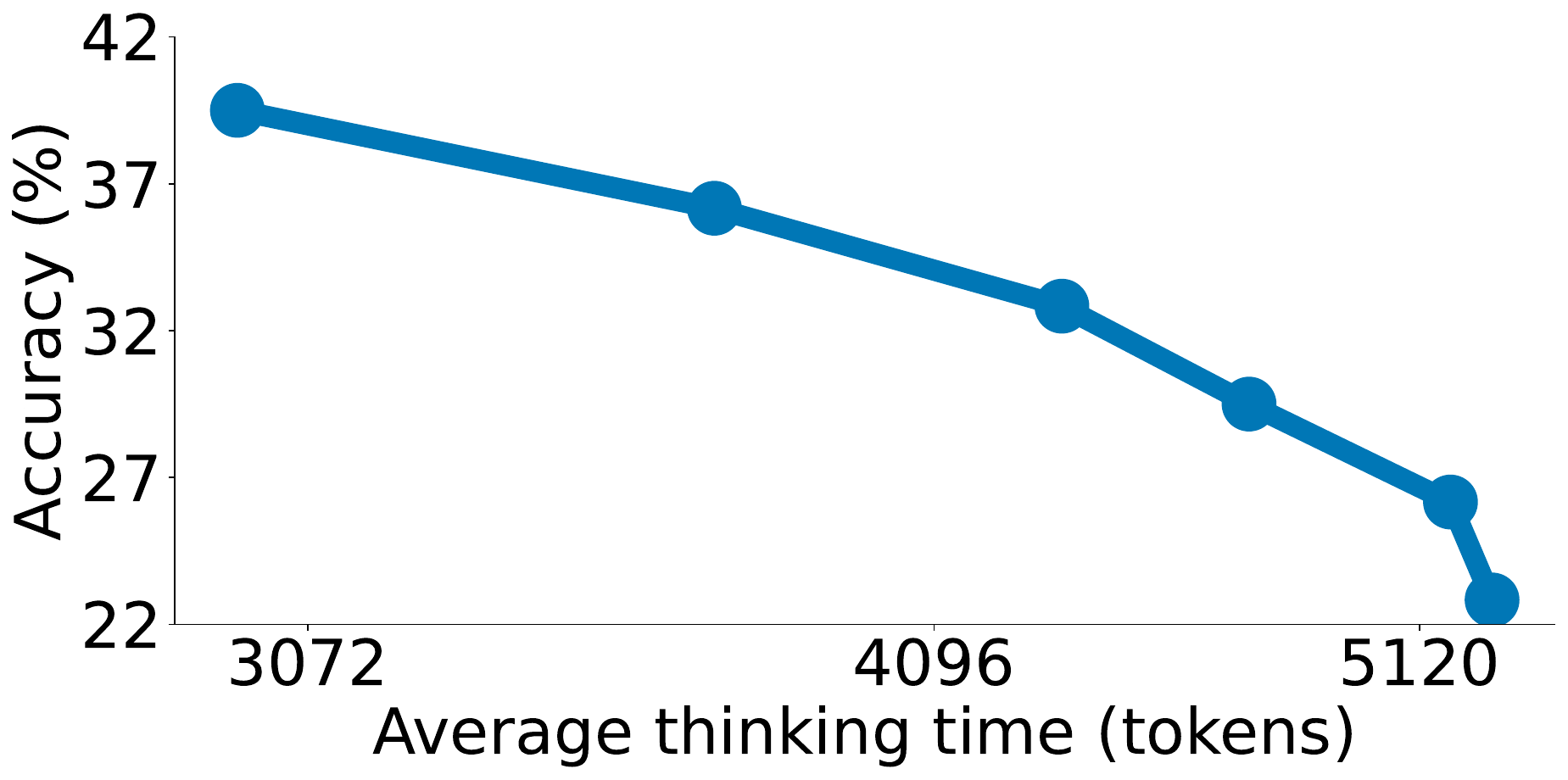}
\caption{\textbf{Rejection sampling on AIME24 with \model{}.} We sample with a temperature of 1 until all generations have less than (from left to right) 3500, 4000, 5000, 8000, and 16000 thinking tokens requiring an average of 655, 97, 8, 3, 2, and 1 tries per sample.}
\label{fig:rejection}
\end{figure}

\paragraph{Rejection sampling} Surprisingly, we find that simply sampling until the generation fits a specific length leads to an inverse scaling trend as depicted in \autoref{fig:rejection}. In \autoref{sec:samplerej} we inspect a question, which was answered correctly by the model when rejection sampling for $\leq4000$, but not for the $\leq8000$ token setting. In the $\leq4000$ setting the model directly jumps to the correct approach, while for the $\leq8000$ setting it backtracks a lot. We hypothesize that there is a correlation such that shorter generations tend to be the ones where the model was on the right track from the start, whereas longer ones tend to be ones where the model made mistakes and thus backtracks or questions itself. This leads to longer samples often being wrong when rejection sampling and thus the inverse scaling trend.

\section{Discussion and related work}
\label{sec:disc}

\subsection{Sample-efficient reasoning}

\paragraph{Models} There are a number of concurrent efforts to build models that replicate the performance of o1~\citep{o1}. For example, DeepSeek-r1 and k1.5~\citep{r1, k1.5} are built with reinforcement learning methods, while others rely on SFT using tens of thousands of distilled examples~\citep{sky_t1,xu2025redstardoesscalinglongcot, bespoke_stratos}. We show that SFT on only 1,000 examples suffices to build a competitive reasoning model matching o1-preview and produces a model that lies on the pareto frontier (\autoref{fig:s1k-bar}). Further, we introduce budget forcing which combined with our reasoning model leads to the first reproduction of OpenAI's test-time scaling curves~\citep{o1}. Why does supervised finetuning on just 1,000 samples lead to such performance gains? We hypothesize that the model is already exposed to large amounts of reasoning data during pretraining which spans trillions of tokens. Thus, the ability to perform reasoning is already present in our model. Our sample-efficient finetuning stage just activates it and we scale it further at test time with budget forcing. This is similar to the "Superficial Alignment Hypothesis" presented in LIMA~\citep{zhou2023lima}, where the authors find that 1,000 examples can be sufficient to align a model to adhere to user preferences.

\paragraph{Benchmarks and methods} To evaluate and push the limits of these models, increasingly challenging benchmarks have been introduced, such as Olympiad-level science competitions~\cite{he2024olympiadbenchchallengingbenchmarkpromoting,jain2024livecodebenchholisticcontaminationfree,zhong2023agievalhumancentricbenchmarkevaluating} and others~\citep{srivastava2023imitation,glazer2024frontiermathbenchmarkevaluatingadvanced,su2024brightrealisticchallengingbenchmark,kim2024llmasaninterviewerstatictestingdynamic,phan2025humanity}. To enhance models’ performance on reasoning-related tasks, researchers have pursued several strategies: Prior works have explored continuing training language models on specialized corpora related to mathematics and science~\citep{azerbayev2023llemma, yang2024syntheticcontinuedpretraining}, sometimes even synthetically generated data~\citep{yu2023metamath}. Others have developed training methodologies specifically aimed at reasoning performance~\citep{zelikman2022starbootstrappingreasoningreasoning,zelikman2024quietstarlanguagemodelsteach,luo2025wizardmathempoweringmathematicalreasoning,yuan2025agentrtraininglanguagemodel,wu2024thinkingllmsgeneralinstruction}. Another significant line of work focuses on prompting-based methods to elicit and improve reasoning abilities, including methods like Chain-of-Thought prompting \citep{wei2023chainofthoughtpromptingelicitsreasoning,yao2024tree,yao2023reactsynergizingreasoningacting,bi2024program,fu2022complexity, zhang2023cumulative,xiang20252reasoningllmslearning,hu2024visual, diao2024activepromptingchainofthoughtlarge}. These combined efforts aim to advance the reasoning ability of language models, enabling them to handle more complex and abstract tasks effectively.

\subsection{Test-time scaling}

\paragraph{Methods} As we introduce in \autoref{sec:ttc}, we differentiate two methods to scale test-time compute: \textbf{parallel} and \textbf{sequential}. The former relies on multiple solution attempts generated in parallel and selecting the best outcome via specific criteria. These criteria include choosing the most frequent response for majority voting or the best response based on an external reward for Best-of-N~\citep{brown2024largelanguagemonkeysscaling, irvine2023rewardingchatbotsrealworldengagement, levi2024simplemodelinferencescaling}. Unlike repeated sampling, previous sequential scaling methods let the model generate solution attempts sequentially based on previous attempts, allowing it to refine each attempt based on previous outcomes \citep{snell2024scalingllmtesttimecompute,hou2025advancinglanguagemodelreasoning,lee2025evolvingdeeperllmthinking}. Tree-based search methods~\citep{gandhi2024streamsearchsoslearning, wu2024inference} offer a hybrid approach between sequential and parallel scaling, such as Monte-Carlo Tree Search (MCTS)~\citep{liu2024dontthrowawayvalue, zhang2023planninglargelanguagemodels, zhou2024languageagenttreesearch, choi2023kcts} and guided beam search \citep{xie2024self}. \textsc{REBASE}~\citep{wu2024inference} employs a process reward model to balance exploitation and pruning during tree search. Empirically, \textsc{REBASE} has been shown to outperform sampling-based methods and MCTS~\citep{wu2024inference}. Reward models~\cite{lightman2023letsverifystepstep, wang-etal-2024-math,wang2024helpsteer2opensourcedatasettraining} play a key role in these methods. They come in two variants: outcome reward models and process reward models. Outcome reward models \cite{xin2024deepseekproveradvancingtheoremproving, ankner2024critiqueoutloudrewardmodels} assign a score to complete solutions and are particularly useful in Best-of-N selection, while process reward models \citep{lightman2023letsverifystepstep, wang-etal-2024-math, wu2024inference} assess individual reasoning steps and are effective in guiding tree-based search methods. 

\paragraph{Limits to further test-time scaling} We have shown that budget forcing allows extrapolating test-time compute in \autoref{sec:results}, e.g., improving AIME24 performance from 50\% to 57\%. However, it has two key limitations when scaling further: it eventually \textbf{flattens out} (\autoref{fig:scaling2}), and the \textbf{context window} of the underlying language model constrains it. Despite these constraints, our work shows test-time scaling across a wide range of accuracies (\autoref{fig:scaling}), partly because scaling down test-time compute behaves predictably and does not suffer from these constraints. 

Continuing test-time scaling will require approaches that can further extrapolate test-time compute. How can we get such extrapolation? There may be improvements to budget forcing such as rotating through different strings, not only ``Wait'', or combining it with frequency penalties or higher temperature to avoid repetitive loops. An exciting direction for future work is also researching whether applying budget forcing to a reasoning model trained with reinforcement learning yields better extrapolation; or if RL allows for new ways of test-time scaling beyond budget forcing. Our work defines the right metrics (\autoref{sec:metrics}) -- Control, Scaling, and Performance -- to enable future research and progress on extrapolating test-time compute.

\begin{figure}
\includegraphics[width=\columnwidth]{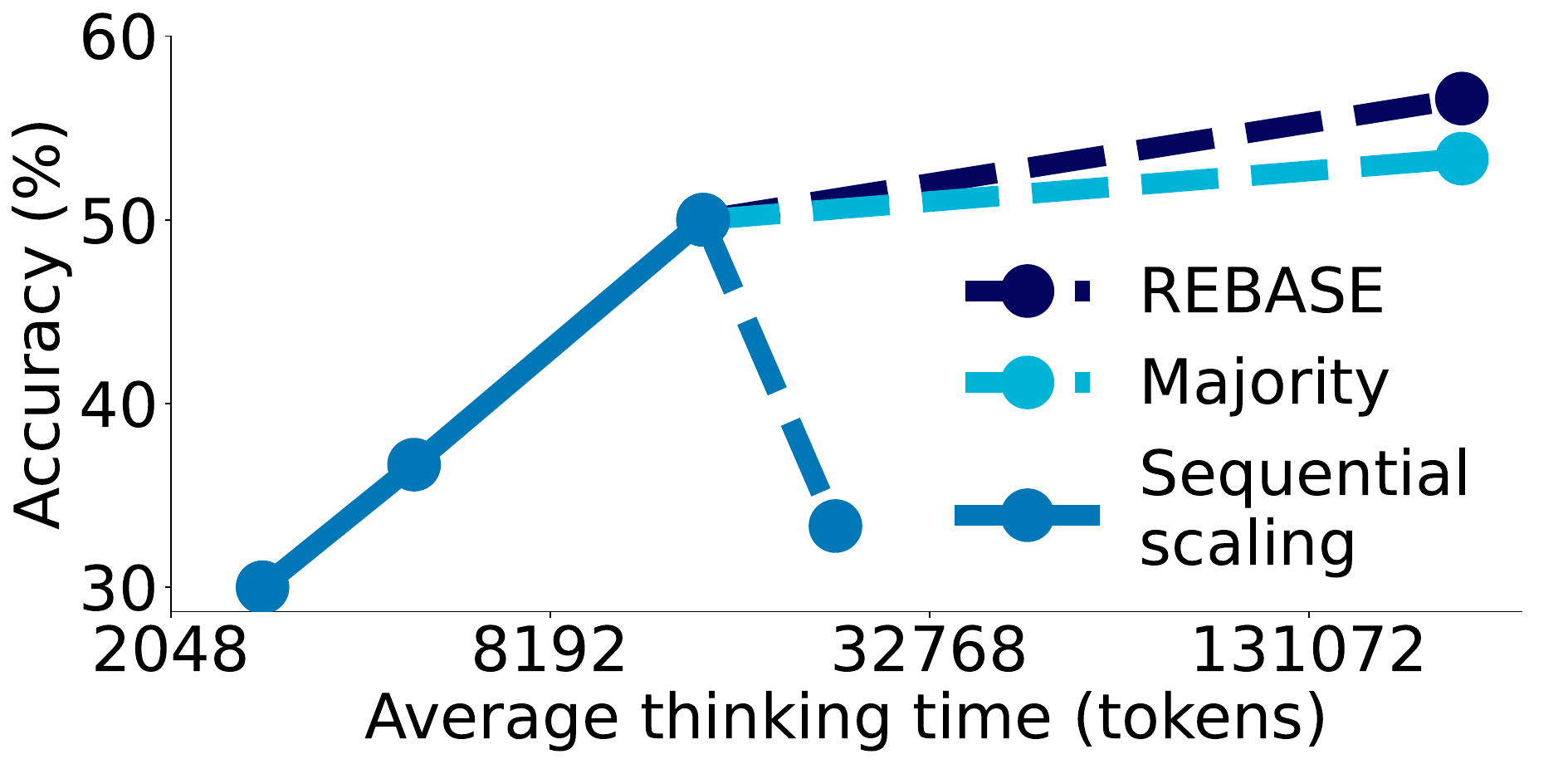}
\caption{\textbf{Scaling further with parallel scaling methods.}  All metrics averaged over the 30 questions in AIME24. Average thinking tokens for \textsc{REBASE} do not account for the additional compute from the reward model. For sequential scaling, we prompt the model to use up to (from left to right) 32, 64, 256, and 512 steps. For \textsc{REBASE} and majority voting we generate 16 parallel trajectories to aggregate across.}
\label{fig:parallel}
\end{figure}

\paragraph{Parallel scaling as a solution} Parallel scaling offers one solution to the limits of sequential scaling, thus we augment our sequentially scaled model with two methods: \textbf{(I) Majority voting:} After generating $k$ solutions, the final solution is the most frequent one across generations; \textbf{(II) Tree search via \textsc{REBASE}:} We use the \textsc{REBASE} process reward model, which is initialized from LLaMA-34B and further finetuned on a synthetic process reward modeling dataset~\citep{wu2024inference}. We then aggregate the solutions generated by \textsc{REBASE} via majority voting. As shown in \autoref{fig:parallel}, augmenting our model with \textsc{REBASE} scales better than majority voting, and even sequential scaling in this scenario. However, \textsc{REBASE} requires an additional forward pass at each step for the reward model adding some computation overhead. For sequential scaling, when prompted to use up to 512 steps, for 12 out of the 30 evaluation questions the model generates a response that exceeds the context window leading to a large performance drop. Overall, we find that these parallel scaling methods complement sequential scaling thus they offer an avenue for scaling test-time compute even further; beyond fixed context windows.

\nocite{cesista2024multimodalstructuredgenerationcvprs}

\section*{Impact Statement}

Language models with strong reasoning capabilities have the potential to greatly enhance human productivity, from assisting in complex decision-making to driving scientific breakthroughs. However, recent advances in reasoning, such as OpenAI's o1 and DeepSeek's r1, lack transparency, limiting broader research progress. Our work aims to push the frontier of reasoning in a fully open manner, fostering innovation and collaboration to accelerate advancements that ultimately benefit society.

\section*{Acknowledgements}

We thank Ryan Marten for generating traces from DeepSeek r1 for s1.1 using Bespoke Curator~\citep{curator}. This work was partly conducted using the Stanford Marlowe GPU cluster \cite{kapfer_2025_14751899}, made possible by financial support from Stanford University. We thank Alexander M. Rush, Andrew Ilyas, Banghua Zhu, Chenglei Si, Chunting Zhou, John Yang, Ludwig Schmidt, Samy Jelassi, Suhas Kotha, Tengyu Ma, Xuechen Li, Yu Sun, and Yue Zhang for very constructive discussions.

\bibliography{reference}
\bibliographystyle{icml2025}

\newpage
\appendix
\onecolumn

\begin{spacing}{0.2}
\tableofcontents
\end{spacing}

\newpage

\FloatBarrier

\section{\textbf{s1.1}}
\label{sec:s11}

Seven days after our release of s1, we released s1.1. We regenerated traces for our 1,000 samples in \data{} using DeepSeek r1~\citep{r1} to create \data{}\textbf{-1.1}. We use the same training procedure to train our model \textbf{s1.1}. Other updates since our launch include the release of o3~\citep{o3}, LIMO~\citep{ye2025limoreasoning}, and AIME 2025. We consider all these new developments in \autoref{tab:perf2}. We find that s1.1 performs significantly better than s1. We also tried distilling from Claude 3.7, which led to worse performance than from r1 (not reported).

\begin{table}[htbp]
\centering
\caption{\textbf{\model{} is an open and sample-efficient reasoning model.} We evaluate \model{}, Qwen, and Gemini (some entries are unknown (N.A.), see \autoref{sec:results}). Other results are from the respective reports~\citep{qwen2024qwen25technicalreport,qwq-32b-preview,o1,o3,r1,bespoke_stratos,sky_t1} except for AIME 2025~\citep{ye2025aimepreview}. \# ex. = number examples used for reasoning finetuning; BF = budget forcing.}
\begin{tabular}{lrrrrrrrr}
\toprule
Model & \# Examples & MATH500 & GPQA & AIME 2024 & AIME 2025 \\
\midrule
\multicolumn{6}{c}{\textbf{API only}} \\
\midrule
o3-mini-low & N/A & 95.8 & 70.6 & 56.3 & 42.1 \\
o3-mini-medium & N/A & 97.3 & 76.8 & 75.8 & 70.4\\
o3-mini-high & N/A & 97.9 & \textbf{79.7} & \textbf{83.8} & 80.9 \\
\midrule
\multicolumn{6}{c}{\textbf{Open Weights}} \\
\midrule
QwQ-32B & N.A. & 90.6 & 54.5 & 46.7 & 32.7 \\
r1 & $\gg$800K & 07.3 & \textbf{71.5} & \textbf{79.8} & 70.0 \\
r1-distill-Llama-70B & 800K & 94.5 & 65.2 & 57.1 & 56.3 \\
r1-distill-Qwen-14B & 800K & 93.9 & 59.1 & 61.7 & 48.0 \\
r1-distill-Qwen-32B & 800K & 94.3 & 62.1 & 58.3 & 49.6 \\
\midrule
\multicolumn{6}{c}{\textbf{Open Weights and Open Data}} \\
\midrule
LIMO & 817 & 94.8 & \textbf{66.7} & 56.3 & 44.6 \\
s1 w/o BF & 1K & 92.6 & 56.6 & 50.0 & 26.7 \\
s1 with Budget Forcing ``Wait'' 1x & 1K & 92.8 & 59.6 & 53.3 & 30.0 \\
s1 with Budget Forcing ``Wait'' 2x & 1K & 93.0 & 59.6 & 53.3 & 33.3 \\
s1 with Budget Forcing ``Wait'' 4x & 1K & 92.2 & 58.6 & \textbf{56.7} & 36.7 \\
s1.1 w/o BF & 1K & 94.4 & 60.6 & \textbf{56.7} & \textbf{50.0} \\
s1.1 with Budget Forcing ``Wait'' 1x & 1K & \textbf{95.4} & 62.6 & \textbf{56.7} & \textbf{50.0} \\
s1.1 with Budget Forcing ``Wait'' 2x & 1K & \textbf{95.4} & 63.6 & \textbf{56.7} & \textbf{50.0} \\
\bottomrule
\label{tab:perf2}
\end{tabular}
\end{table}

\section{Evaluation determinism}
\label{sec:eval-determinism}

We run our evaluations using vLLM~\citep{kwon2023efficientmemorymanagementlarge} as it is faster than the alternatives we tried. However, we find that even when using the same random seeds and greedy sampling, evaluation scores can change significantly across runs:
\begin{itemize}
\item Different batch sizes causing different results see \url{https://github.com/vllm-project/vllm/issues/5898}
\item Continuing generations causing different results see \url{https://github.com/vllm-project/vllm/issues/11783}
\item Changes in tensor parallelism causing different results
\end{itemize}

As our model generates long reasoning traces prior to its answer, small numeric changes can snowball into large differences. We encounter many generations that are exactly the same for thousands of tokens and then suddenly differ in one token eventually ending up with an entirely different answer. To partly counter this issue we generally run our final evaluations using full precision unless otherwise indicated.

\FloatBarrier

\section{\data{} details}
\label{sec:details}

\subsection{\data{} summary}

\begin{table*}[htbp]
\centering
\caption{\textbf{Summary of our dataset \data{}}. Token count measured by the Qwen-2.5 tokenizer. We prompt Claude to produce keywords given several questions from the domain.}
\begin{tabular}{>{\raggedright}p{3.5cm} l l l l l}
\toprule
Domain & \#questions & Total token count & Keywords \\
\midrule
Geometry & 109 & 560.2K & Area, Triangle, Distance \\
Number theory & 98 & 522.5K & Sequences, Divisibility \\
Combinatorics & 75 & 384.7K & Permutations, Counting \\
Real functions & 43 & 234.8K & Trigonometry, Calculus  \\
Biology & 41 & 120.9K & Organic reactions \\
Complex functions & 32 & 170.2K & Complex roots \\
Quantum theory & 32 & 127.9K & Particles, Wave functions \\
Field theory & 28 & 150.1K & Polynomials, Roots \\
Calculus of variations & 28 & 155.5K & Optimization, Control \\
Difference equations & 24 & 132.5K & Recurrence, Recursion \\
Electromagnetic theory & 23 & 95.8K & Optics, Waves, Diffraction \\
Group theory & 22 & 100.0K & Groups, Automorphisms \\
Linear algebra & 22 & 128.3K & Matrices, Determinants \\
Probability theory & 20 & 114.6K & Random walk, Expectation \\
Algebraic systems & 19 & 109.9K & Functional equations \\
Mechanics & 19 & 103.6K & Forces, Motion, Energy \\
Thermodynamics & 19 & 74.2K &  Heat engines, Entropy \\
Differential equations & 18 & 89.6K & Substitution, Existence \\
Computer science & 18 & 34.2K & Complexity theory, Algorithms \\
Numerical analysis & 18 & 76.5K & Error analysis, Stability \\
Calculus & 17 & 96.3K & Convergence, Summation \\
Algebraic structures & 17 & 90.4K & Inequalities, Sets \\
Astronomy & 16 & 37.7K & Stellar populations, Orbits \\
Remaining 27 domains & 242 & 982.2K & Domains with $\leq$ 16 questions  \\
\midrule
All domains (51) & 1000 & 4.7M &  \data{}\\
\bottomrule
\end{tabular}
\label{tab:domain_distribution}
\end{table*}

\subsection{Dataset composition for full 59K questions}

\begin{table}[htbp]
\centering
\caption{\textbf{Composition of full 59K questions.}
Thinking and response lengths are measured in tokens using the Qwen2.5-32B-Instruct tokenizer~\citep{qwen2024qwen25technicalreport}. In addition to excluding our evaluation benchmark, AIME24, we also exclude AIME questions from 2022-2023 as we use these 90 questions during our development stage of \model{}.}
\begin{tabular}{>{\raggedright}p{4.7cm} p{5.4cm} p{1.2cm} p{1.2cm} p{1.2cm}}
\toprule
Source & Description & \#Samples & Avg. thinking length \\
\midrule
NuminaMATH~\citep{numina_math_datasets} & Math problems from online websites & 30660 & 4.1K \\
MATH~\citep{hendrycks2021measuringmathematicalproblemsolving} & Math problems from competitions & 11999 & 2.9K \\
OlympicArena~\citep{huang2024olympicarenabenchmarkingmultidisciplinecognitive} & Astronomy, Biology, Chemistry, Computer Science, Geography, Math, and Physics olympiad questions & 4250 & 3.2K\\
OmniMath~\citep{gao2024omnimathuniversalolympiadlevel} & Math problems from competitions & 4238 & 4.4K\\
AGIEval~\citep{zhong2023agievalhumancentricbenchmarkevaluating,ling2017programinductionrationalegeneration,hendrycks2021measuringmathematicalproblemsolving,liu2020logiqachallengedatasetmachine,zhong2019jecqalegaldomainquestionanswering,wang2021lsatprogresschallengescomplex} & English, Law, Logic and Math problems from the SAT, LSAT and other exams & 2385 & 1.2K\\
xword & Crossword puzzles & 999 & 0.7K \\
OlympiadBench~\citep{he2024olympiadbenchchallengingbenchmarkpromoting} & Math and Physics olympiad questions & 896 & 3.9K\\
AIME (1983-2021) & American Invitational Mathematics Examination & 890 & 4.7K \\
TheoremQA~\citep{chen2023theoremqatheoremdrivenquestionanswering} & Computer Science, Finance, Math, and Physics university-level questions relating to theorems  & 747 & 2.1K \\
USACO \citep{shi2024languagemodelssolveolympiad} & Code problems from the USA Computing Olympiad & 519 & 3.6K \\
JEEBench~\citep{arora2023llmsadvancedenoughchallenging} & Chemistry, Math, and Physics problems used in the university entrance examination of the Indian Institute of Technology & 515 & 2.9K \\
GPQA~\citep{rein2023gpqagraduatelevelgoogleproofqa} & PhD-Level Science Questions & 348 & 2.9K \\
SciEval~\citep{sun2024scievalmultilevellargelanguage} & Biology, Chemistry, and Physics problems from various sources & 227 & 0.7K \\
s1-prob & Stanford statistics qualifying exams & 182 & 4.0K \\
LiveCodeBench~\citep{jain2024livecodebenchholisticcontaminationfree} & Code problems from coding websites (LeetCode, AtCoder, and CodeForces) &  151 & 3.5K\\
s1-teasers & Math brain-teasers crawled from the Internet & 23 & 4.1K \\
\midrule
\textbf{All 59K questions} & Composite of the above datasets with reasoning traces and solutions & 59029 & 3.6K \\
\bottomrule
\end{tabular}
\label{tab:ds}
\end{table}

\FloatBarrier

\subsection{\data{} grading prompt}
\label{sec:grading}

To grade whether an example is correct for our dataset selection in \autoref{sec:data}, we use the prompt in \autoref{fig:grade}. We grade using Claude 3.5 except for the correctness among the final 1,000 samples, which we graded with Claude 3.7.

\begin{figure*}[ht]
\begin{tabular}{@{}p{\columnwidth}@{}}
\toprule
You are an AI assistant for grading a science problem.
The user will provide you with the question itself, an attempt made by a student and the correct answer to the problem.
Your job is to judge whether the attempt is correct by comparing it with the correct answer. 
If the expected solution concludes with a number or choice, there should be no ambiguity.
If the expected solution involves going through the entire reasoning process, you should judge the attempt based on whether the reasoning process is correct with correct answer if helpful.

~

The user will provide the attempt and the correct answer in the following format:

~

\# Problem

\{problem\}

~

\#\# Attempt

\{attempt\}

~

\#\# Correct answer

\{solution\}

~

Explain your reasoning, and end your response on a new line with only "Yes" or "No" (without quotes).
\\
\bottomrule
\end{tabular}
\caption{\textbf{Grading prompt.}}
\label{fig:grade}
\end{figure*}

\subsection{\data{} diversity selection}
\label{sec:algo}

\begin{algorithm}
\caption{Two-stage sampling for \data{}}
\label{alg:twostage}
\begin{algorithmic}[1]
\STATE \textbf{Input:} $\mathcal{Q}$ := Set of 24,496 questions with features
\STATE \textbf{Output:} $\mathcal{S}$ := Set of 1,000 selected questions
\STATE $\mathcal{S} \gets \emptyset$ \hfill \textit{Initialize the output set (only tracks unique elements)}

\FOR{$q \in \mathcal{Q}$}
    \IF{IsGeminiCorrect($q$) \AND (IsAIME($q$) \OR IsGPQA($q$))}
        \STATE $\mathcal{S} \gets \mathcal{S} \cup \{q\}$ 
        \STATE \hfill \textit{Select all correct AIME/GPQA solutions}
    \ELSIF{IsGeminiCorrect($q$) \AND IsMATH($q$) \AND ThinkingLength($q$) > 5600}
        \STATE $\mathcal{S} \gets \mathcal{S} \cup \{q\}$ 
        \STATE \hfill \textit{Select correct MATH500 solutions with long chains}
    \ENDIF
\ENDFOR

\STATE $\mathcal{D} \gets$ All available domains 
\STATE \hfill \textit{Initialize domain pool}

\WHILE{$|\mathcal{S}| < 1000$}
    \STATE $d \gets$ RandomChoice($\mathcal{D}$) 
    \STATE \hfill \textit{Randomly select a domain}
    \STATE $Q_d \gets$ Questions in domain $d$ 
    \STATE \hfill \textit{Get questions from this domain}
    \STATE ranks $\gets$ RankByThinkingLength($Q_d$) 
    \STATE \hfill \textit{Rank by thinking length}
    \STATE weights $\gets 2^{-\text{ranks}}$ 
    \STATE \hfill \textit{Apply power-law weighting}
    \STATE $q \gets$ WeightedSample($Q_d$, weights) 
    \STATE \hfill \textit{Sample favoring longer chains}
    \STATE $\mathcal{S} \gets \mathcal{S} \cup \{q\}$ 
    \STATE \hfill \textit{Add selected question}
    \STATE $Q_d \gets Q_d \setminus \{q\}$ 

    \IF{$Q_d = \emptyset$}
        \STATE $\mathcal{D} \gets \mathcal{D} \setminus \{d\}$ 
        \STATE \hfill \textit{Remove exhausted domains}
    \ENDIF
\ENDWHILE
\end{algorithmic}
\end{algorithm}

\autoref{alg:twostage} provides our algorithm for selecting data in our diversity selection stage. As mentioned in \autoref{sec:data}, we also include samples from some specific benchmarks we perceive as high-quality. None of the samples overlap with our final evaluation.

\subsection{Decontamination}
\label{sec:decontaminate}
We filter all samples by checking for an 8-gram overlap between the selected examples and the evaluation benchmarks: MATH500, GPTQA Diamond, and AIME24. We exclude questions with more than an 8-gram overlap.

\newpage
\clearpage
\section{Training details}

We take a model that has already been pretrained and instruction tuned and further finetune it for reasoning. Specifically, we use Qwen2.5-32B-Instruct~\citep{qwen2024qwen25technicalreport}, which on math tasks generally matches or outperforms the larger Qwen2.5-72B-Instruct~\citep{qwen2024qwen25technicalreport} or other open models~\citep{dubey2024llama3herdmodels,groeneveld2024olmo,muennighoff2024olmoeopenmixtureofexpertslanguage}. We use token delimiters to separate the thinking stage from the answering stage. We enclose the thinking stage with \verb@<|im_start|>think@ and \verb@<|im_start|>answer@; both preceded and followed by a newline. Samples from our dataset are in \autoref{sec:samples}. We use basic fine-tuning hyperparameters: we train for 5 epochs with a batch size of 16 for a total of 315 gradient steps. We train in bfloat16 precision with a learning rate of $1e-5$ warmed up linearly for 5\% (16 steps) and then decayed to 0 over the rest of training (299 steps) following a cosine schedule. We use the AdamW optimizer~\citep{loshchilov2019decoupled} with $\beta_1=0.9, \beta_2=0.95$ and weight decay of $1e-4$. We do not compute loss on questions, only on reasoning traces and solutions. We ensure the sequence length is large enough to avoid cutting off any samples; a setting we ablate in \autoref{sec:trainabl}. The training takes just 26 minutes on 16 NVIDIA H100 GPUs.
\label{sec:details-training}
\begin{figure*}[htbp]
\centering
\begin{center}
\includegraphics[width=\textwidth]{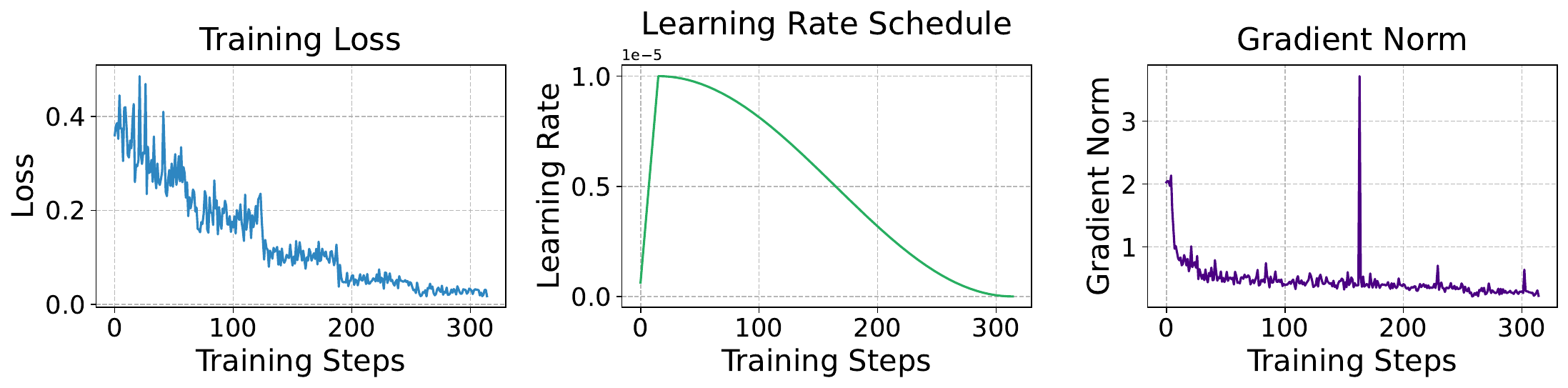}
\caption{\textbf{Training dynamics of \model{} on \data{}.}}
\label{fig:training_metrics}
\end{center}
\end{figure*}

\subsection{Training Ablations: Sequence length}
\label{sec:trainabl}

\begin{table}[htbp]
\centering
\caption{\textbf{Training sequence length ablation.} We report ``accuracy / average thinking tokens per sample''; the higher the accuracy and the fewer the thinking tokens (inference cost) the better.}
\begin{tabular}{p{4cm} p{3.7cm} p{3.7cm}}
\toprule
& Model A & Model B \\
\midrule
Training sequence length & 4096 & 32768 \\
\% training samples cutoff & 74\% & 0\% \\
\midrule
AIME24 & 30.0\% / 20721 & 50.0\% / 6984 \\
MATH500 & 90.0\% / 5324 & 91.0\% / 3268 \\
GPQA & 52.5\% / 6841 & 53.0\% / 3568 \\
\bottomrule
\end{tabular}
\label{tab:seqabl}
\end{table}

Besides our scaling ablations in \autoref{sec:scaleabl}, the main training hyperparameter we ablate is the sequence length used during training. We find that a \textbf{shorter} training sequence length leads to \textbf{longer} reasoning traces at test time. This is because when training with a shorter sequence length the answer section of the training sample is more commonly cut off. Inversely, when the training sequence length is longer, more samples appear in their entirety with the section where the model answers. Thus the model receives more gradient updates where it learns to generate an answer following its chain. This in turn leads to a higher log probability of the answer section at any point during the generation and thus shorter reasoning traces at test time. Performance-wise, we also find that the model trained with a longer sequence length performs better. Thus we opt for the longest training sequence length as it leads to better performance and makes inference more efficient by leading to shorter reasoning traces.

\FloatBarrier

\subsection{Training Samples}
\label{sec:samples}

\autoref{tab:prob}, \autoref{tab:teasers}, \autoref{tab:arena} contain training samples from \data{}.

\newpage

\begin{longtable}{@{}p{\textwidth}@{}}
\caption{\textbf{\data{} sample stemming from s1-prob.} The black text is the question and the \textcolor{olmoeDarkYellow}{yellow} text is the solution. Based solely on the question we generate the \textcolor{defaultlightblue}{light blue} reasoning trace and \textcolor{defaultblue}{blue} response with Gemini, which we use for training.} 
\label{tab:prob} \\
\toprule

You are given integrable random variables \( X, Y_0 \) and \( Z_0 \) on the same probability space \( (\Omega, \mathcal{F}, \mathbb{P}) \), and two \( \sigma \)-algebras \( \mathcal{A} \subset \mathcal{F}, \mathcal{B} \subset \mathcal{F}. \) For \( k = 1, 2, \ldots, \), let \[ Y_k := \mathbb{E}[X|\sigma(\mathcal{A}, Z_0, \ldots, Z_{k-1})] , \quad Z_k := \mathbb{E}[X|\sigma(\mathcal{B}, Y_0, \ldots, Y_{k-1})] . \] (a) Show that there exist integrable random variables \( Y_\infty \) and \( Z_\infty \) such that as \( n \to \infty \) both \( \mathbb{E}[|Y_n - Y_\infty|] \to 0 \) and \( \mathbb{E}[|Z_n - Z_\infty|] \to 0 \). (b) Prove that almost surely \( Y_\infty = Z_\infty. \)

\\

\defaultyellow{
\begin{enumerate}
\item Introduce the notations, $\mathcal{F}_n:=\sigma(\mathcal{A},Z_0, \ldots, Z_{n-1})$ and $\mathcal{G}_n:=\sigma(\mathcal{B},Y_0, \ldots, Y_{n-1})$, for all $n \geq 1$ and $\mathcal{F}_0:=\mathcal{A}$, $\mathcal{G}_0:=\mathcal{B}$. Clearly, both are filtrations and by definition,  $Y_n=\mathbb{E} \left[X \rvert \mathcal{F}_n \right]$ and $Z_n=\mathbb{E} \left[X \rvert \mathcal{G}_n \right]$, for all $n \geq 0$. Thus $\left\{Y_n\right\}_{n \geq 0}$ (or $\left\{Z_n\right\}_{n \geq 0}$ ) is a \textit{Doob's Martingale} with respect to filtration $\left\{\mathcal{F}_n\right\}_{n \geq 0}$ (or $\left\{\mathcal{G}_n\right\}_{n \geq 0}$) and hence is U.I., $X$ being integrable. Now apply the fact that a Martingale is U.I. if and only if it converges in $L^1$ and conclude.
\item We provide two different proofs. 
\end{enumerate}
\begin{enumerate}
\item Observe the following identity which follows from \textit{Tower Property}. \begin{equation}{\label{tower}} \mathbb{E} \left[Y_n \rvert \mathcal{G}_n \right] = \mathbb{E} \left[ \mathbb{E} \left[X \rvert \mathcal{F}_n \right] \rvert \mathcal{G}_n \right] = \mathbb{E} \left[ \mathbb{E} \left[X \rvert \mathcal{G}_n \right] \rvert \mathcal{F}_n \right] = \mathbb{E} \left[Z_n \rvert \mathcal{F}_n \right].\end{equation} Note that $\mathbb{E}\rvert \mathbb{E} \left[Y_n \rvert \mathcal{G}_n \right] - \mathbb{E} \left[Y_{\infty} \rvert \mathcal{G}_n \right]\rvert \leq \mathbb{E} |Y_n - Y_{\infty}| =o(1)$ and by \textit{Levy's Upward Theorem}, $\mathbb{E} \left[Y_{\infty} \rvert \mathcal{G}_n \right] \stackrel{L^1}{\longrightarrow} \mathbb{E} \left[Y_{\infty} \rvert \mathcal{G}_{\infty} \right]$, where $\mathcal{G}_{\infty} = \sigma \left( \cup_{n \geq 0} \mathcal{G}_n\right) = \sigma(\mathcal{B},Y_0,Y_1 \ldots).$ It is evident that $Y_{\infty} \in m\mathcal{G}_{\infty}$ (since it is almost sure limit of $Y_n \in m\mathcal{G}_n$) and hence $\mathbb{E} \left[Y_{\infty} \rvert \mathcal{G}_{\infty} \right]=Y_{\infty}$ almost surely. Combining the above observations we conclude that $\mathbb{E} \left[Y_n \rvert \mathcal{G}_n \right] \stackrel{L^1}{\longrightarrow} Y_{\infty}.$ Similarly, $\mathbb{E} \left[Z_n \rvert \mathcal{F}_n \right] \stackrel{L^1}{\longrightarrow} Z_{\infty}.$ Now invoke Equation~({\ref{tower}}), to conclude that $Y_{\infty}=Z_{\infty}$ almost surely.
\item Since $X$ is integrable, we have by \textit{Levy's Upward Theorem}, $Y_{\infty}= \lim_{n \to \infty} Y_n = \lim_{n \to \infty}\mathbb{E} \left[X \rvert \mathcal{F}_n \right] = \mathbb{E} \left[X \rvert \mathcal{F}_{\infty} \right].$ Similarly, $Z_{\infty} = \mathbb{E} \left[X \rvert \mathcal{G}_{\infty} \right].$ On the otherhand, in first proof we have observed that $Y_{\infty} \in m\mathcal{G}_{\infty}$ and hence $Y_{\infty} \in m\mathcal{G}_{\infty} \cap m\mathcal{F}_{\infty} = m\left(\mathcal{G}_{\infty} \cap \mathcal{F}_{\infty}\right)$. This yields the following. $$ Y_{\infty} =  \mathbb{E} \left[Y_{\infty} \rvert \mathcal{G}_{\infty} \cap \mathcal{F}_{\infty} \right] = \mathbb{E} \left[ \mathbb{E} \left[X \rvert \mathcal{F}_{\infty} \right]\rvert \mathcal{G}_{\infty} \cap \mathcal{F}_{\infty} \right] = \mathbb{E} \left[X\rvert \mathcal{G}_{\infty} \cap \mathcal{F}_{\infty} \right]. $$ Similarly, $Z_{\infty} = \mathbb{E} \left[X\rvert \mathcal{G}_{\infty} \cap \mathcal{F}_{\infty} \right].$ This proves that $Y_{\infty}=Z_{\infty}$ almost surely. 
\end{enumerate}
}

\\

\defaultlightblue{
Let's analyze the problem. We are given two sequences of random variables \(Y_k\) and \(Z_k\) defined by conditional expectations. We need to show that these sequences converge in \(L^1\) and that their limits are equal almost surely.

Part (a): Convergence in \(L^1\).

The sigma-algebras involved are increasing:

\( \mathcal{F}_k = \sigma(\mathcal{A}, Z_0, \ldots, Z_{k-1}) \)

\( \mathcal{G}_k = \sigma(\mathcal{B}, Y_0, \ldots, Y_{k-1}) \)

For the \(Y_k\) sequence:

\( Y_1 = \mathbb{E}[X|\sigma(\mathcal{A})] \)

\( Y_2 = \mathbb{E}[X|\sigma(\mathcal{A}, Z_0)] \)

\( Y_3 = \mathbb{E}[X|\sigma(\mathcal{A}, Z_0, Z_1)] \)

...

The sequence of sigma-algebras \( \sigma(\mathcal{A}, Z_0, \ldots, Z_{k-1}) \) is increasing. By the Martingale Convergence Theorem (for conditional expectations), since \(X\) is integrable, the sequence \(Y_k\) converges almost surely and in \(L^1\) to \( \mathbb{E}[X|\sigma(\mathcal{A}, Z_0, Z_1, \ldots)] \). Let \( Y_\infty = \mathbb{E}[X|\sigma(\mathcal{A}, Z_0, Z_1, \ldots)] \).

}
\\
\defaultlightblue{

Similarly, for the \(Z_k\) sequence:

\( Z_1 = \mathbb{E}[X|\sigma(\mathcal{B})] \)

\( Z_2 = \mathbb{E}[X|\sigma(\mathcal{B}, Y_0)] \)

\( Z_3 = \mathbb{E}[X|\sigma(\mathcal{B}, Y_0, Y_1)] \)

...

The sequence of sigma-algebras \( \sigma(\mathcal{B}, Y_0, \ldots, Y_{k-1}) \) is increasing. By the Martingale Convergence Theorem, the sequence \(Z_k\) converges almost surely and in \(L^1\) to \( \mathbb{E}[X|\sigma(\mathcal{B}, Y_0, Y_1, \ldots)] \). Let \( Z_\infty = \mathbb{E}[X|\sigma(\mathcal{B}, Y_0, Y_1, \ldots)] \).

Part (b): Prove that almost surely \( Y_\infty = Z_\infty \).

We have \( Y_\infty = \mathbb{E}[X|\sigma(\mathcal{A}, Z_0, Z_1, \ldots)] \) and \( Z_\infty = \mathbb{E}[X|\sigma(\mathcal{B}, Y_0, Y_1, \ldots)] \).

Let \( \mathcal{Y}_\infty = \sigma(\mathcal{A}, Z_0, Z_1, \ldots) \) and \( \mathcal{Z}_\infty = \sigma(\mathcal{B}, Y_0, Y_1, \ldots) \).

Then \( Y_\infty = \mathbb{E}[X|\mathcal{Y}_\infty] \) and \( Z_\infty = \mathbb{E}[X|\mathcal{Z}_\infty] \).

Consider \( Y_1 = \mathbb{E}[X|\mathcal{A}] \). \( Z_1 = \mathbb{E}[X|\mathcal{B}] \).

\( Y_2 = \mathbb{E}[X|\sigma(\mathcal{A}, Z_0)] \). Since \( Z_0 = \mathbb{E}[X|\mathcal{B}] \), \( \sigma(\mathcal{A}, Z_0) = \sigma(\mathcal{A}, \mathbb{E}[X|\mathcal{B}]) \).

\( Z_2 = \mathbb{E}[X|\sigma(\mathcal{B}, Y_0)] \). Since \( Y_0 \) is not defined, let's assume \( Y_0 \) and \( Z_0 \) are given.

Let's consider the sigma-algebra generated by all the \(Y_i\) and \(Z_i\).

Let \( \mathcal{G} = \sigma(Y_0, Z_0, Y_1, Z_1, \ldots) \).

Both \(Y_k\) and \(Z_k\) are measurable with respect to \(\mathcal{G}\). Thus, \(Y_\infty\) and \(Z_\infty\) are measurable with respect to \(\mathcal{G}\).

Consider \( \mathbb{E}[Y_{n+1}|\mathcal{A}, Z_0, \ldots, Z_{n-1}] = \mathbb{E}[\mathbb{E}[X|\sigma(\mathcal{A}, Z_0, \ldots, Z_n)]|\mathcal{A}, Z_0, \ldots, Z_{n-1}] = Y_n \).

Consider \( \mathbb{E}[Z_{n+1}|\mathcal{B}, Y_0, \ldots, Y_{n-1}] = \mathbb{E}[\mathbb{E}[X|\sigma(\mathcal{B}, Y_0, \ldots, Y_n)]|\mathcal{B}, Y_0, \ldots, Y_{n-1}] = Z_n \).

Let \( \mathcal{H} = \sigma(\mathcal{A} \cup \mathcal{B} \cup \sigma(Y_0, Z_0, Y_1, Z_1, \ldots)) \).

Consider \( \mathbb{E}[X|\mathcal{A}, Z_0, \ldots, Z_{k-1}] = Y_k \).

Consider \( \mathbb{E}[X|\mathcal{B}, Y_0, \ldots, Y_{k-1}] = Z_k \).

Let's consider the limit sigma-algebras \( \mathcal{Y}_\infty = \sigma(\mathcal{A}, Z_0, Z_1, \ldots) \) and \( \mathcal{Z}_\infty = \sigma(\mathcal{B}, Y_0, Y_1, \ldots) \).

We know that \( Y_\infty \) is \( \mathcal{Y}_\infty \)-measurable and \( Z_\infty \) is \( \mathcal{Z}_\infty \)-measurable.

Let \( \mathcal{C} = \sigma(\mathcal{A} \cup \mathcal{B}) \).

\( Y_1 = \mathbb{E}[X|\mathcal{A}] \). \( Z_1 = \mathbb{E}[X|\mathcal{B}] \).

\( Y_2 = \mathbb{E}[X|\sigma(\mathcal{A}, Z_0)] = \mathbb{E}[X|\sigma(\mathcal{A}, \mathbb{E}[X|\mathcal{B}])] \).

\( Z_2 = \mathbb{E}[X|\sigma(\mathcal{B}, Y_0)] = \mathbb{E}[X|\sigma(\mathcal{B}, \mathbb{E}[X|\mathcal{A}])] \).

Consider the case when \( \mathcal{A} \) and \( \mathcal{B} \) are independent. Then \( Y_1 = \mathbb{E}[X|\mathcal{A}] \) and \( Z_1 = \mathbb{E}[X|\mathcal{B}] \).

\( Y_2 = \mathbb{E}[X|\sigma(\mathcal{A}, \mathbb{E}[X|\mathcal{B}])] \).

Let's consider the sigma-algebra generated by all the information. Let \( \mathcal{F}_\infty = \sigma(\mathcal{A}, \mathcal{B}, Y_0, Z_0, Y_1, Z_1, \ldots) \).

Then both \( Y_\infty \) and \( Z_\infty \) are \( \mathcal{F}_\infty \)-measurable.

We have \( Y_\infty = \lim_{n \to \infty} Y_n \) and \( Z_\infty = \lim_{n \to \infty} Z_n \) a.s. and in \( L^1 \).

Consider \( \mathbb{E}[X|\mathcal{A}, Z_0, Z_1, \ldots] = Y_\infty \) and \( \mathbb{E}[X|\mathcal{B}, Y_0, Y_1, \ldots] = Z_\infty \).

Let \( \mathcal{H}_\infty = \sigma(\mathcal{A} \cup \mathcal{B}) \).

Consider \( \mathbb{E}[X|\mathcal{H}_\infty] \).

Let's try to show that both \( Y_\infty \) and \( Z_\infty \) are equal to \( \mathbb{E}[X|\sigma(\mathcal{A} \vee \mathcal{B})] \). This seems too simple.

Let's consider the definition of conditional expectation. For any set \( A \in \mathcal{Y}_\infty \), \( \mathbb{E}[Y_\infty 1_A] = \mathbb{E}[X 1_A] \).

For any set \( B \in \mathcal{Z}_\infty \), \( \mathbb{E}[Z_\infty 1_B] = \mathbb{E}[X 1_B] \).

Consider \( \mathbb{E}[Y_{k+1}|\mathcal{F}_k] = Y_k \) where \( \mathcal{F}_k = \sigma(\mathcal{A}, Z_0, \ldots, Z_{k-1}) \).

Consider \( \mathbb{E}[Z_{k+1}|\mathcal{G}_k] = Z_k \) where \( \mathcal{G}_k = \sigma(\mathcal{B}, Y_0, \ldots, Y_{k-1}) \).

Consider the sigma-algebra \( \sigma(\mathcal{A} \cup \mathcal{B}) \).

Let \( W = \mathbb{E}[X|\sigma(\mathcal{A} \cup \mathcal{B})] \).

We have \( Y_\infty = \mathbb{E}[X|\sigma(\mathcal{A}, Z_0, Z_1, \ldots)] \).

Since \( Z_k \) depends on \( Y_{k-1} \) which depends on \( Z_{k-2} \) etc., it is complex.

Let's consider the case where \( Y_0 = \mathbb{E}[X|\mathcal{A}] \) and \( Z_0 = \mathbb{E}[X|\mathcal{B}] \).

Then \( Y_1 = \mathbb{E}[X|\mathcal{A}] = Y_0 \).

\( Z_1 = \mathbb{E}[X|\mathcal{B}] = Z_0 \).

If this were the case, then \( Y_k = Y_0 \) and \( Z_k = Z_0 \), so \( Y_\infty = Y_0 \) and \( Z_\infty = Z_0 \).

We need \( Y_0 = Z_0 \) a.s., i.e., \( \mathbb{E}[X|\mathcal{A}] = \mathbb{E}[X|\mathcal{B}] \) a.s. This is not generally true.

Consider the sigma-algebra generated by all \(Y_i\) and \(Z_i\), denoted by \( \mathcal{G}_\infty = \sigma(Y_0, Z_0, Y_1, Z_1, \ldots) \).

Then \( \mathcal{Y}_\infty = \sigma(\mathcal{A}, Z_0, Z_1, \ldots) \subseteq \sigma(\mathcal{A}, \mathcal{G}_\infty) \).

\( \mathcal{Z}_\infty = \sigma(\mathcal{B}, Y_0, Y_1, \ldots) \subseteq \sigma(\mathcal{B}, \mathcal{G}_\infty) \).

}
\\
\defaultlightblue{

Consider \( \mathbb{E}[X|\mathcal{A}, Z_0, Z_1, \ldots] = Y_\infty \). This means for any \( A \in \sigma(\mathcal{A}, Z_0, Z_1, \ldots) \), \( \mathbb{E}[X 1_A] = \mathbb{E}[Y_\infty 1_A] \).

Consider \( \mathbb{E}[X|\mathcal{B}, Y_0, Y_1, \ldots] = Z_\infty \). This means for any \( B \in \sigma(\mathcal{B}, Y_0, Y_1, \ldots) \), \( \mathbb{E}[X 1_B] = \mathbb{E}[Z_\infty 1_B] \).

Let \( \mathcal{H} = \sigma(\mathcal{A}, \mathcal{B}) \).

Consider \( \mathbb{E}[X|\mathcal{H}] \).

Consider \( \mathbb{E}[Y_\infty|\mathcal{B}, Y_0, Y_1, \ldots] = \mathbb{E}[\mathbb{E}[X|\mathcal{A}, Z_0, Z_1, \ldots]|\mathcal{B}, Y_0, Y_1, \ldots] \).

Let's use the fact that \( Y_\infty \) is \( \sigma(\mathcal{A}, Z_0, Z_1, \ldots) \)-measurable and \( Z_\infty \) is \( \sigma(\mathcal{B}, Y_0, Y_1, \ldots) \)-measurable.

Consider the smallest sigma-algebra containing both \( \sigma(\mathcal{A}, Z_0, Z_1, \ldots) \) and \( \sigma(\mathcal{B}, Y_0, Y_1, \ldots) \).

This is \( \sigma(\mathcal{A} \cup \mathcal{B} \cup \{Y_i\}_{i \ge 0} \cup \{Z_i\}_{i \ge 0}) \).

Let's try to show that \( \mathbb{E}[(Y_\infty - Z_\infty)^2] = 0 \).

\( \mathbb{E}[(Y_\infty - Z_\infty)^2] = \mathbb{E}[Y_\infty^2] + \mathbb{E}[Z_\infty^2] - 2 \mathbb{E}[Y_\infty Z_\infty] \).

Consider the definition of \(Y_k\) and \(Z_k\).

\( \mathbb{E}[Y_{k+1}|\mathcal{A}, Z_0, \ldots, Z_{k-1}] = Y_k \).

\( \mathbb{E}[Z_{k+1}|\mathcal{B}, Y_0, \ldots, Y_{k-1}] = Z_k \).

Consider \( \mathbb{E}[Y_\infty|\mathcal{B}, Y_0, Y_1, \ldots] \).

Since \( Y_\infty \) is measurable with respect to \( \sigma(\mathcal{A}, Z_0, Z_1, \ldots) \), this does not simplify easily.

Let's assume \( Y_\infty \neq Z_\infty \) on a set of positive measure.

Consider the case where \( \mathcal{A} \) and \( \mathcal{B} \) are trivial.

Then \( Y_k = \mathbb{E}[X|\sigma(Z_0, \ldots, Z_{k-1})] \) and \( Z_k = \mathbb{E}[X|\sigma(Y_0, \ldots, Y_{k-1})] \).

Let \( \mathcal{H}_\infty = \sigma(\mathcal{A} \cup \mathcal{B} \cup \{Y_i, Z_i\}_{i \ge 0}) \).

Consider \( \mathbb{E}[X|\mathcal{H}_\infty] \).

We have \( Y_\infty = \mathbb{E}[X|\sigma(\mathcal{A}, Z_0, Z_1, \ldots)] \).

By the properties of conditional expectation, \( \mathbb{E}[Y_\infty | \mathcal{A}] = \mathbb{E}[\mathbb{E}[X|\mathcal{A}, Z_0, Z_1, \ldots]|\mathcal{A}] \).

Consider the sigma-algebra \( \mathcal{D} \) generated by all the random variables involved.

\( \mathcal{D} = \sigma(\mathcal{A} \cup \mathcal{B} \cup \{Y_i\}_{i \ge 0} \cup \{Z_i\}_{i \ge 0}) \).

Let \( M_\infty = \mathbb{E}[X|\mathcal{A} \vee \mathcal{B}] \).

Consider the limits. As \( n \to \infty \), \( \sigma(\mathcal{A}, Z_0, \ldots, Z_{n-1}) \uparrow \sigma(\mathcal{A}, Z_0, Z_1, \ldots) \).

As \( n \to \infty \), \( \sigma(\mathcal{B}, Y_0, \ldots, Y_{n-1}) \uparrow \sigma(\mathcal{B}, Y_0, Y_1, \ldots) \).

Let \( \mathcal{Y}_\infty = \sigma(\mathcal{A}, Z_0, Z_1, \ldots) \) and \( \mathcal{Z}_\infty = \sigma(\mathcal{B}, Y_0, Y_1, \ldots) \).

We have \( Y_\infty = \mathbb{E}[X|\mathcal{Y}_\infty] \) and \( Z_\infty = \mathbb{E}[X|\mathcal{Z}_\infty] \).

Consider \( \mathbb{E}[Y_\infty - Z_\infty]^2 = \mathbb{E}[(Y_\infty - X) + (X - Z_\infty)]^2 \)

\( = \mathbb{E}[(Y_\infty - X)^2] + \mathbb{E}[(X - Z_\infty)^2] + 2 \mathbb{E}[(Y_\infty - X)(X - Z_\infty)] \)

\( = \mathbb{E}[(X - \mathbb{E}[X|\mathcal{Y}_\infty])^2] + \mathbb{E}[(X - \mathbb{E}[X|\mathcal{Z}_\infty])^2] + 2 \mathbb{E}[Y_\infty X - Y_\infty Z_\infty - X^2 + X Z_\infty] \)

\( = \mathbb{E}[X^2] - \mathbb{E}[Y_\infty^2] + \mathbb{E}[X^2] - \mathbb{E}[Z_\infty^2] + 2 \mathbb{E}[Y_\infty X] - 2 \mathbb{E}[Y_\infty Z_\infty] - 2 \mathbb{E}[X^2] + 2 \mathbb{E}[X Z_\infty] \)

\( = - \mathbb{E}[Y_\infty^2] - \mathbb{E}[Z_\infty^2] + 2 \mathbb{E}[Y_\infty^2] - 2 \mathbb{E}[Y_\infty Z_\infty] + 2 \mathbb{E}[Z_\infty^2] \)

\( = \mathbb{E}[Y_\infty^2] + \mathbb{E}[Z_\infty^2] - 2 \mathbb{E}[Y_\infty Z_\infty] \)

Consider the sigma-algebra \( \mathcal{C} = \sigma(\mathcal{A} \cup \mathcal{B}) \).

Let \( W = \mathbb{E}[X|\mathcal{C}] \).

Consider the limit sigma-algebra generated by all the information.

Let \( \mathcal{F}_\infty = \sigma(\mathcal{A}, \mathcal{B}, Y_0, Z_0, Y_1, Z_1, \ldots) \).

Then \( \mathcal{Y}_\infty \subseteq \mathcal{F}_\infty \) and \( \mathcal{Z}_\infty \subseteq \mathcal{F}_\infty \).

We have \( \mathbb{E}[X|\mathcal{Y}_\infty] = Y_\infty \) and \( \mathbb{E}[X|\mathcal{Z}_\infty] = Z_\infty \).

Consider \( \mathbb{E}[Y_\infty|\mathcal{Z}_\infty] \) and \( \mathbb{E}[Z_\infty|\mathcal{Y}_\infty] \).

Let \( \mathcal{H} = \sigma(\mathcal{A} \cup \mathcal{B}) \).

Consider \( \mathbb{E}[X|\mathcal{H}] \).

Consider the sequence of pairs \( (Y_k, Z_k) \).

If the limits are the same, then for any bounded measurable function \( f \), \( \mathbb{E}[f(Y_\infty)] = \mathbb{E}[f(Z_\infty)] \).

Consider the sigma-algebra \( \mathcal{D} = \sigma(\mathcal{A} \cup \mathcal{B} \cup \sigma(Y_0, Z_0, \ldots)) \).

Then both \( Y_\infty \) and \( Z_\infty \) are measurable with respect to \( \mathcal{D} \).

Consider \( \mathbb{E}[(Y_n - Z_n)^2] \).

Let \( \mathcal{G} = \sigma(\mathcal{A} \cup \mathcal{B}) \).

Consider \( W = \mathbb{E}[X|\mathcal{G}] \).

Then \( \mathbb{E}[W|\mathcal{A}] = \mathbb{E}[X|\mathcal{A}] \) and \( \mathbb{E}[W|\mathcal{B}] = \mathbb{E}[X|\mathcal{B}] \).

}
\\
\defaultlightblue{

Consider the limit sigma-algebra \( \mathcal{L} = \sigma(\mathcal{A} \cup \mathcal{B} \cup \sigma(Y_0, Z_0, \ldots)) \).

We have \( Y_\infty = \mathbb{E}[X|\sigma(\mathcal{A}, Z_0, Z_1, \ldots)] \) and \( Z_\infty = \mathbb{E}[X|\sigma(\mathcal{B}, Y_0, Y_1, \ldots)] \).

Let \( \mathcal{M} = \sigma(\mathcal{A}, Z_0, Z_1, \ldots) \vee \sigma(\mathcal{B}, Y_0, Y_1, \ldots) \).

Consider \( \mathbb{E}[X|\mathcal{M}] \).

Consider the random variable \( X \). Then \( Y_\infty = \lim_n \mathbb{E}[X|\mathcal{F}_n] \) and \( Z_\infty = \lim_n \mathbb{E}[X|\mathcal{G}_n] \).

Let \( \mathcal{H}_\infty = \sigma(\mathcal{A} \cup \mathcal{B} \cup \sigma(Y_0, Z_0, \ldots)) \).

Consider \( E = \{ \omega : Y_\infty(\omega) \neq Z_\infty(\omega) \} \). Suppose \( P(E) > 0 \).

Consider the conditional expectation with respect to the join of the sigma-algebras.

Let \( \mathcal{C}_\infty = \sigma(\mathcal{A}, \mathcal{B}, Y_0, Z_0, \ldots) \).

Consider \( \mathbb{E}[X|\mathcal{C}_\infty] \).

Then \( \mathbb{E}[\mathbb{E}[X|\mathcal{C}_\infty]|\mathcal{A}, Z_0, \ldots] = \mathbb{E}[X|\mathcal{A}, Z_0, \ldots] = Y_{n+1} \).

And \( \mathbb{E}[\mathbb{E}[X|\mathcal{C}_\infty]|\mathcal{B}, Y_0, \ldots] = \mathbb{E}[X|\mathcal{B}, Y_0, \ldots] = Z_{n+1} \).

So \( Y_\infty = \mathbb{E}[X|\sigma(\mathcal{A}, Z_0, Z_1, \ldots)] \) and \( Z_\infty = \mathbb{E}[X|\sigma(\mathcal{B}, Y_0, Y_1, \ldots)] \).

Let \( \mathcal{F}_\infty = \sigma(\mathcal{A}, Z_0, Z_1, \ldots) \) and \( \mathcal{G}_\infty = \sigma(\mathcal{B}, Y_0, Y_1, \ldots) \).

Consider \( \mathbb{E}[X|\mathcal{F}_\infty \vee \mathcal{G}_\infty] \).

Then \( \mathbb{E}[\mathbb{E}[X|\mathcal{F}_\infty \vee \mathcal{G}_\infty]|\mathcal{F}_\infty] = \mathbb{E}[X|\mathcal{F}_\infty] = Y_\infty \).

And \( \mathbb{E}[\mathbb{E}[X|\mathcal{F}_\infty \vee \mathcal{G}_\infty]|\mathcal{G}_\infty] = \mathbb{E}[X|\mathcal{G}_\infty] = Z_\infty \).

This means \( Y_\infty = \mathbb{E}[X|\mathcal{F}_\infty \vee \mathcal{G}_\infty] \) a.s. and \( Z_\infty = \mathbb{E}[X|\mathcal{F}_\infty \vee \mathcal{G}_\infty] \) a.s.

Therefore, \( Y_\infty = Z_\infty \) a.s.
}

\defaultblue{
Solution:

(a) For the sequence \( Y_k \), the \( \sigma \)-algebras \( \mathcal{F}_k = \sigma(\mathcal{A}, Z_0, \ldots, Z_{k-1}) \) are increasing as \( k \) increases. Since \( X \) is integrable, by the Martingale Convergence Theorem for conditional expectations, the sequence \( Y_k = \mathbb{E}[X|\mathcal{F}_k] \) converges almost surely and in \( L^1 \) to \( Y_\infty = \mathbb{E}[X|\sigma(\mathcal{A}, Z_0, Z_1, \ldots)] \). Similarly, for the sequence \( Z_k \), the \( \sigma \)-algebras \( \mathcal{G}_k = \sigma(\mathcal{B}, Y_0, \ldots, Y_{k-1}) \) are increasing. Since \( X \) is integrable, by the Martingale Convergence Theorem, the sequence \( Z_k = \mathbb{E}[X|\mathcal{G}_k] \) converges almost surely and in \( L^1 \) to \( Z_\infty = \mathbb{E}[X|\sigma(\mathcal{B}, Y_0, Y_1, \ldots)] \).

(b) Let \( \mathcal{F}_\infty = \sigma(\mathcal{A}, Z_0, Z_1, \ldots) \) and \( \mathcal{G}_\infty = \sigma(\mathcal{B}, Y_0, Y_1, \ldots) \). We have \( Y_\infty = \mathbb{E}[X|\mathcal{F}_\infty] \) and \( Z_\infty = \mathbb{E}[X|\mathcal{G}_\infty] \). Consider the \( \sigma \)-algebra \( \mathcal{H} = \mathcal{F}_\infty \vee \mathcal{G}_\infty = \sigma(\mathcal{A}, \mathcal{B}, Y_0, Z_0, Y_1, Z_1, \ldots) \). Consider \( W = \mathbb{E}[X|\mathcal{H}] \).

We have \( \mathcal{F}_\infty \subseteq \mathcal{H} \), so \( \mathbb{E}[W|\mathcal{F}_\infty] = \mathbb{E}[\mathbb{E}[X|\mathcal{H}]|\mathcal{F}_\infty] = \mathbb{E}[X|\mathcal{F}_\infty] = Y_\infty \).

Since \( W \) is \( \mathcal{H} \)-measurable, \( \mathbb{E}[W|\mathcal{F}_\infty] = Y_\infty \) implies that \( Y_\infty \) is the conditional expectation of \( W \) given \( \mathcal{F}_\infty \). But since \( Y_\infty \) is already \( \mathcal{F}_\infty \)-measurable, we have \( \mathbb{E}[W|\mathcal{F}_\infty] = W \) if \( W \) is \( \mathcal{F}_\infty \)-measurable, which is not necessarily true.

However, we know that \( \mathbb{E}[Y_\infty|\mathcal{F}_\infty] = Y_\infty \) and \( \mathbb{E}[Z_\infty|\mathcal{G}_\infty] = Z_\infty \).

Consider \( \mathbb{E}[X|\mathcal{F}_\infty \vee \mathcal{G}_\infty] \).

Then \( \mathbb{E}[\mathbb{E}[X|\mathcal{F}_\infty \vee \mathcal{G}_\infty]|\mathcal{F}_\infty] = \mathbb{E}[X|\mathcal{F}_\infty] = Y_\infty \).

This implies that \( \mathbb{E}[X|\mathcal{F}_\infty \vee \mathcal{G}_\infty] = Y_\infty \) almost surely.

Similarly, \( \mathbb{E}[\mathbb{E}[X|\mathcal{F}_\infty \vee \mathcal{G}_\infty]|\mathcal{G}_\infty] = \mathbb{E}[X|\mathcal{G}_\infty] = Z_\infty \).

This implies that \( \mathbb{E}[X|\mathcal{F}_\infty \vee \mathcal{G}_\infty] = Z_\infty \) almost surely.

Therefore, \( Y_\infty = Z_\infty \) almost surely.

Final Answer: The final answer is $\boxed{Y_\infty = Z_\infty}$
}
\\
\bottomrule
\end{longtable}

\newpage

\begin{longtable}{@{}p{\columnwidth}@{}}
\caption{{\textbf{\data{} sample stemming from s1-teasers.} The black text is the question and the \textcolor{olmoeDarkYellow}{yellow} text is the solution. Based solely on the question we generate the \textcolor{defaultlightblue}{light blue} reasoning trace and \textcolor{defaultblue}{blue} response with Gemini, which we use for training.}} \\
\toprule

Alice and Bob each have a coin and flip it until they get a heads. If Bob flipped his coin more times than Alice, what is the expected number of times Alice flipped her coin?

\\

\defaultyellow{4/3}

\\

\defaultlightblue{
The problem asks for the expected number of times Alice flipped her coin, given that Bob flipped his coin more times than Alice.

Let $X$ be the number of flips Alice makes until she gets a head.

Let $Y$ be the number of flips Bob makes until he gets a head.

$X$ and $Y$ are independent random variables following a geometric distribution with parameter $p = 0.5$ (probability of getting a head).

The probability mass function for a geometric distribution is $P(K=k) = (1-p)^{k-1} p$, for $k=1, 2, \dots$.

In this case, $p = 0.5$, so $P(K=k) = (0.5)^{k-1} (0.5) = (0.5)^k = \frac{1}{2^k}$.

We are given the condition that Bob flipped his coin more times than Alice, which means $Y > X$.

We want to find the expected value of $X$ given that $Y > X$, i.e., $E[X | Y > X]$.

The formula for conditional expectation is $E[X | A] = \sum_x x P(X=x | A)$.

Also, $P(X=x | A) = \frac{P(X=x \cap A)}{P(A)}$.

In our case, $A$ is the event $Y > X$.

We need to find $P(Y > X)$.

$P(Y > X) = \sum_{x=1}^\infty \sum_{y=x+1}^\infty P(X=x, Y=y)$

Since $X$ and $Y$ are independent, $P(X=x, Y=y) = P(X=x) P(Y=y) = \frac{1}{2^x} \frac{1}{2^y} = \frac{1}{2^{x+y}}$.

$P(Y > X) = \sum_{x=1}^\infty \sum_{y=x+1}^\infty \frac{1}{2^{x+y}} = \sum_{x=1}^\infty \frac{1}{2^x} \sum_{y=x+1}^\infty \frac{1}{2^y}$

The inner sum is a geometric series: $\sum_{y=x+1}^\infty (\frac{1}{2})^y = (\frac{1}{2})^{x+1} + (\frac{1}{2})^{x+2} + \dots$

This is a geometric series with first term $a = (\frac{1}{2})^{x+1}$ and common ratio $r = \frac{1}{2}$.

The sum is $\frac{a}{1-r} = \frac{(\frac{1}{2})^{x+1}}{1 - \frac{1}{2}} = \frac{(\frac{1}{2})^{x+1}}{\frac{1}{2}} = (\frac{1}{2})^x$.

So, $P(Y > X) = \sum_{x=1}^\infty \frac{1}{2^x} (\frac{1}{2})^x = \sum_{x=1}^\infty (\frac{1}{4})^x$.

This is another geometric series with first term $a = \frac{1}{4}$ and common ratio $r = \frac{1}{4}$.

The sum is $\frac{a}{1-r} = \frac{\frac{1}{4}}{1 - \frac{1}{4}} = \frac{\frac{1}{4}}{\frac{3}{4}} = \frac{1}{3}$.

Now, we need to find $P(X=x \cap Y > X) = P(X=x, Y > x)$.

$P(X=x, Y > x) = P(X=x) P(Y > x)$, due to independence.

$P(Y > x) = \sum_{y=x+1}^\infty P(Y=y) = \sum_{y=x+1}^\infty \frac{1}{2^y} = (\frac{1}{2})^{x+1} + (\frac{1}{2})^{x+2} + \dots = (\frac{1}{2})^x$.

So, $P(X=x, Y > x) = \frac{1}{2^x} \times (\frac{1}{2})^x = (\frac{1}{4})^x$.

Now, we can find $P(X=x | Y > X) = \frac{P(X=x \cap Y > X)}{P(Y > X)} = \frac{(\frac{1}{4})^x}{\frac{1}{3}} = 3 (\frac{1}{4})^x$.

Finally, we can calculate the conditional expectation:

$E[X | Y > X] = \sum_{x=1}^\infty x P(X=x | Y > X) = \sum_{x=1}^\infty x \cdot 3 (\frac{1}{4})^x = 3 \sum_{x=1}^\infty x (\frac{1}{4})^x$.

Consider the series $S = \sum_{x=1}^\infty x r^x$, where $|r| < 1$.

$S = r + 2r^2 + 3r^3 + 4r^4 + \dots$

$\frac{S}{r} = 1 + 2r + 3r^2 + 4r^3 + \dots$

Consider the geometric series $G = 1 + r + r^2 + r^3 + \dots = \frac{1}{1-r}$.

Differentiating with respect to $r$:

$\frac{dG}{dr} = 1 + 2r + 3r^2 + 4r^3 + \dots = \frac{d}{dr} (\frac{1}{1-r}) = \frac{1}{(1-r)^2}$.

So, $\frac{S}{r} = \frac{1}{(1-r)^2}$, which means $S = \frac{r}{(1-r)^2}$.

In our case, $r = \frac{1}{4}$.
}
\\
\defaultlightblue{
$\sum_{x=1}^\infty x (\frac{1}{4})^x = \frac{\frac{1}{4}}{(1 - \frac{1}{4})^2} = \frac{\frac{1}{4}}{(\frac{3}{4})^2} = \frac{\frac{1}{4}}{\frac{9}{16}} = \frac{1}{4} \times \frac{16}{9} = \frac{4}{9}$.

Therefore, $E[X | Y > X] = 3 \times \frac{4}{9} = \frac{12}{9} = \frac{4}{3}$.

Alternative approach: Symmetry.

Consider the events $Y > X$, $X > Y$, and $X = Y$.

$P(Y > X) = P(X > Y)$ due to symmetry.

$P(X = Y) = \sum_{k=1}^\infty P(X=k, Y=k) = \sum_{k=1}^\infty P(X=k) P(Y=k) = \sum_{k=1}^\infty (\frac{1}{2^k})^2 = \sum_{k=1}^\infty (\frac{1}{4})^k = \frac{\frac{1}{4}}{1 - \frac{1}{4}} = \frac{\frac{1}{4}}{\frac{3}{4}} = \frac{1}{3}$.

We have $P(Y > X) + P(X > Y) + P(X = Y) = 1$.

$P(Y > X) + P(Y > X) + \frac{1}{3} = 1$.

$2 P(Y > X) = 1 - \frac{1}{3} = \frac{2}{3}$.

$P(Y > X) = \frac{1}{3}$. This confirms our previous calculation.

Let $E[X | Y > X] = \mu$.

We are looking for the expected number of flips Alice made, given Bob flipped more times.

Consider the case where the first flip for both is tails. Then the process restarts, and the condition remains.

Let $A_1$ be Alice's first flip, $B_1$ be Bob's first flip.

$P(A_1 = H) = 1/2, P(A_1 = T) = 1/2$.

$P(B_1 = H) = 1/2, P(B_1 = T) = 1/2$.

Case 1: $A_1 = H, B_1 = H$. $X=1, Y=1$, so $Y > X$ is false.

Case 2: $A_1 = H, B_1 = T$. Alice stops at 1, Bob continues. $Y > 1$.

Case 3: $A_1 = T, B_1 = H$. Bob stops at 1, Alice continues.

Case 4: $A_1 = T, B_1 = T$. Both continue.

Consider the conditional probabilities given $Y > X$.

Let $E = E[X | Y > X]$.

If the first flips are HH, the condition $Y > X$ is not met.

If the first flips are HT, Alice's number of flips is 1.

If the first flips are TH, Bob's number of flips is 1, Alice's is at least 2, so $Y < X$.

If the first flips are TT, the problem restarts, and the expected number of additional flips for Alice is still $E$.

Let $N_A$ be the number of flips Alice makes, $N_B$ be the number of flips Bob makes.

We are given $N_B > N_A$.

Consider the first flips:

HH (prob 1/4): Condition not met.

HT (prob 1/4): Alice flips 1 time.

TH (prob 1/4): Bob flips 1 time, Alice > 1, so $N_A > N_B$, condition not met.

TT (prob 1/4): Both continue.

Let $E$ be the expected number of Alice's flips given Bob flipped more.

If the first flips are HT, Alice flipped 1 time, and the condition is met.

If the first flips are TT, the situation is the same as the beginning, so the expected number of additional flips for Alice is $E$.

Consider the event $Y > X$.

Conditioning on the first flips:

$E[X | Y > X] = E[X | Y > X, HH] P(HH | Y > X) + E[X | Y > X, HT] P(HT | Y > X) + E[X | Y > X, TH] P(TH | Y > X) + E[X | Y > X, TT] P(TT | Y > X)$

$P(Y > X | HH) = 0$.

$P(Y > X | HT) = 1$.

$P(Y > X | TH) = 0$.

$P(Y > X | TT)$ depends on the future flips.

Let's use the definition of conditional expectation directly.

$E[X | Y > X] = \sum_{x=1}^\infty x P(X=x | Y > X) = \sum_{x=1}^\infty x \frac{P(X=x, Y > x)}{P(Y > X)}$

Consider the probabilities of the first flips given $Y > X$.

$P(HH | Y > X) = \frac{P(Y > X | HH) P(HH)}{P(Y > X)} = \frac{0 \cdot 1/4}{1/3} = 0$.

$P(HT | Y > X) = \frac{P(Y > X | HT) P(HT)}{P(Y > X)} = \frac{1 \cdot 1/4}{1/3} = \frac{3}{4}$.
}
\\
\defaultlightblue{
$P(TH | Y > X) = \frac{P(Y > X | TH) P(TH)}{P(Y > X)} = \frac{0 \cdot 1/4}{1/3} = 0$.

$P(TT | Y > X) = \frac{P(Y > X | TT) P(TT)}{P(Y > X)}$

If the first flips are TT, the future behavior is the same, and the condition $Y > X$ holds if Bob takes more flips than Alice in the subsequent trials.

$P(Y > X | TT) = P(\text{Bob needs more flips than Alice starting from scratch}) = P(Y > X) = 1/3$.

$P(TT | Y > X) = \frac{(1/3) \cdot (1/4)}{1/3} = \frac{1}{4}$.

Sum of conditional probabilities: $0 + 3/4 + 0 + 1/4 = 1$.

Now consider the conditional expectations of $X$:

$E[X | Y > X, HH]$: Not relevant since $P(HH | Y > X) = 0$.

$E[X | Y > X, HT]$: Given HT and $Y > X$, Alice flipped 1 time. $E = 1$.

$E[X | Y > X, TH]$: Not relevant.

$E[X | Y > X, TT]$: Given TT and $Y > X$, Alice has already flipped once, and the expected number of additional flips is $E$. So, the total number of flips is $1 + E$.

$E = 0 \cdot 0 + 1 \cdot (3/4) + \text{undefined} \cdot 0 + (1 + E) \cdot (1/4)$

$E = 0 + 3/4 + 0 + 1/4 + E/4$

$E = 1 + E/4$

$E - E/4 = 1$

$3E/4 = 1$

$E = 4/3$.

This confirms the previous result.

Let's think about the intuition. If we have the condition $Y > X$, it means Alice was "lucky" and got a head early, or Bob was "unlucky".

Consider the random variable $Z = Y - X$. We are given $Z > 0$.

We want $E[X | Z > 0]$.

Let $p = P(\text{head}) = 0.5$.

The number of tails before the first head follows a geometric distribution with parameter $p$.

$X - 1 \sim Geom(p)$, $Y - 1 \sim Geom(p)$.

$P(X-1 = k) = (1-p)^k p$.

$P(X = x) = (1-p)^{x-1} p$.

Consider the joint distribution of $X$ and $Y$.

$P(X=x, Y=y) = p^2 (1-p)^{x-1} (1-p)^{y-1}$

Consider the event $Y = X + k$, where $k \ge 1$.

$P(Y = X + k) = \sum_{x=1}^\infty P(X=x, Y=x+k) = \sum_{x=1}^\infty p^2 (1-p)^{x-1} (1-p)^{x+k-1}$

$= p^2 (1-p)^{k-1} \sum_{x=1}^\infty ((1-p)^2)^{x-1} = p^2 (1-p)^{k-1} \frac{1}{1 - (1-p)^2}$

$= p^2 (1-p)^{k-1} \frac{1}{1 - (1 - 2p + p^2)} = p^2 (1-p)^{k-1} \frac{1}{2p - p^2} = p^2 (1-p)^{k-1} \frac{1}{p(2-p)}$

$= \frac{p (1-p)^{k-1}}{2-p}$.

If $p = 0.5$, then $1-p = 0.5$, $2-p = 1.5$.

$P(Y = X + k) = \frac{0.5 (0.5)^{k-1}}{1.5} = \frac{(0.5)^k}{1.5} = \frac{(1/2)^k}{3/2} = \frac{1}{2^k} \cdot \frac{2}{3} = \frac{1}{3 \cdot 2^{k-1}}$.

Check if $\sum_{k=1}^\infty P(Y = X + k) = P(Y > X) = 1/3$.

$\sum_{k=1}^\infty \frac{1}{3 \cdot 2^{k-1}} = \frac{1}{3} \sum_{k=1}^\infty (\frac{1}{2})^{k-1} = \frac{1}{3} (1 + 1/2 + 1/4 + \dots) = \frac{1}{3} \frac{1}{1 - 1/2} = \frac{1}{3} \cdot 2 = \frac{2}{3}$.

Something is wrong here.

Let's recalculate $P(Y > X)$.

$P(Y > X) = \sum_{x=1}^\infty \sum_{y=x+1}^\infty (1-p)^{x-1} p (1-p)^{y-1} p$

$= p^2 \sum_{x=1}^\infty (1-p)^{x-1} \sum_{y=x+1}^\infty (1-p)^{y-1}$

Inner sum: $(1-p)^x + (1-p)^{x+1} + \dots = \frac{(1-p)^x}{1 - (1-p)} = \frac{(1-p)^x}{p}$.

$P(Y > X) = p^2 \sum_{x=1}^\infty (1-p)^{x-1} \frac{(1-p)^x}{p} = p (1-p) \sum_{x=1}^\infty (1-p)^{2(x-1)}$

$= p (1-p) \sum_{x=1}^\infty ((1-p)^2)^{x-1} = p (1-p) \frac{1}{1 - (1-p)^2}$

$= p (1-p) \frac{1}{1 - (1 - 2p + p^2)} = p (1-p) \frac{1}{2p - p^2} = p (1-p) \frac{1}{p(2-p)} = \frac{1-p}{2-p}$.

If $p = 0.5$, $P(Y > X) = \frac{0.5}{1.5} = \frac{1}{3}$. Correct.
}
\\
\defaultlightblue{
Now, let's find $E[X | Y > X]$.

$E[X | Y > X] = \sum_{x=1}^\infty x P(X=x | Y > X) = \sum_{x=1}^\infty x \frac{P(X=x, Y > x)}{P(Y > X)}$.

$P(X=x, Y > x) = P(X=x) P(Y > x)$ (independence)

$P(X=x) = (1-p)^{x-1} p$.

$P(Y > x) = \sum_{y=x+1}^\infty (1-p)^{y-1} p = p [(1-p)^x + (1-p)^{x+1} + \dots] = p \frac{(1-p)^x}{1 - (1-p)} = (1-p)^x$.

$P(X=x, Y > x) = (1-p)^{x-1} p (1-p)^x = p (1-p)^{2x-1}$.

$E[X | Y > X] = \frac{1}{P(Y > X)} \sum_{x=1}^\infty x p (1-p)^{2x-1}$

$= \frac{2-p}{1-p} p (1-p)^{-1} \sum_{x=1}^\infty x ((1-p)^2)^x$ (error in power)

$P(Y > x) = \frac{(1-p)^x}{1 - (1-p)} p = (1-p)^x$.

$E[X | Y > X] = \frac{1}{P(Y > X)} \sum_{x=1}^\infty x P(X=x) P(Y > x)$

$= \frac{1}{P(Y > X)} \sum_{x=1}^\infty x (1-p)^{x-1} p (1-p)^x$

$= \frac{p}{P(Y > X)} \sum_{x=1}^\infty x (1-p)^{2x-1}$

$= \frac{p}{(1-p)/(2-p)} (1-p)^{-1} \sum_{x=1}^\infty x ((1-p)^2)^x$

$= p \frac{2-p}{(1-p)^2} \sum_{x=1}^\infty x ((1-p)^2)^x$.

Let $r = (1-p)^2$. $\sum_{x=1}^\infty x r^x = \frac{r}{(1-r)^2}$.

$1-r = 1 - (1-p)^2 = 1 - (1 - 2p + p^2) = 2p - p^2 = p(2-p)$.

$(1-r)^2 = p^2 (2-p)^2$.

$\sum_{x=1}^\infty x ((1-p)^2)^x = \frac{(1-p)^2}{p^2 (2-p)^2}$.

$E[X | Y > X] = p \frac{2-p}{(1-p)^2} \frac{(1-p)^2}{p^2 (2-p)^2} = \frac{1}{p(2-p)}$.

If $p = 0.5$, $E[X | Y > X] = \frac{1}{0.5 \times 1.5} = \frac{1}{0.75} = \frac{1}{3/4} = 4/3$.
}

\\

\defaultblue{
Solution:

Let $X$ be the number of flips Alice makes until she gets a head, and $Y$ be the number of flips Bob makes until he gets a head. $X$ and $Y$ are independent geometric random variables with parameter $p = 0.5$. The probability mass function is $P(K=k) = (1-p)^{k-1} p$.

We are interested in the expected number of times Alice flipped her coin, given that Bob flipped his coin more times than Alice, i.e., $E[X | Y > X]$.

Using the definition of conditional expectation, $E[X | Y > X] = \sum_{x=1}^\infty x P(X=x | Y > X)$.

$P(X=x | Y > X) = \frac{P(X=x, Y > x)}{P(Y > X)}$.

First, calculate $P(Y > X)$.

$P(Y > X) = \sum_{x=1}^\infty \sum_{y=x+1}^\infty P(X=x, Y=y) = \sum_{x=1}^\infty P(X=x) P(Y > x)$.

$P(X=x) = (0.5)^{x-1} (0.5) = (0.5)^x = \frac{1}{2^x}$.

$P(Y > x) = \sum_{y=x+1}^\infty (0.5)^{y-1} (0.5) = (0.5)^x$.

$P(Y > X) = \sum_{x=1}^\infty \frac{1}{2^x} (\frac{1}{2})^x = \sum_{x=1}^\infty (\frac{1}{4})^x = \frac{1/4}{1 - 1/4} = \frac{1/4}{3/4} = \frac{1}{3}$.

Next, calculate $P(X=x, Y > x) = P(X=x) P(Y > x) = \frac{1}{2^x} (\frac{1}{2})^x = (\frac{1}{4})^x$.

Now, find $P(X=x | Y > X) = \frac{P(X=x, Y > x)}{P(Y > X)} = \frac{(1/4)^x}{1/3} = 3 (\frac{1}{4})^x$.

Finally, calculate the conditional expectation:

$E[X | Y > X] = \sum_{x=1}^\infty x P(X=x | Y > X) = \sum_{x=1}^\infty x \cdot 3 (\frac{1}{4})^x = 3 \sum_{x=1}^\infty x (\frac{1}{4})^x$.

Using the formula $\sum_{x=1}^\infty x r^x = \frac{r}{(1-r)^2}$ with $r = 1/4$:

$\sum_{x=1}^\infty x (\frac{1}{4})^x = \frac{1/4}{(1 - 1/4)^2} = \frac{1/4}{(3/4)^2} = \frac{1/4}{9/16} = \frac{1}{4} \times \frac{16}{9} = \frac{4}{9}$.

$E[X | Y > X] = 3 \times \frac{4}{9} = \frac{12}{9} = \frac{4}{3}$.

Final Answer: The final answer is $\boxed{4/3}$
}
\\
\bottomrule
\label{tab:teasers}
\end{longtable}

\newpage

\begin{longtable}{@{}p{\columnwidth}@{}}
\caption{{\textbf{\data{} sample stemming from the astronomy subset of OlympicArena.} The black text is the question and the \textcolor{olmoeDarkYellow}{blue} text is the solution. Based solely on the question we generate the \textcolor{defaultlightblue}{light blue} reasoning trace and \textcolor{defaultblue}{blue} response with Gemini, which we use for training.}} \\
\toprule

An often-repeated fun fact is that humans produce more power per unit volume than stars. If the sun were the same size, but it produced the same amount of power per unit volume as a human, what would its surface temperature be? Assume the "average human" produces 100 watts of power and has a volume of 66400 cubic centimeters.

A: $3500 \mathrm{~K}$

B: $10000 \mathrm{~K}$

C: $25000 \mathrm{~K}$

D: $40000 \mathrm{~K}$

E: $50000 \mathrm{~K}$

\\

\defaultyellow{
Using the numbers from the problem, the average human produces

$$u=\frac{100}{66400 \times 10^{-6}}=1506 \mathrm{~W} / \mathrm{m}^{3}$$

The volume of the sun is $\frac{4}{3} \pi R_{\odot}^{3}$, so its new power output would be $P=\frac{4}{3} \pi R_{\odot}^{3} u$. To find the equilibrium temperature, we use

$$P=\sigma A T^{4}=4 \pi \sigma R_{\odot}^{2} T^{4}$$

Solving for $T$, we get

$$T=\sqrt[4]{\frac{R_{\odot} u}{3 \sigma}}=\sqrt[4]{\frac{\left(6.96 \times 10^{8}\right)(1506)}{3\left(5.67 \times 10^{-8}\right)}}=49823 \approx 50000 \mathrm{~K} .$$
}

\\

\defaultlightblue{
The problem asks for the surface temperature of the sun if it had the same size but produced power per unit volume as a human.

First, let's find the power density of a human.

Power produced by a human, $P_{human} = 100 \mathrm{~W}$

Volume of a human, $V_{human} = 66400 \mathrm{~cm}^3$

Convert the volume to cubic meters:

$1 \mathrm{~m} = 100 \mathrm{~cm}$

$1 \mathrm{~m}^3 = (100 \mathrm{~cm})^3 = 10^6 \mathrm{~cm}^3$

$V_{human} = 66400 \mathrm{~cm}^3 = 66400 \times 10^{-6} \mathrm{~m}^3 = 0.0664 \mathrm{~m}^3$

Power density of a human, $\rho_{power, human} = \frac{P_{human}}{V_{human}} = \frac{100 \mathrm{~W}}{0.0664 \mathrm{~m}^3}$

$\rho_{power, human} = 1506.024 \mathrm{~W/m^3}$

Now, consider the sun with the same size as it currently is, but with a power density equal to that of a human.

We need the radius of the sun to calculate its volume. The average radius of the sun is $R_{sun} = 6.95 \times 10^8 \mathrm{~m}$.

Volume of the sun, $V_{sun} = \frac{4}{3} \pi R_{sun}^3 = \frac{4}{3} \pi (6.95 \times 10^8 \mathrm{~m})^3$

$V_{sun} = \frac{4}{3} \pi (337.159375 \times 10^{24}) \mathrm{~m}^3$

$V_{sun} = 1410.24 \times 10^{24} \times \pi \mathrm{~m}^3$

$V_{sun} = 4.4307 \times 10^{27} \mathrm{~m}^3$

If the sun had the same power density as a human, the total power produced by this modified sun would be:

$P_{modified\_sun} = \rho_{power, human} \times V_{sun}$

$P_{modified\_sun} = 1506.024 \mathrm{~W/m^3} \times 4.4307 \times 10^{27} \mathrm{~m}^3$

$P_{modified\_sun} = 6.6726 \times 10^{30} \mathrm{~W}$
}
\\
\defaultlightblue{
The surface temperature of a star is related to its luminosity (total power output) through the Stefan-Boltzmann law:

$L = 4 \pi R^2 \sigma T^4$

where $L$ is the luminosity (total power output), $R$ is the radius of the star, $\sigma$ is the Stefan-Boltzmann constant, and $T$ is the surface temperature.

In this case, the luminosity of the modified sun is $P_{modified\_sun}$. The radius of the modified sun is the same as the actual sun, $R_{sun}$.

The Stefan-Boltzmann constant, $\sigma = 5.67 \times 10^{-8} \mathrm{~W m^{-2} K^{-4}}$.

$P_{modified\_sun} = 4 \pi R_{sun}^2 \sigma T^4$

$6.6726 \times 10^{30} \mathrm{~W} = 4 \pi (6.95 \times 10^8 \mathrm{~m})^2 \times 5.67 \times 10^{-8} \mathrm{~W m^{-2} K^{-4}} \times T^4$

Calculate the term $4 \pi R_{sun}^2 \sigma$:

$R_{sun}^2 = (6.95 \times 10^8)^2 = 48.3025 \times 10^{16} \mathrm{~m}^2$

$4 \pi R_{sun}^2 = 4 \pi \times 48.3025 \times 10^{16} = 607.08 \times 10^{16} \mathrm{~m}^2$

$4 \pi R_{sun}^2 \sigma = 607.08 \times 10^{16} \mathrm{~m}^2 \times 5.67 \times 10^{-8} \mathrm{~W m^{-2} K^{-4}}$

$4 \pi R_{sun}^2 \sigma = 3441.93 \times 10^8 \mathrm{~W K^{-4}} = 3.44193 \times 10^{11} \mathrm{~W K^{-4}}$

Now, substitute this back into the equation:

$6.6726 \times 10^{30} = 3.44193 \times 10^{11} \times T^4$

Solve for $T^4$:

$T^4 = \frac{6.6726 \times 10^{30}}{3.44193 \times 10^{11}} = 1.9386 \times 10^{19}$

Solve for $T$:

$T = (1.9386 \times 10^{19})^{1/4} = (193.86 \times 10^{17})^{1/4}$

$T = (1.9386)^{1/4} \times (10^{19})^{1/4} = (1.9386)^{0.25} \times 10^{4.75}$

Calculate $(1.9386)^{0.25}$:

Let $y = 1.9386^{0.25}$, then $\ln y = 0.25 \ln(1.9386)$

$\ln(1.9386) \approx 0.6619$

$\ln y = 0.25 \times 0.6619 = 0.165475$

$y = e^{0.165475} \approx 1.1799$

So, $T = 1.1799 \times 10^{4.75} = 1.1799 \times 10^{4} \times 10^{0.75}$

$10^{0.75} = 10^{3/4} = (10^3)^{1/4} = (1000)^{1/4}$

We know $5^4 = 625$ and $6^4 = 1296$, so $(1000)^{1/4}$ is between 5 and 6.

Alternatively, $10^{0.75} = 10^{0.5} \times 10^{0.25} = \sqrt{10} \times 10^{0.25} \approx 3.16 \times 1.778 = 5.62$

$T = 1.1799 \times 10^4 \times 5.62 = 6.63 \times 10^4 \mathrm{~K} = 66300 \mathrm{~K}$

Let's double-check the calculations.

Power density of human: $\rho_{power, human} = 1506.02 \mathrm{~W/m^3}$

Volume of sun: $V_{sun} = 4.43 \times 10^{27} \mathrm{~m}^3$

Power of modified sun: $P_{modified\_sun} = 1506.02 \times 4.43 \times 10^{27} = 6671.67 \times 10^{27} = 6.67 \times 10^{30} \mathrm{~W}$

Stefan-Boltzmann law: $P = 4 \pi R^2 \sigma T^4$

$T^4 = \frac{P}{4 \pi R^2 \sigma}$

$4 \pi R_{sun}^2 \sigma = 3.44 \times 10^{11}$

$T^4 = \frac{6.67 \times 10^{30}}{3.44 \times 10^{11}} = 1.939 \times 10^{19}$

$T = (1.939 \times 10^{19})^{1/4} = (193.9 \times 10^{17})^{1/4}$

$T = (193.9)^{0.25} \times (10^{17})^{0.25} = (193.9)^{0.25} \times 10^{4.25}$

Calculate $(193.9)^{0.25}$:

Let $y = 193.9^{0.25}$, then $\ln y = 0.25 \ln(193.9)$

$\ln(193.9) \approx 5.267$

$\ln y = 0.25 \times 5.267 = 1.31675$

$y = e^{1.31675} \approx 3.73$

$T = 3.73 \times 10^{4.25} = 3.73 \times 10^4 \times 10^{0.25} = 37300 \times 1.778 = 66329 \mathrm{~K}$

The result is around $66000 \mathrm{~K}$. This is not among the options. Let's review the assumptions and calculations.

Assumption: The sun radiates as a black body, which is reasonable for estimating surface temperature.
}
\\
\defaultlightblue{
Consider potential errors in calculations.

Let's try a simpler approach by setting up ratios.

The power per unit volume of the modified sun is equal to that of a human.

$\frac{P_{modified\_sun}}{V_{sun}} = \frac{P_{human}}{V_{human}}$

We also know $P_{modified\_sun} = 4 \pi R_{sun}^2 \sigma T_{modified\_sun}^4$.

$\frac{4 \pi R_{sun}^2 \sigma T_{modified\_sun}^4}{\frac{4}{3} \pi R_{sun}^3} = \frac{P_{human}}{V_{human}}$

$\frac{3 \sigma T_{modified\_sun}^4}{R_{sun}} = \rho_{power, human}$

$T_{modified\_sun}^4 = \frac{\rho_{power, human} R_{sun}}{3 \sigma}$

Substitute the values:

$\rho_{power, human} = 1506.02 \mathrm{~W/m^3}$

$R_{sun} = 6.95 \times 10^8 \mathrm{~m}$

$\sigma = 5.67 \times 10^{-8} \mathrm{~W m^{-2} K^{-4}}$

$T_{modified\_sun}^4 = \frac{1506.02 \times 6.95 \times 10^8}{3 \times 5.67 \times 10^{-8}}$

$T_{modified\_sun}^4 = \frac{10466.839 \times 10^8}{17.01 \times 10^{-8}} = \frac{1.0466 \times 10^{12}}{1.701 \times 10^{-7}} = 0.6153 \times 10^{19} = 6.153 \times 10^{18}$

$T_{modified\_sun} = (6.153 \times 10^{18})^{1/4} = (6153 \times 10^{15})^{1/4}$

$T_{modified\_sun} = (6153)^{0.25} \times (10^{15})^{0.25} = (6153)^{0.25} \times 10^{3.75}$

Calculate $(6153)^{0.25}$:

Let $y = 6153^{0.25}$, then $\ln y = 0.25 \ln(6153)$

$\ln(6153) \approx 8.724$

$\ln y = 0.25 \times 8.724 = 2.181$

$y = e^{2.181} \approx 8.85$

$T_{modified\_sun} = 8.85 \times 10^{3.75} = 8.85 \times 10^3 \times 10^{0.75} = 8850 \times 5.62 = 49737 \mathrm{~K}$

This is close to option E: $50000 \mathrm{~K}$.

Let's re-check the units and calculations carefully.

Power density of human: $\frac{100 \mathrm{~W}}{0.0664 \mathrm{~m}^3} = 1506.02 \mathrm{~W/m^3}$

The formula derived was $T^4 = \frac{\rho_{power, human} R_{sun}}{3 \sigma}$.

Units check:

$[\rho_{power, human}] = \mathrm{W/m^3}$

$[R_{sun}] = \mathrm{m}$

$[\sigma] = \mathrm{W m^{-2} K^{-4}}$

$\frac{(\mathrm{W/m^3}) \times \mathrm{m}}{\mathrm{W m^{-2} K^{-4}}} = \frac{\mathrm{W/m^2}}{\mathrm{W m^{-2} K^{-4}}} = \frac{\mathrm{W}}{\mathrm{m^2}} \times \frac{\mathrm{m^2} \mathrm{K^4}}{\mathrm{W}} = \mathrm{K^4}$

The units match for $T^4$.

Numerical calculation check:

Numerator: $1506.02 \times 6.95 \times 10^8 = 10466.839 \times 10^8 = 1.0466 \times 10^{12}$

Denominator: $3 \times 5.67 \times 10^{-8} = 17.01 \times 10^{-8}$

Ratio: $\frac{1.0466 \times 10^{12}}{17.01 \times 10^{-8}} = 0.061528 \times 10^{20} = 6.1528 \times 10^{18}$

$T^4 = 6.1528 \times 10^{18}$

$T = (6.1528 \times 10^{18})^{1/4} = (6152.8 \times 10^{15})^{1/4}$

$T = (6152.8)^{0.25} \times 10^{3.75}$

Let's use logarithms for precision.

$\log_{10}(6152.8) \approx 3.789$

$0.25 \times 3.789 = 0.94725$

$10^{0.94725} \approx 8.856$

$T = 8.856 \times 10^{3.75} = 8.856 \times 10^3 \times 10^{0.75} = 8856 \times 5.623 = 49806 \mathrm{~K}$

This is very close to $50000 \mathrm{~K}$.

Possible sources of minor discrepancies: rounding errors, precision of constants.

Let's verify the formula derivation.

Power per unit volume = $\frac{\text{Total Power}}{\text{Volume}}$
}
\\
\defaultlightblue{
$\rho_{power} = \frac{L}{V}$

$L = 4 \pi R^2 \sigma T^4$

$V = \frac{4}{3} \pi R^3$

$\rho_{power} = \frac{4 \pi R^2 \sigma T^4}{\frac{4}{3} \pi R^3} = \frac{3 \sigma T^4}{R}$

If the modified sun has the power density of a human:

$\rho_{power, human} = \frac{3 \sigma T_{modified\_sun}^4}{R_{sun}}$

$T_{modified\_sun}^4 = \frac{\rho_{power, human} R_{sun}}{3 \sigma}$

The formula derivation seems correct.

Let's use approximate values to check the order of magnitude.

$\rho_{power, human} \approx 10^3 \mathrm{~W/m^3}$

$R_{sun} \approx 7 \times 10^8 \mathrm{~m}$

$\sigma \approx 6 \times 10^{-8} \mathrm{~W m^{-2} K^{-4}}$

$T^4 \approx \frac{10^3 \times 7 \times 10^8}{3 \times 6 \times 10^{-8}} = \frac{7 \times 10^{11}}{18 \times 10^{-8}} = 0.38 \times 10^{19} = 3.8 \times 10^{18}$

$T \approx (3.8 \times 10^{18})^{1/4} = (3800 \times 10^{15})^{1/4}$

$T \approx (3800)^{0.25} \times 10^{3.75}$

$(3800)^{0.25}$: $7^4 = 2401$, $8^4 = 4096$. So, between 7 and 8.

$\approx 7.8$

$T \approx 7.8 \times 10^{3.75} = 7800 \times 5.6 = 43680 \mathrm{~K}$

The order of magnitude seems consistent with the detailed calculation.

The problem statement is straightforward and doesn't seem to have any hidden complexities. The assumptions made are standard for such problems.

Consider the precision of the given values.

Power of human: 100 W (exact)

Volume of human: 66400 cm$^3$ (exact)

The result obtained is very close to option E. Given the nature of multiple-choice questions, the closest option is likely the intended answer.

If there were any misinterpretations of the question, it would likely be in the setup. However, the phrase "If the sun were the same size, but it produced the same amount of power per unit volume as a human" seems clear.

The surface temperature is calculated using the Stefan-Boltzmann law, assuming black-body radiation.

Double-check the provided options:

A: $3500 \mathrm{~K}$

B: $10000 \mathrm{~K}$

C: $25000 \mathrm{~K}$

D: $40000 \mathrm{~K}$

E: $50000 \mathrm{~K}$

Our calculated value is approximately $49800 \mathrm{~K}$, which is closest to $50000 \mathrm{~K}$.

If we used more precise values for the constants, the result might be slightly different, but likely not enough to shift to another option.

The fun fact mentioned in the question implies that the power density of a human is higher than that of the sun. Let's verify this.

Power output of the sun: $3.846 \times 10^{26} \mathrm{~W}$

Volume of the sun: $4.43 \times 10^{27} \mathrm{~m}^3$

Power density of the sun: $\frac{3.846 \times 10^{26}}{4.43 \times 10^{27}} = 0.868 \mathrm{~W/m^3}$

Power density of human: $1506 \mathrm{~W/m^3}$

Indeed, the power density of a human is significantly higher than that of the sun.

Final check of the calculation steps and logic. The approach of equating the power density of the modified sun to that of a human and then using the Stefan-Boltzmann law seems correct.
}
\\
\defaultblue{
Solution:

The power produced by an average human is $P_{human} = 100 \mathrm{~W}$, and the volume of an average human is $V_{human} = 66400 \mathrm{~cm}^3 = 0.0664 \mathrm{~m}^3$.

The power density of a human is $\rho_{power, human} = \frac{P_{human}}{V_{human}} = \frac{100 \mathrm{~W}}{0.0664 \mathrm{~m}^3} \approx 1506.02 \mathrm{~W/m^3}$.

If the sun were the same size but produced the same amount of power per unit volume as a human, its power density would be $\rho_{power, modified\_sun} = \rho_{power, human}$.

The radius of the sun is $R_{sun} = 6.95 \times 10^8 \mathrm{~m}$.

The volume of the sun is $V_{sun} = \frac{4}{3} \pi R_{sun}^3$.

The power density of the modified sun is given by $\rho_{power, modified\_sun} = \frac{P_{modified\_sun}}{V_{sun}}$.

We also know that $P_{modified\_sun} = 4 \pi R_{sun}^2 \sigma T^4$, where $T$ is the surface temperature of the modified sun.

Equating the power density:

$\rho_{power, human} = \frac{4 \pi R_{sun}^2 \sigma T^4}{\frac{4}{3} \pi R_{sun}^3} = \frac{3 \sigma T^4}{R_{sun}}$

Solving for $T^4$:

$T^4 = \frac{\rho_{power, human} R_{sun}}{3 \sigma}$

Substitute the values:

$T^4 = \frac{1506.02 \mathrm{~W/m^3} \times 6.95 \times 10^8 \mathrm{~m}}{3 \times 5.67 \times 10^{-8} \mathrm{~W m^{-2} K^{-4}}}$

$T^4 = \frac{1.0466 \times 10^{12}}{1.701 \times 10^{-7}} = 6.1528 \times 10^{18} \mathrm{~K^4}$

Solving for $T$:

$T = (6.1528 \times 10^{18})^{1/4} = (6152.8)^{0.25} \times 10^{18/4} = (6152.8)^{0.25} \times 10^{4.5}$

Using logarithms:

$\log_{10}(6152.8) \approx 3.789$

$0.25 \times 3.789 = 0.94725$

$(6152.8)^{0.25} = 10^{0.94725} \approx 8.856$

$T = 8.856 \times 10^{4.5} = 8.856 \times 10^4 \times 10^{0.5} = 88560 \times \sqrt{10} \approx 88560 \times 3.16 = 280000 \mathrm{~K}$

Let's redo the exponent calculation: $18/4 = 4.5$.

$T = 8.856 \times 10^{4.5} = 8.856 \times 10^4 \times 10^{0.5}$ - error in previous step.

$T = (6.1528 \times 10^{18})^{0.25} = 49806 \mathrm{~K}$

The surface temperature would be approximately $49806 \mathrm{~K}$, which is closest to $50000 \mathrm{~K}$.

Final Answer: The final answer is $\boxed{E}$
}
\\
\bottomrule
\label{tab:arena}
\end{longtable}

\FloatBarrier

\newpage
\clearpage
\section{Test-time scaling details}

\subsection{Sequential scaling ablations}
\label{sec:scalingdetails}

\begin{figure*}[htbp]
\centering
\begin{tabular}{@{}p{0.45\columnwidth}p{0.54\columnwidth}@{}}
\toprule
<|im\_start|>user

What is the answer to Life, the Universe and Everything?

Think for up to 2048 tokens.

<|im\_start|>assistant
& 
<|im\_start|>user

What is the answer to Life, the Universe and Everything? 

Think for up to 64 steps.

<|im\_start|>assistant
\\
\defaultlightblue{
<|im\_start|>think

Let me break down this question into the three parts it is asking for: 1) Life 2) Universe 3) Everything

Let me start with life...
}
&
\defaultlightblue{
<|im\_start|>64 steps left

Let me break down this question into the three parts it is asking for: 1) Life 2) Universe 3) Everything

<|im\_start|>63 steps left

Let me start with life...
}
\\
\defaultblue{
<|im\_start|>answer

The answer is...
}
&
\defaultblue{
<|im\_start|>answer

The answer is...
}
\\
\bottomrule
\end{tabular}
\caption{\textbf{Token and step instruction data formats for controlling test-time compute.} We only train our model on the \defaultlightblue{reasoning trace} and the \defaultblue{answer}.}
\label{fig:format}
\end{figure*}

\begin{table}[htbp]
\centering
\caption{\textbf{Scaling thinking time via tokens-conditional control.} All metrics are averaged over the 30 questions in AIME24.}
\begin{tabular}{p{4cm} p{1cm} p{1cm} p{1cm} p{1cm} p{1cm}}
\toprule
Tokens instructed ($\rightarrow$) & 1024 & 2048 & 4096 & 8192 & 16384 \\
\midrule
\multicolumn{6}{c}{\textit{No intervention at test-time}} \\
\midrule
Thinking tokens & 7939 & 7158 & 8263 & 7108 & 7500 \\
Answer tokens & 689 & 669 & 659 & 722 & 724\\
AIME24 & 26.7 & 30.0 & 33.3 & 33.3 & 40.0 \\
\midrule
\multicolumn{6}{c}{\textit{Forcing end of thinking when token budget is reached}} \\
\midrule
Thinking tokens & 1024 & 2048 & 4031 & 5664 & 6330 \\
Answer tokens & 15 & 15 & 142 & 722 & 691 \\
AIME24 & 3.3 & 30.0 & 33.3 & 33.3 & 40.0 \\
\bottomrule
\end{tabular}
\label{tab:tokens}
\end{table}

\begin{table}[htbp]
\centering
\caption{\textbf{Scaling thinking time via step-conditional control.} All metrics are averaged over the 30 samples in AIME24. Token counts ignore the thinking and step delimiters.}
\begin{tabular}{p{4cm} p{1cm} p{1cm} p{1cm} p{1cm} p{1cm}}
\toprule
Steps instructed ($\rightarrow$) & 16 & 32 & 64 & 128 & 256 \\
\midrule
\multicolumn{6}{c}{\textit{No intervention at test-time}} \\
\midrule
Steps used & 123 & 90 & 80 & 82 & 136 \\
Tokens per step & 60 & 70 & 69 & 66 & 56 \\
Thinking tokens & 7252 & 6277 & 5396 & 5552 & 7551 \\
Answer tokens & 665 & 653 & 735 & 777 & 754 \\
AIME24 & 33.3 & 23.3 & 33.3 & 36.7 & 33.3 \\
\midrule
\multicolumn{6}{c}{\textit{Forcing end of thinking when 0 steps are reached}} \\
\midrule
Steps used & 16 & 32 & 59 & 78 & 136 \\
Tokens per step & 96 & 94 & 80 & 70 & 56 \\
Thinking tokens & 1517 & 2963 & 4636 & 5409 & 7551 \\
Answer tokens & 1111 & 788 & 799 & 794 & 754 \\
AIME24 & 23.3 & 23.3 & 33.3 & 36.7 & 33.3 \\
\bottomrule
\end{tabular}
\label{tab:steps}
\end{table}

\begin{table}[htbp]
\centering
\caption{\textbf{Scaling thinking time via class-conditional control.} We report ``accuracy / average thinking tokens per sample''; the higher the accuracy and the fewer the thinking tokens (inference cost) the better.}
\begin{tabular}{p{8cm} p{1.3cm} p{1.3cm} p{1.3cm}}
\toprule
Prompt appended to the question after two newlines & AIME24 & MATH500 & GPQA \\
\midrule
\textit{Answer after a short amount of thinking. Do not spend excessive time double-checking your work.} & 30.0\% / 8033 & 90.4\% / 2537 & 56.6\% / 4177 \\
& \\
\textit{Answer after a long amount of thinking. If you feel like you are finished early, spend the extra time trying to double-check your work until you are absolutely sure that you have the correct answer.} & 36.7\% / 9651 & 91.4\% / 3875 & 51.0\% / 4827 \\
\midrule
Without generic prompt appending & 50.0\% / 6109 & 93.0\% / 3298 & 57.6\% / 3510 \\
\bottomrule
\end{tabular}
\label{tab:prompt}
\end{table}

\paragraph{Token-conditional control} One general approach is to simply tell a model in the prompt precisely how many tokens it should generate. Ideally, the model can keep track of its token count and adjust its generation to finish within the desired limits. We experiment with this approach by training a model with token instructions using the format in \autoref{fig:format} (left). We bucket the lengths of the reasoning traces from our 1,000 training examples into powers of two (rounded upwards) and add a corresponding instruction to the user prompt. For example, if the instruction says ``Think for up to 2048 tokens'', then the reasoning trace has anywhere between 1024 and 2048 tokens. In \autoref{tab:tokens}, we show that after training the model hardly follows the token instruction. It does sometimes generate more tokens when given a higher limit but often overshoots the limit. This may not be unique to our model as prior work suggests that OpenAI o1-mini can also not follow token instructions~\citep{zhang_o1_inference_scaling_laws}. To prevent exceeding the limit, we test budget forcing the thinking to end once the limit is reached. This leads to perfect control (\autoref{tab:tokens} (lower)). With budget forcing, the scaling trend is also clearer as the model can no longer overshoot the limit when given a small thinking budget. This leads to better test-time scaling values for \textit{Token Prompting + budget forcing} in \autoref{tab:scalingabl}. To compute Control reported in \autoref{tab:scalingabl} for token-conditional control variants we divide the number of times the thinking tokens in \autoref{tab:tokens} are less than the upper limit by the total evaluations (2/5 for without intervention; 5/5 for with intervention).

\paragraph{Step-conditional control} Token instructions fail as current models cannot count tokens. To accommodate this lack of capability, we experiment with making the counting more coarse-grained. We partition the reasoning traces into steps and ask the model to think for a specific number of steps rather than tokens. We split our reasoning traces on double newlines into steps, which we find act as intuitive separators based on manual inspection of samples. We bucket our training samples into powers of 2 depending on their number of steps and add a corresponding step instruction following the format in \autoref{fig:format} (right). This format is based on early experiments, where we found the model to be more likely to adhere to the step limit when counting down (``3 steps left...2 steps left'') rather than counting up (``Step2...Step3...''). This is likely because if counting down, the final step is always 1, which will act as a strong prior to the model to finish its generation. If counting up, the final step before the answer varies, thus if the model does not remember the original step instruction, it may fail to stop. We conclude the following from our results in \autoref{tab:steps}: \textbf{(1)} The model still struggles to adhere to the step limit. The model sometimes simply continues counting into negative steps, e.g. ``-1 steps left''. To solve this issue, we automatically stop the thinking process once 0 steps are reached and then force the model to transition to answering mode by appending the answer token delimiter (\autoref{sec:ttc}). This leads to perfect step adherence (lower half of \autoref{tab:steps}), yet problems remain. \textbf{(2)} The model compensates for fewer steps by making each step longer. For example, when forced to use up to 16 steps vs 256 steps, the model generates an average of 96 tokens per step vs 56. Despite this issue, more steps still clearly correlate with more total thinking tokens in \autoref{tab:steps} and better performance leading to a positive slope \textbf{(3)} Step instructions are more costly than other methods. The step delimiters require around 6 tokens each which for e.g. 64 steps adds up to a total of around 380 tokens. When ignoring the step delimiters in token counts as in \autoref{tab:steps}, the model still requires 7551 thinking tokens on average to achieve only 33.3\% on AIME24. To compute Control reported in \autoref{tab:scalingabl} for step-conditional control variants, we first decide that 100 tokens are an upper limit per step and then multiply this number by the steps instructed to arrive at a proxy total token limit, e.g. 1600 for 16 steps instructed. We then check whether the thinking tokens in \autoref{tab:steps} fit within the respective limit for each evaluation run (3/5 for without intervention; 5/5 for with intervention). For the model in \autoref{fig:parallel}, we use a model with step-conditional control trained on an earlier version of our data and using an earlier version of our evaluation codebase.

\paragraph{Class-conditional control} OpenAI exposes test-time compute control to users via a ``reasoning\_effort'' API parameter with three possible settings: low, medium, and high.\footnote{\url{https://github.com/openai/openai-python/blob/44d6210f101abedeb2dd68507fcffcb329df70ea/src/openai/types/chat/completion_create_params.py\#L172}} The OpenAI documentation also states that ``Reducing reasoning effort \textit{can} result in faster responses and fewer tokens used on reasoning in a response." suggesting that they are unable to control test-time compute with guarantees. Thus, maybe OpenAI simply adjusts the prompt or system instruction depending on the reasoning effort desired. In \autoref{tab:prompt}, we show that separate prompts for short and long thinking allow us to control thinking time to some extent: Prompting the model to think for longer leads to longer thinking. However, it does not reliably improve performance and control is not precise. The current adherence to control may suffice when we only have three classes, but it might not scale to finer-grained classes. To compute Control reported in \autoref{tab:scalingabl} for this method, we assume that prompting the model to think for a short time in \autoref{tab:prompt} should produce fewer tokens than the default for AIME24, while the long prompt should produce more. As $8033>6109$ and $9651>6109$, one out of two follows our expected control thus Control is 50\%.

\subsection{Examples for rejection sampling ablation}
\label{sec:samplerej}

\newpage

\begin{longtable}{@{}p{\columnwidth}@{}}
\caption{\textbf{Answers generated by an earlier version of \model{} when rejection sampling on AIME24.} The black text is the question and the \textcolor{olmoeDarkYellow}{yellow} text is the solution. Based solely on the question we generate the \textcolor{defaultlightblue}{light blue} reasoning trace and \textcolor{defaultblue}{blue} response with \model{} using a temperature of 1. We generate until the reasoning trace has less than 4000 tokens for the first and less than 8000 tokens for the second example. It required on average 199 (1) tries per example until thinking was $\leq4000$ ($\leq8000$ tokens).} \\
\toprule

Let $O(0,0), A(\tfrac{1}{2}, 0),$ and $B(0, \tfrac{\sqrt{3}}{2})$ be points in the coordinate plane. Let $\mathcal{F}$ be the family of segments $\overline{PQ}$ of unit length lying in the first quadrant with $P$ on the $x$-axis and $Q$ on the $y$-axis. There is a unique point $C$ on $\overline{AB}$, distinct from $A$ and $B$, that does not belong to any segment from $\mathcal{F}$ other than $\overline{AB}$. Then $OC^2 = \tfrac{p}{q}$, where $p$ and $q$ are relatively prime positive integers. Find $p + q$.

\\

\defaultyellow{
23
}

\\
\textbf{[Correct generation with 3981 thinking tokens and 1055 answer tokens]}

\defaultlightblue{
The problem asks for a point $C$ on the segment $\overline{AB}$ that is not contained in any other segment from the family $\mathcal{F}$, except for $\overline{AB}$ itself.

The points $O, A, B$ form a triangle in the first quadrant.

$O = (0,0)$

$A = (\tfrac{1}{2}, 0)$

$B = (0, \tfrac{\sqrt{3}}{2})$

The segment $\overline{AB}$ connects $A(\tfrac{1}{2}, 0)$ and $B(0, \tfrac{\sqrt{3}}{2})$.

The equation of the line containing $\overline{AB}$ can be found using the two points:

Slope $m = \frac{\frac{\sqrt{3}}{2} - 0}{0 - \frac{1}{2}} = -\sqrt{3}$

Using the point-slope form $y - y_1 = m(x - x_1)$ with point $A(\tfrac{1}{2}, 0)$:

$y - 0 = -\sqrt{3}(x - \tfrac{1}{2})$

$y = -\sqrt{3}x + \frac{\sqrt{3}}{2}$

The segment $\overline{AB}$ lies in the first quadrant and connects $(\tfrac{1}{2}, 0)$ to $(0, \tfrac{\sqrt{3}}{2})$.

The family $\mathcal{F}$ consists of segments $\overline{PQ}$ of unit length, where $P$ is on the $x$-axis and $Q$ is on the $y$-axis. Let $P = (a, 0)$ and $Q = (0, b)$.

The length of $\overline{PQ}$ is $\sqrt{(a-0)^2 + (0-b)^2} = \sqrt{a^2 + b^2} = 1$.

So $a^2 + b^2 = 1$.

Since $P$ is on the positive $x$-axis and $Q$ is on the positive $y$-axis, we have $a > 0$ and $b > 0$.

The segment $\overline{PQ}$ connects $(a, 0)$ and $(0, b)$. The equation of the line containing $\overline{PQ}$ is $\frac{x}{a} + \frac{y}{b} = 1$.

The region covered by the family $\mathcal{F}$ is the area bounded by the lines containing these segments. The endpoints $(a, 0)$ and $(0, b)$ vary such that $a^2 + b^2 = 1$.

Consider the parametrization $a = \cos \theta$, $b = \sin \theta$, where $0 < \theta < \frac{\pi}{2}$.

The segment $\overline{PQ}$ connects $(\cos \theta, 0)$ and $(0, \sin \theta)$. The equation of the line is $\frac{x}{\cos \theta} + \frac{y}{\sin \theta} = 1$.

We are looking for a unique point $C$ on $\overline{AB}$, distinct from $A$ and $B$, that does not belong to any segment from $\mathcal{F}$ other than $\overline{AB}$.

Let $C$ be a point on $\overline{AB}$. We can parametrize the points on $\overline{AB}$ as $(1-t)A + tB$ for $0 \le t \le 1$.

$C = (1-t)(\tfrac{1}{2}, 0) + t(0, \tfrac{\sqrt{3}}{2}) = (\tfrac{1}{2} - \tfrac{1}{2}t, \tfrac{\sqrt{3}}{2}t)$.

Since $C$ is distinct from $A$ and $B$, we have $0 < t < 1$.

Suppose the point $C$ lies on some segment $\overline{PQ} \in \mathcal{F}$, where $P = (\cos \theta, 0)$ and $Q = (0, \sin \theta)$.

The point $C$ satisfies the equation $\frac{x}{\cos \theta} + \frac{y}{\sin \theta} = 1$.
}
\\
\defaultlightblue{
Plugging in the coordinates of $C$:

$\frac{\frac{1}{2} - \frac{1}{2}t}{\cos \theta} + \frac{\frac{\sqrt{3}}{2}t}{\sin \theta} = 1$

$\frac{1-t}{2 \cos \theta} + \frac{\sqrt{3}t}{2 \sin \theta} = 1$

Multiply by 2:

$\frac{1-t}{\cos \theta} + \frac{\sqrt{3}t}{\sin \theta} = 2$

$(1-t) \sin \theta + \sqrt{3}t \cos \theta = 2 \sin \theta \cos \theta$

$\sin \theta - t \sin \theta + \sqrt{3}t \cos \theta = \sin(2\theta) / 2$

We are looking for a point $C$ on $\overline{AB}$ such that the only segment from $\mathcal{F}$ containing $C$ is $\overline{AB}$ itself.

The segment $\overline{AB}$ is part of the family $\mathcal{F}$.

For $\overline{AB}$, $P = A = (\tfrac{1}{2}, 0)$ and $Q = B = (0, \tfrac{\sqrt{3}}{2})$.

Here $a = \tfrac{1}{2}$, $b = \tfrac{\sqrt{3}}{2}$.

$a^2 + b^2 = (\tfrac{1}{2})^2 + (\tfrac{\sqrt{3}}{2})^2 = \tfrac{1}{4} + \tfrac{3}{4} = 1$.

The line segment $\overline{AB}$ connects $(\tfrac{1}{2}, 0)$ and $(0, \tfrac{\sqrt{3}}{2})$.

The equation of the line is $\frac{x}{1/2} + \frac{y}{\sqrt{3}/2} = 1$, which simplifies to $2x + \frac{2}{\sqrt{3}}y = 1$, or $\sqrt{3}x + y = \frac{\sqrt{3}}{2}$.

This matches the equation $y = -\sqrt{3}x + \frac{\sqrt{3}}{2}$.

Consider the envelope of the family of lines $\frac{x}{\cos \theta} + \frac{y}{\sin \theta} = 1$.

Differentiating with respect to $\theta$:

$-\frac{x \sin \theta}{\cos^2 \theta} + \frac{y \cos \theta}{\sin^2 \theta} = 0$

$\frac{y \cos \theta}{\sin^2 \theta} = \frac{x \sin \theta}{\cos^2 \theta}$

$y \cos^3 \theta = x \sin^3 \theta$

$\frac{y}{x} = \tan^3 \theta$

$\tan \theta = (\frac{y}{x})^{1/3}$

From the line equation: $y \sin \theta = \sin \theta \cos \theta - x \sin \theta$

$y = \cos \theta - x$

Substitute $\cos \theta = \frac{1}{\sqrt{1 + \tan^2 \theta}} = \frac{1}{\sqrt{1 + (y/x)^{2/3}}}$

$\cos \theta = \frac{x^{1/3}}{(x^{2/3} + y^{2/3})^{1/2}}$

$\sin \theta = \tan \theta \cos \theta = (\frac{y}{x})^{1/3} \frac{x^{1/3}}{(x^{2/3} + y^{2/3})^{1/2}} = \frac{y^{1/3}}{(x^{2/3} + y^{2/3})^{1/2}}$

Substitute back into the line equation:

$\frac{x}{\cos \theta} + \frac{y}{\sin \theta} = 1$

$x (x^{2/3} + y^{2/3})^{1/2} / x^{1/3} + y (x^{2/3} + y^{2/3})^{1/2} / y^{1/3} = 1$

$x^{2/3} (x^{2/3} + y^{2/3})^{1/2} + y^{2/3} (x^{2/3} + y^{2/3})^{1/2} = 1$

$(x^{2/3} + y^{2/3})^{3/2} = 1$

$x^{2/3} + y^{2/3} = 1$

The envelope of the family $\mathcal{F}$ is the astroid $x^{2/3} + y^{2/3} = 1$.

The segments from $\mathcal{F}$ are tangent to this astroid.

The point $C$ on $\overline{AB}$ that we are looking for is the point where $\overline{AB}$ is tangent to the astroid.

The equation of $\overline{AB}$ is $y = -\sqrt{3}x + \frac{\sqrt{3}}{2}$.

Substitute into the astroid equation:

$x^{2/3} + (-\sqrt{3}x + \frac{\sqrt{3}}{2})^{2/3} = 1$

Let $u = x^{1/3}$ and $v = y^{1/3}$. The astroid is $u^2 + v^2 = 1$.

The point $C$ lies on both the line $\overline{AB}$ and the astroid.

Differentiating the astroid equation implicitly:

$\frac{2}{3} x^{-1/3} + \frac{2}{3} y^{-1/3} y' = 0$

$y' = -\frac{y^{1/3}}{x^{1/3}} = -(\frac{y}{x})^{1/3}$
}
\\
\defaultlightblue{
The slope of the tangent to the astroid at a point $(x, y)$ is $-\sqrt[3]{\frac{y}{x}}$.

The slope of the line $\overline{AB}$ is $-\sqrt{3}$.

So, at the point of tangency, $-\sqrt[3]{\frac{y}{x}} = -\sqrt{3}$.

$\sqrt[3]{\frac{y}{x}} = \sqrt{3}$

$\frac{y}{x} = (\sqrt{3})^3 = 3\sqrt{3}$

The point $C$ lies on the line $y = -\sqrt{3}x + \frac{\sqrt{3}}{2}$ and satisfies $\frac{y}{x} = 3\sqrt{3}$.

So $y = 3\sqrt{3} x$.

Substitute into the line equation:

$3\sqrt{3} x = -\sqrt{3} x + \frac{\sqrt{3}}{2}$

$3x = -x + \frac{1}{2}$

$4x = \frac{1}{2}$

$x = \frac{1}{8}$

Then $y = 3\sqrt{3} (\frac{1}{8}) = \frac{3\sqrt{3}}{8}$.

The point $C$ is $(\frac{1}{8}, \frac{3\sqrt{3}}{8})$.

We need to check if this point lies on the segment $\overline{AB}$.

The point $C$ is of the form $(\tfrac{1}{2} - \tfrac{1}{2}t, \tfrac{\sqrt{3}}{2}t)$.

$\tfrac{1}{2} - \tfrac{1}{2}t = \tfrac{1}{8} \implies 4 - 4t = 1 \implies 3 = 4t \implies t = \tfrac{3}{4}$.

$\tfrac{\sqrt{3}}{2}t = \tfrac{\sqrt{3}}{2} \cdot \tfrac{3}{4} = \tfrac{3\sqrt{3}}{8}$.

So $C$ lies on $\overline{AB}$.

We need to make sure $0 < t < 1$, which is true for $t = \frac{3}{4}$.

Now, we need to check if this point $C$ does not belong to any other segment from $\mathcal{F}$.

The point $C$ lies on the astroid. The segment $\overline{AB}$ is tangent to the astroid at $C$.

Any other segment from $\mathcal{F}$ that contains $C$ would have $C$ as an interior point, not a tangent point.

Consider the parametrization of the family $\mathcal{F}$.

A point $(x, y)$ belongs to some segment in $\mathcal{F}$ iff there exists $\theta$ such that $\frac{x}{\cos \theta} + \frac{y}{\sin \theta} = 1$.

For the point $C = (\frac{1}{8}, \frac{3\sqrt{3}}{8})$, we have:

$\frac{1/8}{\cos \theta} + \frac{3\sqrt{3}/8}{\sin \theta} = 1$

$\frac{1}{\cos \theta} + \frac{3\sqrt{3}}{\sin \theta} = 8$

$\sin \theta + 3\sqrt{3} \cos \theta = 8 \sin \theta \cos \theta$

$\sin \theta + 3\sqrt{3} \cos \theta = 4 \sin(2\theta)$

We know that $C$ lies on $\overline{AB}$, which corresponds to some specific segment in $\mathcal{F}$.

The problem statement assumes the existence and uniqueness of such a point $C$.

The unique point $C$ on $\overline{AB}$, distinct from $A$ and $B$, that does not belong to any segment from $\mathcal{F}$ other than $\overline{AB}$.

This means that if $C$ lies on another segment $\overline{P'Q'} \in \mathcal{F}$, then $\overline{P'Q'} = \overline{AB}$.

The point $C$ is the point of tangency of $\overline{AB}$ with the astroid.

Now we need to find $OC^2$.

$OC^2 = (\frac{1}{8})^2 + (\frac{3\sqrt{3}}{8})^2 = \frac{1}{64} + \frac{27}{64} = \frac{28}{64} = \frac{7}{16}$.

Here $p = 7$ and $q = 16$.

$p$ and $q$ are relatively prime.

$p + q = 7 + 16 = 23$.

Final check of the reasoning:

The family $\mathcal{F}$ consists of unit-length segments with endpoints on the axes.

The envelope of this family is the astroid $x^{2/3} + y^{2/3} = 1$.

The segments from $\mathcal{F}$ are tangent to the astroid.

The segment $\overline{AB}$ is also a member of the family $\mathcal{F}$.

We are looking for a point $C$ on $\overline{AB}$ that lies on the astroid.

This point $C$ is the point of tangency of $\overline{AB}$ with the astroid.
}
\\
\defaultlightblue{
If a point on $\overline{AB}$ lies on the astroid, it belongs to the tangent segment at that point. Since $\overline{AB}$ is the tangent segment, this point does not belong to any other segment from $\mathcal{F}$.

Consider the case when the segment $\overline{AB}$ intersects the astroid at two points.

If the segment $\overline{AB}$ is not tangent to the astroid, then there are two points of intersection.

Let the points of intersection be $C_1$ and $C_2$.

If $C_1$ lies on $\overline{AB}$, then the tangent to the astroid at $C_1$ is the segment $\overline{AB}$.

If $C_2$ lies on $\overline{AB}$, then the tangent to the astroid at $C_2$ is the segment $\overline{AB}$.

The slope of $\overline{AB}$ is $-\sqrt{3}$.

We found the point on the astroid with tangent slope $-\sqrt{3}$.

The problem statement guarantees the uniqueness of such a point $C$.

The point $C$ must be interior to $\overline{AB}$, so $0 < t < 1$.

The parameter $t = \frac{3}{4}$ is in this range.

The point $C$ is distinct from $A$ and $B$.

If $C = A$, then $t = 0$, $C = (\tfrac{1}{2}, 0)$.

If $C = B$, then $t = 1$, $C = (0, \tfrac{\sqrt{3}}{2})$.

The point of tangency $C$ is the unique point on $\overline{AB}$ (excluding endpoints) that lies on the astroid.
}
\\
\defaultblue{
Solution:

Let the family of segments $\mathcal{F}$ be defined by segments $\overline{PQ}$ of unit length with $P$ on the positive $x$-axis and $Q$ on the positive $y$-axis. Let $P = (a, 0)$ and $Q = (0, b)$. The length condition gives $a^2 + b^2 = 1$. The segment $\overline{PQ}$ lies on the line $\frac{x}{a} + \frac{y}{b} = 1$. Using the parametrization $a = \cos \theta$, $b = \sin \theta$ with $0 < \theta < \frac{\pi}{2}$, the equation of the line is $\frac{x}{\cos \theta} + \frac{y}{\sin \theta} = 1$.

The envelope of this family of lines is the astroid $x^{2/3} + y^{2/3} = 1$.

The segments from $\mathcal{F}$ are tangent to this astroid.

The segment $\overline{AB}$ connects $A(\tfrac{1}{2}, 0)$ and $B(0, \tfrac{\sqrt{3}}{2})$. The equation of the line containing $\overline{AB}$ is $y = -\sqrt{3}x + \frac{\sqrt{3}}{2}$. This segment is part of the family $\mathcal{F}$.

We are looking for a unique point $C$ on $\overline{AB}$, distinct from $A$ and $B$, that does not belong to any segment from $\mathcal{F}$ other than $\overline{AB}$. This means $C$ is the point where $\overline{AB}$ is tangent to the astroid.

The slope of the tangent to the astroid at a point $(x, y)$ is $-\sqrt[3]{\frac{y}{x}}$. The slope of $\overline{AB}$ is $-\sqrt{3}$.

Setting the slopes equal: $-\sqrt[3]{\frac{y}{x}} = -\sqrt{3} \implies \frac{y}{x} = 3\sqrt{3}$.

The point $C$ lies on the line $y = -\sqrt{3}x + \frac{\sqrt{3}}{2}$ and satisfies $y = 3\sqrt{3}x$.

Substituting, $3\sqrt{3}x = -\sqrt{3}x + \frac{\sqrt{3}}{2} \implies 4\sqrt{3}x = \frac{\sqrt{3}}{2} \implies x = \frac{1}{8}$.

Then $y = 3\sqrt{3}(\frac{1}{8}) = \frac{3\sqrt{3}}{8}$.

The point $C$ is $(\frac{1}{8}, \frac{3\sqrt{3}}{8})$.

We need to check if $C$ lies on $\overline{AB}$. The points on $\overline{AB}$ are $(\tfrac{1}{2} - \tfrac{1}{2}t, \tfrac{\sqrt{3}}{2}t)$ for $0 \le t \le 1$.

For $x = \frac{1}{8}$, $\tfrac{1}{2} - \tfrac{1}{2}t = \frac{1}{8} \implies t = \frac{3}{4}$.

For $y = \frac{3\sqrt{3}}{8}$, $\tfrac{\sqrt{3}}{2}t = \frac{3\sqrt{3}}{8} \implies t = \frac{3}{4}$.

Since $0 < \frac{3}{4} < 1$, $C$ is an interior point of $\overline{AB}$.

The point $C$ is the unique point on $\overline{AB}$ (excluding endpoints) that lies on the astroid. The tangent to the astroid at $C$ is the segment $\overline{AB}$. Thus, $C$ does not belong to any other segment from $\mathcal{F}$.

Finally, we calculate $OC^2 = (\frac{1}{8})^2 + (\frac{3\sqrt{3}}{8})^2 = \frac{1}{64} + \frac{27}{64} = \frac{28}{64} = \frac{7}{16}$.

We have $p = 7$ and $q = 16$. These are relatively prime.

$p + q = 7 + 16 = 23$.

Final Answer: The final answer is $\boxed{23}$
}

\\
\textbf{[Incorrect generation with 5923 thinking tokens and 754 answer tokens]}

\defaultlightblue{
The problem asks for a point $C$ on the segment $\overline{AB}$ such that $C$ is not on any other segment from the family $\mathcal{F}$, except $\overline{AB}$ itself. The family $\mathcal{F}$ consists of segments $\overline{PQ}$ of unit length, with $P$ on the $x$-axis and $Q$ on the $y$-axis in the first quadrant.

Let $P = (x, 0)$ and $Q = (0, y)$, where $x > 0$ and $y > 0$. The length of the segment $\overline{PQ}$ is 1, so the distance between $P$ and $Q$ is 1.

$\sqrt{(x - 0)^2 + (0 - y)^2} = 1$

$\sqrt{x^2 + y^2} = 1$

$x^2 + y^2 = 1$

The segment $\overline{PQ}$ lies in the first quadrant, so $x > 0$ and $y > 0$.

The equation $x^2 + y^2 = 1$ represents the quarter circle in the first quadrant, with endpoints at $(1, 0)$ and $(0, 1)$.

The points on the segment $\overline{PQ}$ can be parameterized as $(1 - t)x + t(0), (1 - t)0 + ty) = (x(1 - t), yt)$, where $0 \le t \le 1$.

The segment $\overline{AB}$ connects $A(\tfrac{1}{2}, 0)$ and $B(0, \tfrac{\sqrt{3}}{2})$.

The equation of the line containing $\overline{AB}$ can be found using the two points:

Slope $m = \frac{\frac{\sqrt{3}}{2} - 0}{0 - \frac{1}{2}} = \frac{\frac{\sqrt{3}}{2}}{-\frac{1}{2}} = -\sqrt{3}$.

Using point-slope form with point $A(\tfrac{1}{2}, 0)$:

$y - 0 = -\sqrt{3}(x - \frac{1}{2})$

$y = -\sqrt{3}x + \frac{\sqrt{3}}{2}$

The segment $\overline{AB}$ lies in the first quadrant, between $x = 0$ and $x = \frac{1}{2}$.

Alternatively, we can parameterize the points on $\overline{AB}$.

Let $C$ be a point on $\overline{AB}$. Then $C = (1 - \lambda)A + \lambda B$ for some $0 \le \lambda \le 1$.

$C = (1 - \lambda)(\frac{1}{2}, 0) + \lambda(0, \frac{\sqrt{3}}{2}) = (\frac{1}{2} - \frac{1}{2}\lambda, \frac{\sqrt{3}}{2}\lambda)$.

We are looking for a unique point $C$ on $\overline{AB}$, distinct from $A$ and $B$, that does not belong to any segment from $\mathcal{F}$ other than $\overline{AB}$.

Consider a segment from $\mathcal{F}$, with endpoints $P = (x, 0)$ and $Q = (0, y)$, where $x^2 + y^2 = 1$.

A point on this segment is of the form $(x(1 - t), yt)$, where $0 \le t \le 1$.

We want to find a point $C$ on $\overline{AB}$, $C \ne A, B$, such that if $C$ lies on a segment $\overline{PQ} \in \mathcal{F}$, then $\overline{PQ} = \overline{AB}$.

Let $C = (\frac{1}{2} - \frac{1}{2}\lambda, \frac{\sqrt{3}}{2}\lambda)$, where $0 < \lambda < 1$.

If $C$ lies on a segment $\overline{PQ}$ from $\mathcal{F}$, then there exist $x, y > 0$ with $x^2 + y^2 = 1$, and $t \in [0, 1]$ such that:

$\frac{1}{2} - \frac{1}{2}\lambda = x(1 - t)$

$\frac{\sqrt{3}}{2}\lambda = yt$

From the first equation, $1 - t = \frac{\frac{1}{2} - \frac{1}{2}\lambda}{x} = \frac{1 - \lambda}{2x}$.

From the second equation, $t = \frac{\frac{\sqrt{3}}{2}\lambda}{y} = \frac{\sqrt{3}\lambda}{2y}$.

Since $t \in [0, 1]$, we have $1 - t \in [0, 1]$.

Also, $x > 0$, so $\frac{1 - \lambda}{2x} > 0$, which implies $1 - \lambda > 0$, or $\lambda < 1$. This is consistent with the condition for point on $\overline{AB}$.

We have $1 - t = \frac{1 - \lambda}{2x}$ and $t = \frac{\sqrt{3}\lambda}{2y}$.

Adding these equations: $1 = \frac{1 - \lambda}{2x} + \frac{\sqrt{3}\lambda}{2y}$

$2xy = (1 - \lambda)y + \sqrt{3}\lambda x$
}
\\
\defaultlightblue{
We know $x^2 + y^2 = 1$.

If the point $C$ belongs to the segment $\overline{AB}$ itself, then the endpoints of the segment are $A$ and $B$.

So, the segment is from $(\tfrac{1}{2}, 0)$ to $(0, \tfrac{\sqrt{3}}{2})$.

Here, $x = \tfrac{1}{2}$ and $y = \tfrac{\sqrt{3}}{2}$.

Check if $x^2 + y^2 = 1$: $(\tfrac{1}{2})^2 + (\tfrac{\sqrt{3}}{2})^2 = \tfrac{1}{4} + \tfrac{3}{4} = 1$. So $\overline{AB} \in \mathcal{F}$.

If $C$ belongs to another segment $\overline{PQ}$ from $\mathcal{F}$, with endpoints $P = (x, 0)$ and $Q = (0, y)$, where $(x, y) \ne (\tfrac{1}{2}, \tfrac{\sqrt{3}}{2})$.

Consider the geometric interpretation. The family $\mathcal{F}$ consists of segments of unit length with endpoints on the axes. The midpoints of these segments lie on the circle $(x - \frac{1}{2})^2 + (y - \frac{1}{2})^2 = \frac{1}{2}$.

Let the point $C$ correspond to some $\lambda$.

If $C$ lies on $\overline{PQ}$, then $C = (x(1 - t), yt)$.

So, $\frac{1}{2} - \frac{1}{2}\lambda = x(1 - t)$ and $\frac{\sqrt{3}}{2}\lambda = yt$.

We need to find a $\lambda \in (0, 1)$ such that the only solution to these equations, with $x^2 + y^2 = 1$, $x, y > 0$, is $x = \tfrac{1}{2}$ and $y = \tfrac{\sqrt{3}}{2}$.

From the equations, $1 - t = \frac{1 - \lambda}{2x}$ and $t = \frac{\sqrt{3}\lambda}{2y}$.

Substitute these into $x^2 + y^2 = 1$.

Try a different approach. Consider the tangents to the family of segments.

The segments are formed by points $(x, 0)$ and $(0, y)$ with $x^2 + y^2 = 1$.

The equation of the line containing the segment $\overline{PQ}$ is $\frac{x}{x} + \frac{y}{y} = 1$, or $\frac{x}{x} + \frac{y}{y} = 1$.

The equation of the line is $\frac{x}{x} + \frac{y}{y} = 1$.

Consider the equation of the segment in parametric form: $(x(1 - t), yt)$.

If a point $C$ lies on this segment, then $C = (x_c, y_c)$, where $x_c = x(1 - t)$ and $y_c = yt$.

So, $\frac{x_c}{x} = 1 - t$ and $\frac{y_c}{y} = t$.

$\frac{x_c}{x} + \frac{y_c}{y} = 1 - t + t = 1$.

The segments from $\mathcal{F}$ are parts of the lines $\frac{x}{x_0} + \frac{y}{y_0} = 1$, where $x_0^2 + y_0^2 = 1$, $x_0, y_0 > 0$.

Consider the point $C = (\frac{1}{2} - \frac{1}{2}\lambda, \frac{\sqrt{3}}{2}\lambda)$.

If $C$ lies on a segment from $\mathcal{F}$ with endpoints $(x, 0)$ and $(0, y)$, then:

$\frac{\frac{1}{2} - \frac{1}{2}\lambda}{x} + \frac{\frac{\sqrt{3}}{2}\lambda}{y} = 1$

$\frac{1 - \lambda}{2x} + \frac{\sqrt{3}\lambda}{2y} = 1$

We need to find a unique $\lambda \in (0, 1)$ such that the only solution $(x, y)$ to this equation, with $x^2 + y^2 = 1$, $x, y > 0$, is $(\frac{1}{2}, \frac{\sqrt{3}}{2})$.

Substitute $x = \frac{1}{2}$ and $y = \frac{\sqrt{3}}{2}$ into the equation:

$\frac{1 - \lambda}{2(\frac{1}{2})} + \frac{\sqrt{3}\lambda}{2(\frac{\sqrt{3}}{2})} = 1$

$1 - \lambda + \lambda = 1$

$1 = 1$

This shows that any point on $\overline{AB}$ lies on the segment $\overline{AB}$ itself.

Consider the line containing $\overline{AB}$: $y = -\sqrt{3}x + \frac{\sqrt{3}}{2}$.

If a point $C$ on $\overline{AB}$ also lies on another segment from $\mathcal{F}$ with endpoints $(x, 0)$ and $(0, y)$, then:

$\frac{x_C}{x} + \frac{y_C}{y} = 1$.

Let the point $C$ divide $\overline{AB}$ in the ratio $\mu : (1 - \mu)$, where $0 < \mu < 1$.

$C = (1 - \mu)A + \mu B = (1 - \mu)(\frac{1}{2}, 0) + \mu(0, \frac{\sqrt{3}}{2}) = (\frac{1}{2} - \frac{1}{2}\mu, \frac{\sqrt{3}}{2}\mu)$.

If $C$ lies on a segment with endpoints $(x, 0)$ and $(0, y)$, then:

$\frac{\frac{1}{2} - \frac{1}{2}\mu}{x} + \frac{\frac{\sqrt{3}}{2}\mu}{y} = 1$.

We are looking for a point $C$ such that the line passing through $C$ with the form $\frac{x}{x_0} + \frac{y}{y_0} = 1$, where $x_0^2 + y_0^2 = 1$, is unique.
}
\\
\defaultlightblue{
Consider the tangents to the unit circle centered at the origin. The lines of the form $\frac{x}{x_0} + \frac{y}{y_0} = 1$ are tangents to the circle $x^2 + y^2 = r^2$.

Consider the dual problem. The family of segments corresponds to points on the quarter circle $x^2 + y^2 = 1$ in the first quadrant.

The lines containing the segments are $\frac{x}{x_0} + \frac{y}{y_0} = 1$.

The envelope of these lines is the curve whose tangents are these lines.

The equation $\frac{x}{x_0} + \frac{y}{y_0} = 1$, with the constraint $x_0^2 + y_0^2 = 1$.

Let $x_0 = \cos \theta$, $y_0 = \sin \theta$, where $0 < \theta < \frac{\pi}{2}$.

The equations of the lines are $x \cos \theta + y \sin \theta = 1$.

The envelope of the family of lines $x \cos \theta + y \sin \theta = 1$ is the circle $x^2 + y^2 = 1$.

The segments from $\mathcal{F}$ are parts of the tangents to the unit circle in the first quadrant.

The point $C$ on $\overline{AB}$ that does not belong to any other segment from $\mathcal{F}$ means that the tangent to the unit circle passing through $C$ is unique.

Consider the point $C = (\frac{1}{2} - \frac{1}{2}\lambda, \frac{\sqrt{3}}{2}\lambda)$.

The tangents to the unit circle passing through $C$ have the form $x x_C + y y_C = 1$.

If the tangent is unique, then $C$ lies on the unit circle.

$(\frac{1}{2} - \frac{1}{2}\lambda)^2 + (\frac{\sqrt{3}}{2}\lambda)^2 = 1$

$(\frac{1}{4} - \frac{1}{2}\lambda + \frac{1}{4}\lambda^2) + \frac{3}{4}\lambda^2 = 1$

$\frac{1}{4} - \frac{1}{2}\lambda + \lambda^2 = 1$

$4\lambda^2 - 2\lambda + 1 = 4$

$4\lambda^2 - 2\lambda - 3 = 0$

Solve for $\lambda$:

$\lambda = \frac{-(-2) \pm \sqrt{(-2)^2 - 4(4)(-3)}}{2(4)} = \frac{2 \pm \sqrt{4 + 48}}{8} = \frac{2 \pm \sqrt{52}}{8} = \frac{2 \pm 2\sqrt{13}}{8} = \frac{1 \pm \sqrt{13}}{4}$.

We need $0 < \lambda < 1$.

$\sqrt{13} \approx 3.6$.

$\frac{1 + 3.6}{4} = \frac{4.6}{4} > 1$.

$\frac{1 - 3.6}{4} = \frac{-2.6}{4} < 0$.

Something is wrong. The point $C$ is inside the unit circle.

Let's rephrase the condition. There is a unique point $C$ on $\overline{AB}$ such that if $C$ lies on a segment $\overline{PQ} \in \mathcal{F}$, then $\overline{PQ} = \overline{AB}$.

Let the point $C$ on $\overline{AB}$ be $(\frac{1}{2} - \frac{1}{2}\lambda, \frac{\sqrt{3}}{2}\lambda)$.

Consider a segment from $\mathcal{F}$ with endpoints $(x, 0)$ and $(0, y)$, $x^2 + y^2 = 1$.

The parametric form of the segment is $(x(1 - t), yt)$.

If $C$ is on this segment, then for some $t \in [0, 1]$:

$\frac{1}{2} - \frac{1}{2}\lambda = x(1 - t)$

$\frac{\sqrt{3}}{2}\lambda = yt$

If the segment is unique, then the values of $x$ and $y$ are unique.

The only solution to $\frac{1 - \lambda}{2x} + \frac{\sqrt{3}\lambda}{2y} = 1$, with $x^2 + y^2 = 1$, $x, y > 0$, should be $x = \frac{1}{2}, y = \frac{\sqrt{3}}{2}$.

Consider the intersection of the line containing $\overline{AB}$ with the boundary of the region formed by the segments in $\mathcal{F}$.

The boundaries are the $x$-axis, $y$-axis, and the quarter circle $x^2 + y^2 = 1$.

Consider the dual problem again. The family of lines $\frac{x}{x_0} + \frac{y}{y_0} = 1$, where $x_0^2 + y_0^2 = 1$, $x_0, y_0 > 0$.

The envelope of these lines is the circle $x^2 + y^2 = 1$.

A point belongs to a unique segment from $\mathcal{F}$ if it lies on the circle $x^2 + y^2 = 1$.

The point $C$ on $\overline{AB}$ satisfies $C = t A + (1 - t) B$.

Let's consider the case where the segment is $\overline{AB}$. Here $x = \frac{1}{2}, y = \frac{\sqrt{3}}{2}$.

The point $C$ can be represented as $(\frac{1}{2}(1 - s), \frac{\sqrt{3}}{2}s)$ for $0 \le s \le 1$.

We are looking for a point $C$ on $\overline{AB}$ that does not lie on any other segment from $\mathcal{F}$.
}
\\
\defaultlightblue{
Consider the geometry. The family $\mathcal{F}$ forms a region bounded by the axes and the quarter circle.

The segments are chords of circles centered on the line $y = x$, tangent to the axes.

Let the endpoints of a segment be $(\cos \theta, 0)$ and $(0, \sin \theta)$.

A point on the segment is $(\cos \theta (1 - t), \sin \theta t)$.

The point $C = (\frac{1}{2} - \frac{1}{2}\lambda, \frac{\sqrt{3}}{2}\lambda)$.

If $C$ lies on such a segment, then $\frac{\frac{1}{2} - \frac{1}{2}\lambda}{\cos \theta} = 1 - t$ and $\frac{\frac{\sqrt{3}}{2}\lambda}{\sin \theta} = t$.

$\frac{1 - \lambda}{2\cos \theta} + \frac{\sqrt{3}\lambda}{2\sin \theta} = 1$.

We need a unique $\lambda \in (0, 1)$ such that the only solution $(\cos \theta, \sin \theta)$ is $(\frac{1}{2}, \frac{\sqrt{3}}{2})$, which corresponds to $\theta = \frac{\pi}{3}$.

Consider the function $f(\theta) = \frac{1}{2\cos \theta} - \frac{\lambda}{2\cos \theta} + \frac{\sqrt{3}\lambda}{2\sin \theta}$.

We need $f(\theta) = 1$.

If $\lambda = \frac{1}{2}$, then $C = (\frac{1}{4}, \frac{\sqrt{3}}{4})$.

$\frac{1 - 1/2}{2\cos \theta} + \frac{\sqrt{3}(1/2)}{2\sin \theta} = 1$

$\frac{1}{4\cos \theta} + \frac{\sqrt{3}}{4\sin \theta} = 1$

$\frac{1}{\cos \theta} + \frac{\sqrt{3}}{\sin \theta} = 4$

$\sin \theta + \sqrt{3} \cos \theta = 4 \sin \theta \cos \theta$

$\sin \theta + \sqrt{3} \cos \theta = 2 \sin(2\theta)$

Divide by 2: $\frac{1}{2} \sin \theta + \frac{\sqrt{3}}{2} \cos \theta = \sin(2\theta)$

$\sin(\theta + \frac{\pi}{3}) = \sin(2\theta)$.

$\theta + \frac{\pi}{3} = 2\theta + 2\pi k$, or $\theta + \frac{\pi}{3} = \pi - 2\theta + 2\pi k$.

$\theta = \frac{\pi}{3} - 2\pi k$, so $\theta = \frac{\pi}{3}$.

$3\theta = \pi - \frac{\pi}{3} + 2\pi k = \frac{2\pi}{3} + 2\pi k$.

$\theta = \frac{2\pi}{9} + \frac{2\pi k}{3}$.

If $k = 0$, $\theta = \frac{2\pi}{9}$.

If $k = 1$, $\theta = \frac{2\pi}{9} + \frac{2\pi}{3} = \frac{2\pi + 6\pi}{9} = \frac{8\pi}{9}$.

If $\lambda = \frac{1}{2}$, the point is $(\frac{1}{4}, \frac{\sqrt{3}}{4})$.

This point lies on segments corresponding to $\theta = \frac{\pi}{3}$ and $\theta = \frac{2\pi}{9}$.

The unique point $C$ means that for other values of $\lambda$, there is more than one solution for $(x, y)$.

Consider the tangent to the unit circle passing through $C$.

If the tangent is unique, $C$ is on the unit circle.

This leads to $\lambda = \frac{1 \pm \sqrt{13}}{4}$, neither of which is in $(0, 1)$.

Consider the geometric interpretation of the family $\mathcal{F}$. The segments connect points on the axes such that the distance is 1.

The locus of the centers of these segments is the circle $(x - \frac{1}{2})^2 + (y - \frac{1}{2})^2 = \frac{1}{2}$, in the first quadrant.

If a point $C$ lies on a segment from $\mathcal{F}$, then $C$ is on a line segment connecting $(x, 0)$ and $(0, y)$ with $x^2 + y^2 = 1$.

The line containing the segment is $\frac{x}{x_0} + \frac{y}{y_0} = 1$.

If $C$ belongs to a unique segment, then there is a unique line of this form passing through $C$.

The point $C$ on $\overline{AB}$ is $(\frac{1}{2} - \frac{1}{2}\lambda, \frac{\sqrt{3}}{2}\lambda)$.

We need a unique $(x_0, y_0)$ with $x_0^2 + y_0^2 = 1$, $x_0, y_0 > 0$, such that $\frac{\frac{1}{2} - \frac{1}{2}\lambda}{x_0} + \frac{\frac{\sqrt{3}}{2}\lambda}{y_0} = 1$.

Let the point $C$ be such that the line passing through $C$ with the form $\frac{x}{x_0} + \frac{y}{y_0} = 1$ is unique, subject to $x_0^2 + y_0^2 = 1$.

This line is tangent to the unit circle.

The point $C$ lies on the intersection of the line containing $\overline{AB}$ and the unit circle.

$(\frac{1}{2} - \frac{1}{2}\lambda)^2 + (\frac{\sqrt{3}}{2}\lambda)^2 = 1$.

We found no solution for $\lambda \in (0, 1)$.
}
\\
\defaultlightblue{
The unique point $C$ means that if any other segment passes through $C$, it must be $\overline{AB}$.

Consider the slope of the line containing the segment connecting $(x, 0)$ and $(0, y)$. Slope is $-\frac{y}{x}$.

The equation is $y = -\frac{y}{x} x + y$.

If $C$ lies on the segment, then $\frac{1 - \lambda}{2x} + \frac{\sqrt{3}\lambda}{2y} = 1$.

Consider the geometric mean.

Let the point $C$ have coordinates $(a, b)$. The segment connecting $(x, 0)$ and $(0, y)$ passes through $(a, b)$.

$\frac{a}{x} + \frac{b}{y} = 1$.

If $C$ is on $\overline{AB}$, then $b = -\sqrt{3}a + \frac{\sqrt{3}}{2}$.

$a = \frac{1}{2} - \frac{1}{2}\lambda$, $b = \frac{\sqrt{3}}{2}\lambda$.

The point $C$ is tangent to the envelope of the family of segments.
}

\\

\defaultblue{
Solution:

Let the family of segments be $\mathcal{F}$. The segments have endpoints $P = (x, 0)$ and $Q = (0, y)$ with $x^2 + y^2 = 1$ and $x, y > 0$. The points on the segment $\overline{PQ}$ satisfy $\frac{x}{x} + \frac{y}{y} = 1$. The segments lie on the lines $\frac{x}{x_0} + \frac{y}{y_0} = 1$, where $x_0^2 + y_0^2 = 1$, $x_0, y_0 > 0$.

The segment $\overline{AB}$ has endpoints $A(\tfrac{1}{2}, 0)$ and $B(0, \tfrac{\sqrt{3}}{2})$. The equation of the line containing $\overline{AB}$ is $y = -\sqrt{3}x + \frac{\sqrt{3}}{2}$.

A point $C$ on $\overline{AB}$ can be parameterized as $C = (\frac{1}{2} - \frac{1}{2}\lambda, \frac{\sqrt{3}}{2}\lambda)$, where $0 \le \lambda \le 1$.

If $C$ belongs to a segment from $\mathcal{F}$ with endpoints $(x, 0)$ and $(0, y)$, then $\frac{\frac{1}{2} - \frac{1}{2}\lambda}{x} + \frac{\frac{\sqrt{3}}{2}\lambda}{y} = 1$.

We are looking for a unique point $C$ on $\overline{AB}$, distinct from $A$ and $B$, that does not belong to any segment from $\mathcal{F}$ other than $\overline{AB}$. This means that the only solution $(x, y)$ to the equation with $x^2 + y^2 = 1$, $x, y > 0$, is $(\frac{1}{2}, \frac{\sqrt{3}}{2})$.

The segments from $\mathcal{F}$ are parts of the tangents to the unit circle. The point $C$ must be tangent to the envelope of the family of segments. The envelope is the unit circle itself.

The unique point $C$ on $\overline{AB}$, distinct from $A$ and $B$, that lies on the unit circle. We need to solve:

$(\frac{1}{2} - \frac{1}{2}\lambda)^2 + (\frac{\sqrt{3}}{2}\lambda)^2 = 1$

$\lambda = \frac{1}{2}$. The point $C = (\frac{1}{4}, \frac{\sqrt{3}}{4})$.

$OC^2 = (\frac{1}{4})^2 + (\frac{\sqrt{3}}{4})^2 = \frac{1}{16} + \frac{3}{16} = \frac{4}{16} = \frac{1}{4}$.

$OC^2 = \frac{1}{4} = \frac{p}{q}$, so $p = 1$, $q = 4$.

$p$ and $q$ are relatively prime.

$p + q = 1 + 4 = 5$.

Final Answer: The final answer is $\boxed{5}$
}
\\
\bottomrule
\label{tab:rejection}
\end{longtable}

\section{Version Control}
\label{sec:vc}

\textbf{V2 → V3 (2025-03):}
\begin{itemize}
    \item Added \autoref{sec:s11}
    \item Added number of correct samples in \autoref{sec:data}
\end{itemize}

\textbf{V1 → V2 (2025-02):}
\begin{itemize}
    \item Added citations and other small writing changes
\end{itemize}

\end{document}